\documentclass[12pt]{article} 
\usepackage{helvet} 
\usepackage{graphicx} 
\usepackage{subcaption} 
\usepackage{multirow} 
\usepackage[table, dvipsnames]{xcolor} 
\usepackage{amsmath} 
\usepackage{bm} 
\usepackage{authblk} 
\usepackage{float} 
\usepackage{array} 
\usepackage{longtable} 
\usepackage{setspace} 
\usepackage{siunitx} 
\DeclareSIUnit{\wtpercent}{wt\%} 
\usepackage{nomencl} 
\makenomenclature 
\usepackage{etoolbox}
\renewcommand\nomgroup[1]{%
  \item[\bfseries
  \ifstrequal{#1}{A}{Abbreviations}{%
  \ifstrequal{#1}{S}{Symbols}{%
  \ifstrequal{#1}{G}{Greek letters}{}}}%
]}
\renewcommand{\figurename}{Fig.} 
\usepackage[sectionbib]{bibunits} 
\defaultbibliographystyle{ieeetr}
\defaultbibliography{bibliography}

\title{Determining Concentration from Raman Spectroscopy with Manifold Learning and Y-shaped Conformal Autoencoders}
\author[1]{Anna Taka}
\author[2]{Eleni D. Koronaki}
\author[3]{Ioannis G. Kevrekidis}
\author[4]{Eike Cramer}
\author[1]{Joern Viell}
\author[5,1,6*]{Alexander Mitsos}
\affil[1]{Process Systems Engineering (AVT.SVT), RWTH Aachen University, Forckenbeckstr. 51, 52074 Aachen, Germany}
\affil[2]{Luxembourg Institute of Science and Technology (LIST), 5 Av. des Hauts-Fourneaux, 4362 Esch-Belval Esch-sur-Alzette, Luxembourg}
\affil[3]{Department of Chemical and Biomolecular Engineering and Department of Applied Mathematics and Statistics, Whiting School of Engineering, John Hopkins University, 3400 North Charles Street, Baltimore, MD 21218, USA}
\affil[4]{Department of Chemical Engineering, University College London,  Gower St, London WC1E 6BT, United Kingdom}
\affil[5]{JARA-CSD, 52056 Aachen, Germany}
\affil[6]{Institute of Climate and Energy Systems, Energy Systems Engineering (ICE-1), Forschungszentrum Jülich GmbH, 52425 Jülich, Germany}
\affil[*]{Corresponding author: amitsos@alum.mit.edu}

\date{} 

\begin{document}
\begin{bibunit} 
\maketitle

\begin{abstract}
    \noindent Monitoring of chemical processes is crucial in chemical engineering, motivating sophisticated  process analytical technology (PAT). 
    In-line analytics are advantageous, with Raman spectroscopy being an established and reliable tool in various applications, from which many properties can be determined. We recently [Koronaki et al., AIChEJ 2024] proposed determining polymer particle size from Raman spectroscopic data, by combining Diffusion Maps (a manifold learning method) with a Y-shaped Conformal Autoencoder Neural Network.
    Herein, we extend this workflow to predict liquid concentrations, allowing also for multiple target predictions from the same model.
    We apply the methods on two datasets consisting of Raman measurements and concentration measurements: one from microgel polymerization and one from citric acid dissociation.
    The single-target concentration prediction results, where one target is determined from each model at a time, are accurate.
    Comparing against dedicated methods such as Indirect Hard Modeling (IHM) is not straightforward, due to the different validation strategies and the considered spectra range; nevertheless, it is fair to say that IHM performs slightly better than the proposed workflow.
    We further compare this approach with other regression methods, such as artificial neural networks, extreme gradient boosting and partial least squares, and the results show that the Conformal Autoencoders achieve higher predictive performance, demonstrating that the workflow can also be used for concentration determination.
    Finally we find that single-target concentration predictions are more accurate than the multiple target predictions for both case studies.
    
\end{abstract}

\section{Introduction} \label{intro}
In chemical engineering, the process has a major influence on the final product properties. Thus, monitoring and controlling chemical processes using process analytical technology (PAT) tools is essential.
Offline analytics such as gas chromatography (GC), high performance liquid chromatography and nuclear magnetic resonance (NMR) spectroscopy have been applied to many processes, for instance in polymerization~\cite{NMR_polymerization, GC_polymerization}, or in acid production and purification processes~\cite{offline_analytics_carboxylic_acids, offline_analytics_itaconic_acid, offline_analytics_purification}.
However, substantial effort and time is required to get the desired information from the measurements, as these methods require sampling, sample preparation, and further analysis~\cite{julian16, Echtermeyer20}.
In-line analytics conversely, enable direct measurement in the process at short acquisition time~\cite{nomenclature_PAT, short_acquisition_times}, allowing control of the process.
Near-infrared (NIR), mid-infrared (MIR) and Raman spectroscopy are common in-line spectroscopic methods~\cite{inline_analytics_pharmaceutic, inline_analytics_overview, inline_analytics_PAT_assesment}. 
Raman measurements are barely affected by the scattering of water~\cite{Raman_advantages}, making the method an established and reliable tool for different applications~\cite{Raman_polymerization_Asua, Raman_polymerization_ps_Lapkin, Raman_VCL_NIPAM_MeyerKirschner2018}.

Raman spectroscopic measurements are typically used to monitor species concentrations~\cite{raman_spectroscopy_for_concentrations_Pollard2020, raman_spectroscopy_for_concentrations_Mayorova2024}.
In order to analyze the spectra, a method is required to correlate the spectroscopic signal information with concentration~\cite{Echtermeyer20}. Such methods include peak integration (PI)~\cite{concentration_determination_PI_BeerLambert_law_Beer1852, concentration_determination_PI_AlKhanbashi1998, concentration_determination_PI_Hergeth1998}, partial least squares (PLS)~\cite{concentration_determination_PLS_Wold2001}, multivariate curve resolution-alternating least squares (MCR-ALS)~\cite{concentration_determination_mcr-als_Garrido2008} and indirect hard modeling (IHM)~\cite{concentration_determination_IHM_Alsmeyer_Marquardt2004, concentration_determination_IHM_Kriesten_Marquardt2008, concentration_determination_IHM_Kriesten_Mayer_Marquardt2008}.

Another property of interest is particle size. Particles affect spectroscopic measurements through light scattering~\cite{particles_affect_scattering_process_bohren2008absorption}. Moreover, it is shown that there is a correlation between the Raman scattering and polymer particle size~\cite{correlation_Raman_ps_Reis2003}.
Early methods to correlate the spectroscopic signal information with polymer particle size include PLS or Principal Component Analysis (PCA), both based on linear approaches~\cite{ps_determination_IHM+PLS_MeyerKirschner2018, ps_determination_PLS_Lapkin2015, correlation_Raman_ps_Reis2003}.
We recently~\cite{koronaki24} proposed the use of manifold learning approaches, namely Diffusion Maps (DMAPs): in a first step, DMAPs identifies an underlying low-dimensional manifold of the spectra; in a second step this reduced representation of the spectra is the input to a Y-shaped Conformal Autoencoder (Y-AE)~\cite{evangelou22} which predicts the polymer particle size.
DMAPs are based on mathematical diffusion processes on a graph constructed from the data and aim to discover the low-dimensional underlying manifold from which the data is sampled ~\cite{coifman06}. By employing DMAPs, we achieved significant dimensionality reduction of the input spectra, deriving a reduced representation of six dimensions from the initial $11084$ Raman wavenumbers.
After the dimensionality reduction, the reduced representation of the spectra found with DMAPs was used as input to a Y-AE to predict the polymer particle sizes.
The results indicate that the proposed method is very promising for online polymer size determination.

A similar workflow was proposed by Wang et al.~\cite{SimilarApproachToDMAPs+Y-AE_NIR_WANG2020} to predict the gasoline octane number from NIR spectroscopic measurements~\cite{NIR_KALIVAS1997}. 
The authors combined covariance matrix estimation (CME) and Landmark-Isometric feature mapping (L-Isomap) to reduce the dimensionality of the NIR spectra, and in a second step they proposed a neural network method, called the Beetle Antennae Search optimization Back Propagation neural network (BAS-BP), to predict the octane number from the reduced spectra representation.

In this contribution, we aim to determine species concentrations with the DMAPs and Y-AE approach to evaluate the capability of the approach, and to predict multiple targets simultaneously from the same model. 
We use data from two case studies in which we had applied IHM: microgel polymers from Meyer-Kirschner et al.~\cite{julian16}; citric acid solutions from Echtermeyer et al.~\cite{Echtermeyer20,Echtermeyer2024_dataset}.

We follow the same workflow as in our recent publication~\cite{koronaki24}, i.e., applying initially DMAPs to find the low-dimensional underlying manifold of spectra on both case studies, and then using the manifold as input to a Y-AE to predict the species concentrations separately as well as simultaneously.
We also apply three other regression methods: artificial neural networks (ANN) and Extreme Gradient Boosting, (XGB), to predict the targets from the reduced representation of the spectra found with DMAPs, and Partial Least Squares regression (PLS) to predict the targets directly from the spectra measurements, similarly to~\cite{koronaki24}.

The article is structured as follows: Section~\ref{datasets} presents the process of data collection for the two case studies. Section~\ref{IHM, PLS, DMAPS, Y-AEs} reviews the IHM, PLS, DMAPs and Y-AEs. 
Section~\ref{DMAPs results, predictions} presents the underlying manifolds of each dataset determined with DMAPs and the concentration prediction results comparing Y-AEs, ANN, XGB and PLS. Finally, in Section~\ref{conclusions}, we draw conclusions and suggest future work.

\section{Case studies}\label{datasets}
We analyze two case studies, using the data from Meyer-Kirschner et al.~\cite{julian16} and Echtermeyer et al.~\cite{Echtermeyer20,Echtermeyer2024_dataset}.
In the following, Section~\ref{julian_data} describes the process of data collection for the dataset from Meyer-Kirschner et al.~\cite{julian16} and Section~\ref{alex echte_data} reviews the process of data collection for the dataset from Echtermeyer et al.~\cite{Echtermeyer20}.

\subsection{Microgel synthesis dataset}\label{julian_data}
Meyer-Kirschner et al.~\cite{julian16} considered microgels based on N-vinylcaprolactam (VCL, 98$\%$ purity), which starting materials were distilled under a vacuum for purification, and cross-linked via N,N'-methylenbisacrylamide (BIS, \(>99.5\)$\%$ purity). 
The microgel synthesis procedure is described by Pich et al.~\cite{synthesis_procedure_Pich2006}, and the experimental setup is summarized in Meyer-Kirschner et al.~\cite{julian16}. 
The microgels (PVCL) underwent dialysis for purification, where the surfactant is assumed to be washed out, without affecting the final yield. After purification, the correlation between the spectroscopic signal information and the concentration will be more significant.

The authors prepared an additional set of 31 samples for the model calibration, consisting of aqueous monomer solutions (up to 1.47wt$\%$ VCL), aqueous microgel solutions (up to 2.22wt$\%$ PVCL), and solutions of both monomer and microgels in water, and acquired further offline Raman measurements of the samples at room temperature. After a preliminary study that we conduct, we notice that the measurements for one sample with high polymer concentration significantly deteriorate the results. A similar behavior was observed also by Meyer-Kirschner et al.~\cite{julian16}, where it is explained that this could be due to the fact that the dispersion with high microgel fraction cannot be mixed ideally anymore, creating a microgel-rich fraction at the bottom of the vial. Hence, this sample is excluded from our study, with the data consisting of Raman spectra and concentration values of 30 monomer and microgel samples.

For all samples we consider 11084 wavenumbers per spectrum in the Raman shift range between 99.9 cm\(^{-1}\) and 3424.8 cm\(^{-1}\).
Then we do peak maximum normalization on the intensities, introducing thus spectra pretreatment. We are aware that having sample preparation and spectra pretreatment present is not the ideal case for applying DMAPs, as DMAPs are robust to noise perturbation~\cite{Delaporte2008AnIT}, but we do that since the dataset is small.

\subsection{Citric acid dissociation dataset}\label{alex echte_data}
Echtermeyer et al.~\cite{Echtermeyer20} considered four carboxylic acid solutions, namely itaconic acid (Alfa Aesar, \(\geq99\)$\%$), formic acid (Merck, \(98-100\)$\%$), acetic acid (Merck, \(99-100\)$\%$) and citric acid (KMF Laborchemie, \(\geq99\)$\%$). The authors prepared the solutions with defined concentrations in deionized water for titration with non-Raman-active, aqueous sodium hydroxide NaOH (VWR Chemicals, 1 mol L\(^{-1}\)).
Raman spectra were recorded using a RXN2 Raman Analyzer (Kaiser Optical Systems) at 785 nm with 400 mW.
We use the data of the citric acid solution, containing 45 datapoints. The citric acid dataset consists of Raman spectra and concentration values of the citric acid and its dissociated acid species in water, CA\(^{-}\), CA\(^{2-}\), CA\(^{3-}\). The Raman intensities of the spectra are measured over a Raman shift range of 99.9 cm\(^{-1}\) and 3424.8 cm\(^{-1}\), relating to 11084 wavenumbers per spectrum.

\section{Methods} \label{IHM, PLS, DMAPS, Y-AEs}
In the following, Section~\ref{IHM} reviews IHM for concentration determination, Section~\ref{PLS} reviews PLS for target prediction through linear dimensionality reduction of the input, Section~\ref{DMAPS} reviews the DMAPs approach for dimensionality reduction, and Section~\ref{Y-AEs} reviews the Y-AEs for target estimation.

\subsection{Indirect Hard Modeling} \label{IHM}
Indirect Hard Modeling (IHM) is a dedicated method for concentration determination from Raman spectra. 
Unlike MCR-ALS, which is based on ``soft models'' that arrange the data in matrices and rely on statistics and data variance within the matrices~\cite{concentration_determination_mcr-als_Garrido2008}, IHM employs ``hard models'' of each component present in the reaction by summation of parametrized peak functions (Gaussian and Lorentzian functions, i.e., pseudo-Voigt profiles)~\cite{concentration_determination_IHM_Alsmeyer_Marquardt2004, concentration_determination_IHM_Kriesten_Mayer_Marquardt2008}.
Subsequently, the ``hard model'' of the mixture is formed as a weighted sum of the pure component models.
The component weight parameters are then calibrated (small datasets are sufficient) to predict concentration values from new mixture spectra measurements.
Thus, the model can be used to monitor the concentration from measured Raman spectra~\cite{julian16, Echtermeyer20}.

The mass balance information can also be included as an additional equation in the regression of the component weights, resulting to the regression of weight ratios, i.e., ratiometric regression, which neutralizes the impact of intensity variations, e.g., by turbidity, excitation intensity, volume density or measurement time.
IHM is available in the commercial software PEAXACT~\cite{PEAXACT_IHM}.
Typically, the spectra are pretreated and only the fingerprint region is considered, since it contains most of the information. The prediction accuracy of the model is typically validated using the Root-Mean-Squared-Error of Cross-validation (RMSECV) metric.

\subsection{Partial Least Squares regression} \label{PLS}
Partial Least Squares regression (PLS) is a statistical method that finds the correlation between input variables and target variables by projecting them into a new space of uncorrelated latent variables~\cite{concentration_determination_PLS_Wold2001}.
The uncorrelated latent variables are created by finding linear combinations of the input and target variables that maximize their covariance.
During this step, a dimensionality reduction of the input variables is achieved when the number of the latent variables is smaller than the dimensions of the input variables.

On a second step, a linear regression is performed using the latent variables (instead of the initial input variables) to predict the target variables. Both single target predictions (known as PLS1) and multiple target predictions (PLS2) are available on the scikit-learn library~\cite{scikit-learn}.
PLS bears similarity to principal component analysis.

\subsection{Diffusion Maps} \label{DMAPS}
Diffusion Maps (DMAPs) are a mathematical technique for dimensionality reduction. 
They are based on diffusion processes on a graph constructed from the data, and aim to find the low-dimensional geometric description from which the data is sampled. Dimensionality reduction is achieved by selecting only a few parameters (the components of the data in the leading eigenvectors of the graph Laplacian of this graph) for the representation of the initial high-dimensional dataset~\cite{coifman06}. 
These parameters are called coordinates.
The DMAP coordinates are chosen through the decay of the eigenvalues of the matrix~\cite{coifman06}, defining a parametrization of the low-dimensional representation of the dataset, also called manifold.
The DMAP coordinates can also be selected based on the residuals between the eigenvector sets after performing a local regression~\cite{LocalRegressionSelection_Dsilva2018}. In that case, the eigenvectors with the larger residuals are selected for the manifold representation as they add more information.

The next step consists of the inverse mapping from the reduced to the initial ambient space. The mapping from low to high dimensions is achieved with the Laplacian pyramid extension \cite{Lapla_Pyramids_1_Rabin2012, Lapla_Pyramids_2, DATAFOLD_Lehmberg2020}. This is to ensure that the manifold is accurate.

After finding the low-dimensional manifold, it is possible to map between the initial ambient space and the reduced space. This is to extend functions that are defined on the dataset and evaluate them outside of it, for points called ``out-of-sample''. This mapping from high to low dimensions is performed with the Nyström extension, which uses some functions defined on the dataset, called Geometric Harmonics~\cite{koronaki24, coifman06}.
For more details on the DMAPs implementation, we refer to~\cite{coifman06}.

\subsection{Y-shaped conformal Autoencoders} \label{Y-AEs}
\begin{figure} [ht]
    \centering
    \includegraphics[width=0.5\linewidth]{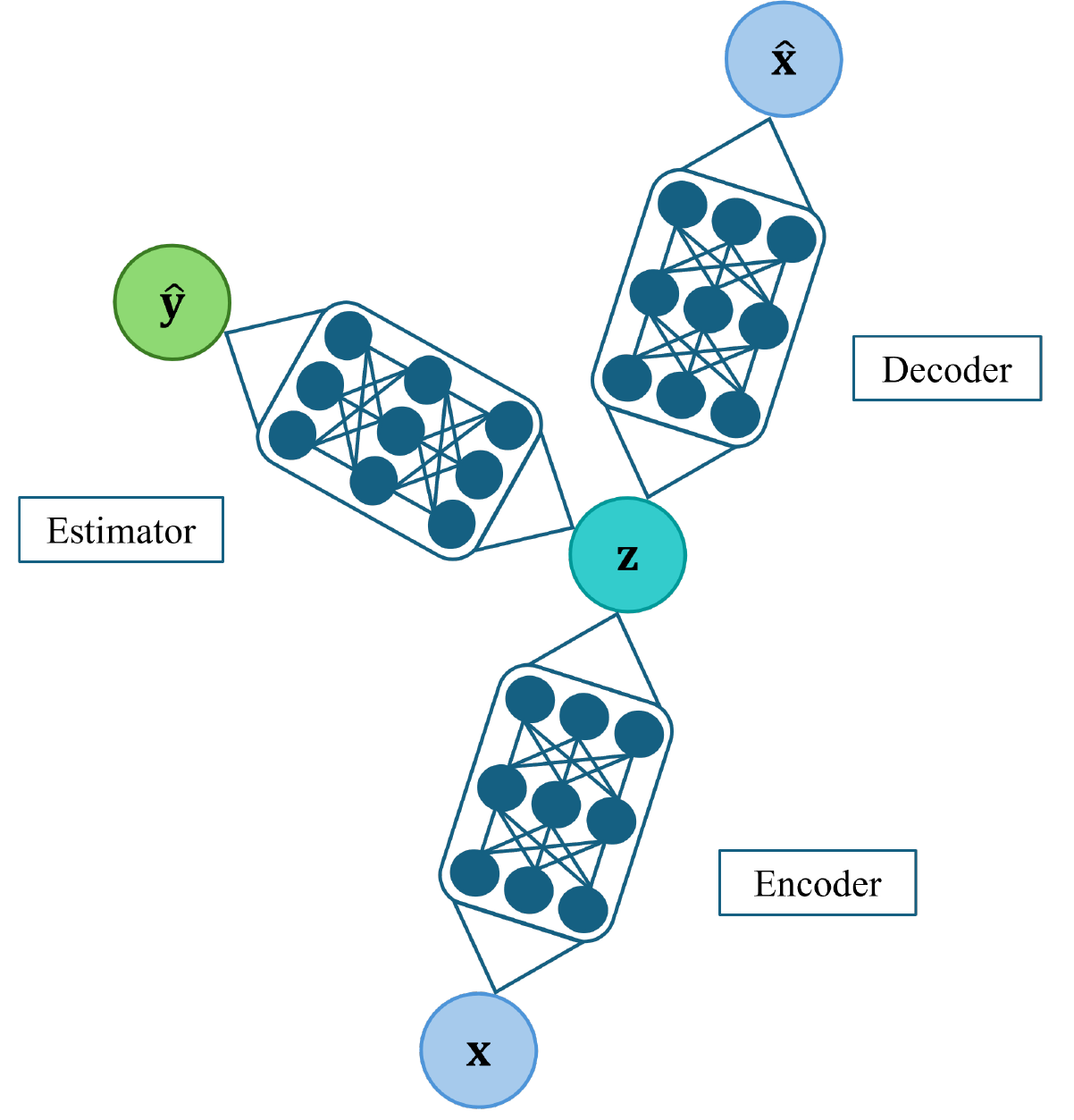}
    \caption{The Y-shaped Conformal Autoencoder combines a conformal autoencoder with a neural network, thus consisting of three connected subnetworks: the encoder, the decoder, and the estimator. The encoder maps the inputs to the autoencoder latent variables, the decoder does the reverse mapping, from the autoencoder latent variables to the inputs, and the estimator maps from a subset of the autoencoder latent variables that affect the outputs to the desired targets.}
    \label{fig:Y-AE}
\end{figure}

Y-shaped Conformal Autoencoders (Y-AEs)~\cite{evangelou22} combine a conformal autoencoder with a neural network, comprising three connected subnetworks, as shown in Fig.~\ref{fig:Y-AE}. 
Y-AEs present an ANN-based regression model with additional capabilities. 
An orthogonality constraint is imposed on the autoencoder's latent variables. As a result, Y-AEs first disentangle the combination of the inputs that affect the outputs from those that do not affect them, and then they predict the outputs. When implementing Y-AE, the outputs are defined as a function of the latent variables that affect them, and we believe that even in the cases where the data are limited, the predictions of Y-AE will be accurate.

The three constituent neural networks of a Y-AE are the encoder, the decoder, and the estimator. 
The encoder maps the inputs to the autoencoder latent variables, disentangling what matters to the outputs from what does not, $\textbf{x} \mapsto \textbf{z}$. 
The decoder reconstructs the inputs by doing the reverse mapping, from the autoencoder latent variables to the inputs, $\textbf{z} \mapsto \hat{\textbf{x}}$.
Finally, the estimator maps from a subset of the autoencoder latent variables that affect the outputs to the desired targets, $\textbf{z} \mapsto \hat{\textbf{y}}$.

The loss function for the training of Y-AE consists of three components:
\[\mathcal{L} = \mathcal{L}_{AE} + \alpha \mathcal{L}_{\bm{z}} + \mathcal{L}_{E}\]
Here, $\alpha$ is a hyperparameter used to scale the relative importance of the orthogonality constraint.
The first component describes the reconstruction loss of the autoencoder section of the Y-AE:  
\[\mathcal{L}_{AE} = MSE(\hat{\textbf{x}}, \textbf{x})\]
Here, \textbf{x} is the vector of the input data, $\hat{\textbf{x}}$ is the vector of the reconstructed inputs, and MSE denotes the Mean-Squared-Error (MSE) metric.
The second component comes from implementing the orthogonality constraint
\[\langle \bm{dz_i}, \bm{dz_j} \rangle = 0 , \forall i \neq j\]
on the autoencoder's latent variables \textbf{z} as penalty:
\[\mathcal{L}_{\bm{z}} = \sum_{(i,j): j>i} MSE (\langle \bm{dz_i}, \bm{dz_j} \rangle, 0)\]
Here, $\bm{dz_i}$ indicates the vector of partial derivatives of the latent variable $z_i$ with respect to the inputs, and $\langle \cdot,\cdot \rangle$ indicates the inner dot-product.  
The third loss component describes the reconstruction loss of the estimator:
\[\mathcal{L}_{E} = MSE(\hat{\textbf{y}}, \textbf{y})\]
Here, \textbf{y} is the vector of the true output values, i.e., the data, and $\hat{\textbf{y}}$ is the vector of the predicted output values from the network.
For more details on the Y-AEs implementation, we refer to the original publication~\cite{evangelou22}. 
We use the same training approach as explained in~\cite{koronaki24}.

\section{Results}\label{DMAPs results, predictions}
In the following, Section~\ref{manifolds} presents the representation of the low-dimensional underlying manifold of Raman spectra for both case studies, Section~\ref{separate predictions} presents the separate concentration prediction results with Y-AE, ANNs and XGB for both case studies, Section~\ref{simultaneous predictions} presents the simultaneous concentration prediction results with Y-AE for both case studies and Section~\ref{summary} presents a summary of all the results for both case studies. 

\subsection{Low-dimensional parametrization of spectra} \label{manifolds}
\begin{figure}[ht] 
\centering
\begin{subfigure}{1\textwidth}
\includegraphics[width=1\linewidth]{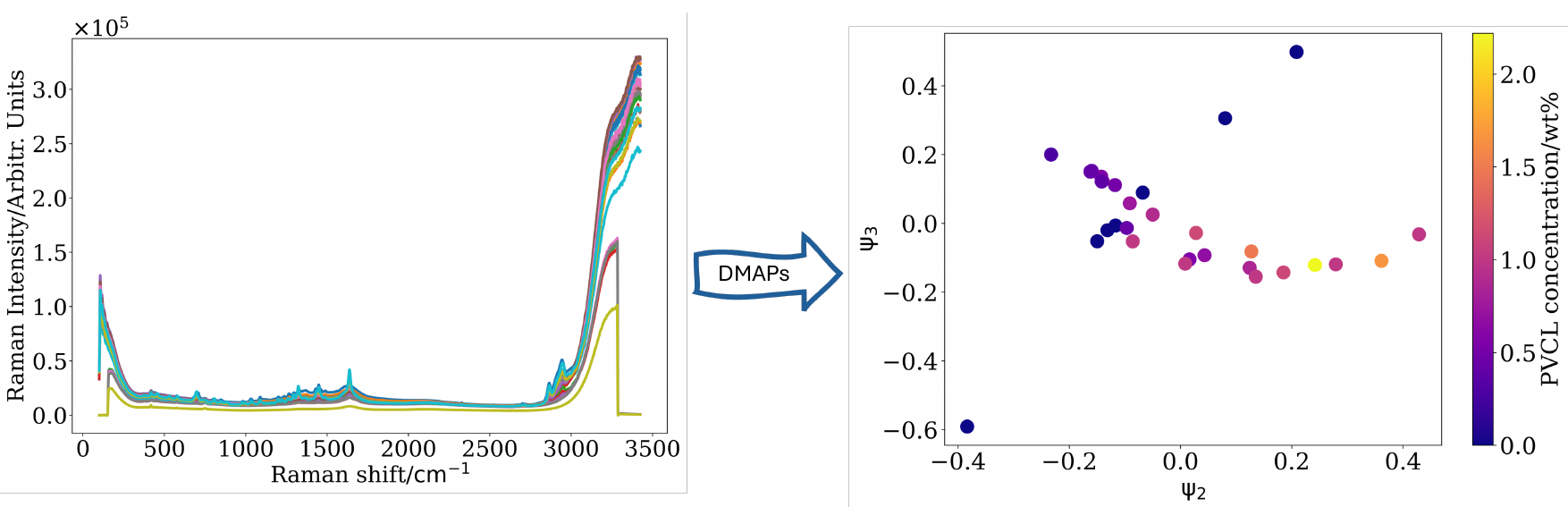} 
\caption{}
\label{fig:manifold_Raman_case 1}
\end{subfigure}

\begin{subfigure}{1\textwidth}
\includegraphics[width=1\linewidth]{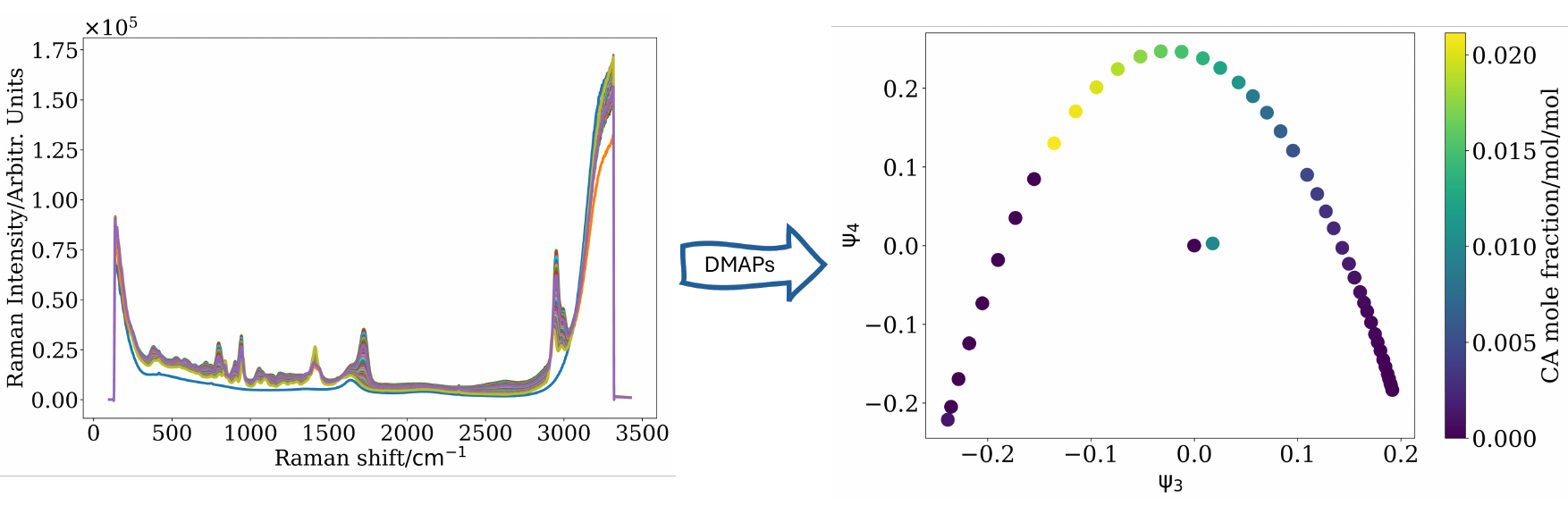}
\caption{}
\label{fig:manifold_Raman_case 2}
\end{subfigure}
\caption{Low-dimensional parametrization of Raman spectra found with DMAPs, (a) for the first case study and (b) for the second case study. For both case studies, on the left we see the Raman spectra measured for all the samples: each line presents the measured spectrum of one sample. On the right we see the parametrization of the low-dimensional underlying manifold found with DMAPs. In both representations we see that there are some offset points: this is due to the value of the kernel scale.}
\label{fig:manifolds_Raman}
\end{figure}

We initially apply DMAPs to find the underlying manifold of the Raman spectra for both case studies, as shown in Fig.~\ref{fig:manifolds_Raman}.

\begin{figure}[p] 
    \centering
    \begin{subfigure}{1\textwidth}
    \includegraphics[width=1\linewidth]{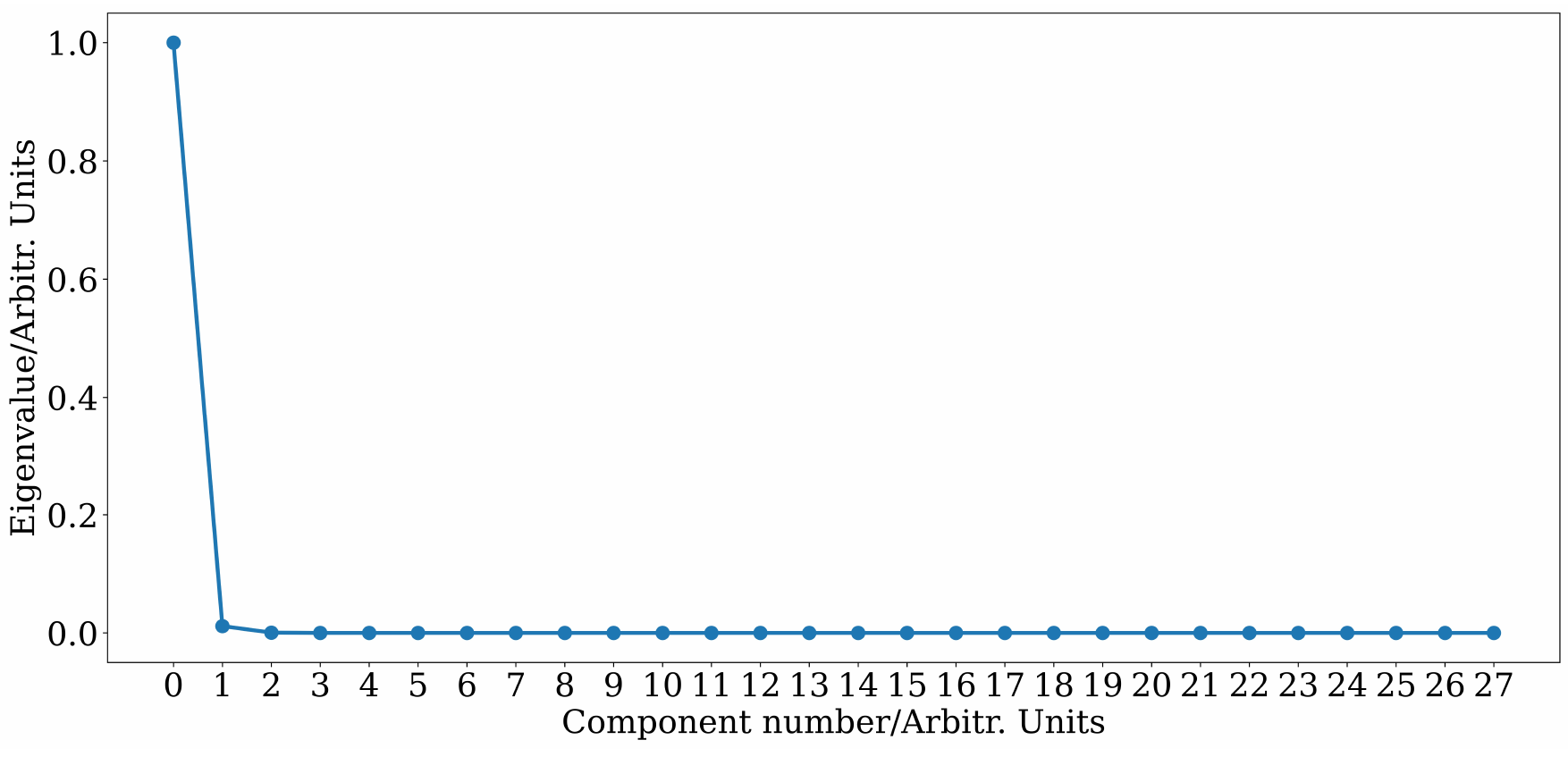}
    \caption{}
    \label{fig:scree_plot_case_1}
    \end{subfigure}
    
    \begin{subfigure}{1\textwidth}
    \includegraphics[width=1\linewidth]{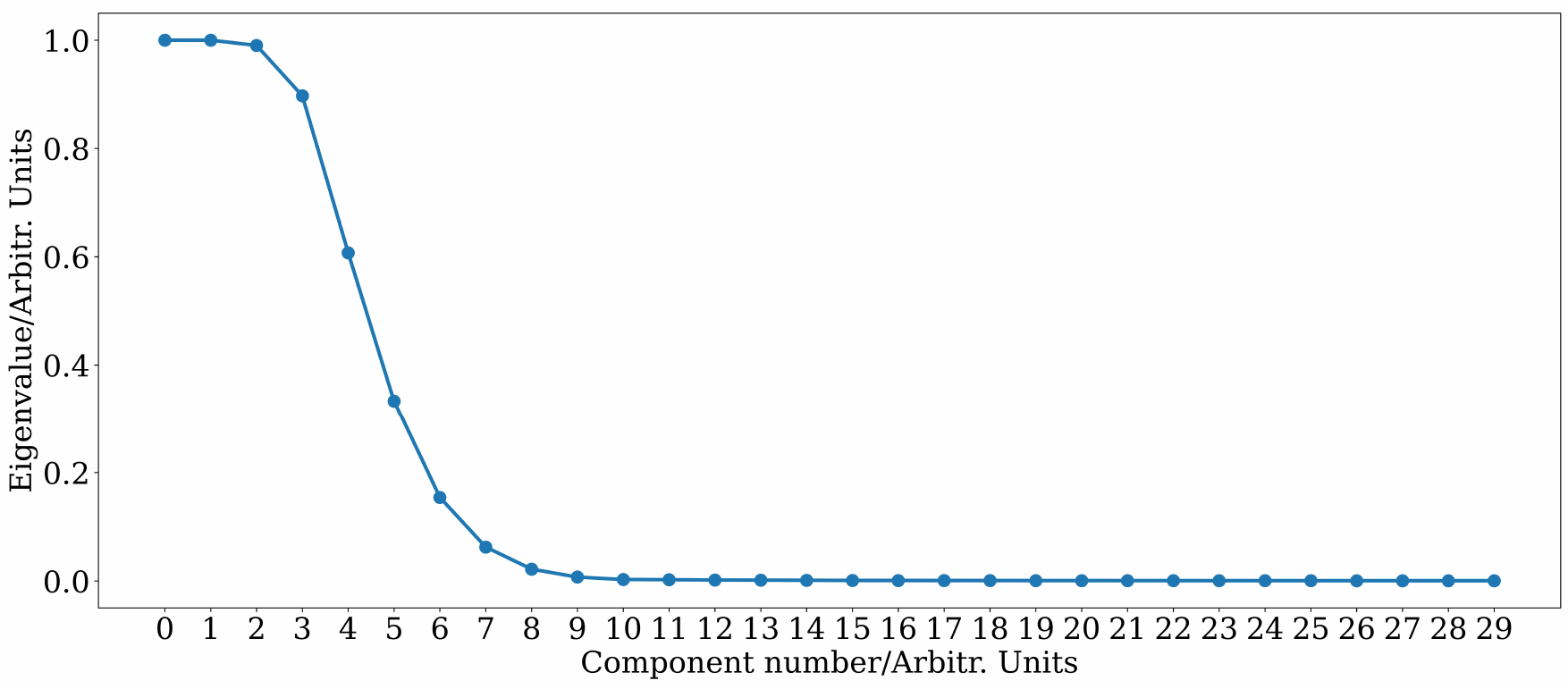}
    \caption{}
    \label{fig:scree_plot_case_2}
    \end{subfigure}
    
    \caption{Scree plot of the eigenvalues of the matrix from DMAPs on the spectra of the first case study (a) and for the second case study (b). For both case studies, the first eigenvalue is equal to one and the corresponding eigenvector is constant, thus we do not consider it for the selection.}
    \label{fig:scree_plots}
\end{figure}

We select seven DMAP coordinates $(\psi_1, \psi_2, \psi_3, \psi_4, \psi_5, \psi_6, \psi_7)$ to represent the spectra for the first case study, and two DMAP coordinates $(\psi_3, \psi_4)$ to represent the spectra for the second case study, both initially consisting of 11084 wavenumbers per spectrum.
We make the selection based on the decay of the eigenvalues of the scaled affinity matrix~\cite{coifman06}, and the accuracy of the reconstruction of the original spectra from the selected coordinates.
The eigenvalues of the matrices for each case study are presented on Fig.~\ref{fig:scree_plots}.

For the first case study, the value of the kernel scale is large and the decay of the eigenvalues is steep. With the first eigenvalue $(\lambda_0)$ being equal to one, the corresponding eigenvector is constant, thus we do not consider it for the selection. Starting from $\lambda_1$, and keeping the corresponding eigenvector $(\psi_1)$ as the first dimension of the manifold, we continue adding more dimensions to the parametrization of the manifold based on the reconstruction of the original spectra, through the inverse mapping from the reduced to the initial ambient space, with the Laplacian pyramid extension. The eigenvectors whose eigenvalues have a high value, add more information, and we consider them for the low-dimensional embedding. Thus we get the seven-dimensional parametrization shown in Fig.~\ref{fig:manifold_Raman_case 1}.
Similarly, we select the two-dimensional representation for the second case study, shown in Fig.~\ref{fig:manifold_Raman_case 2}.

\begin{figure}[p] 
    \centering
    \begin{subfigure}{1\textwidth}
    \includegraphics[width=1\linewidth]{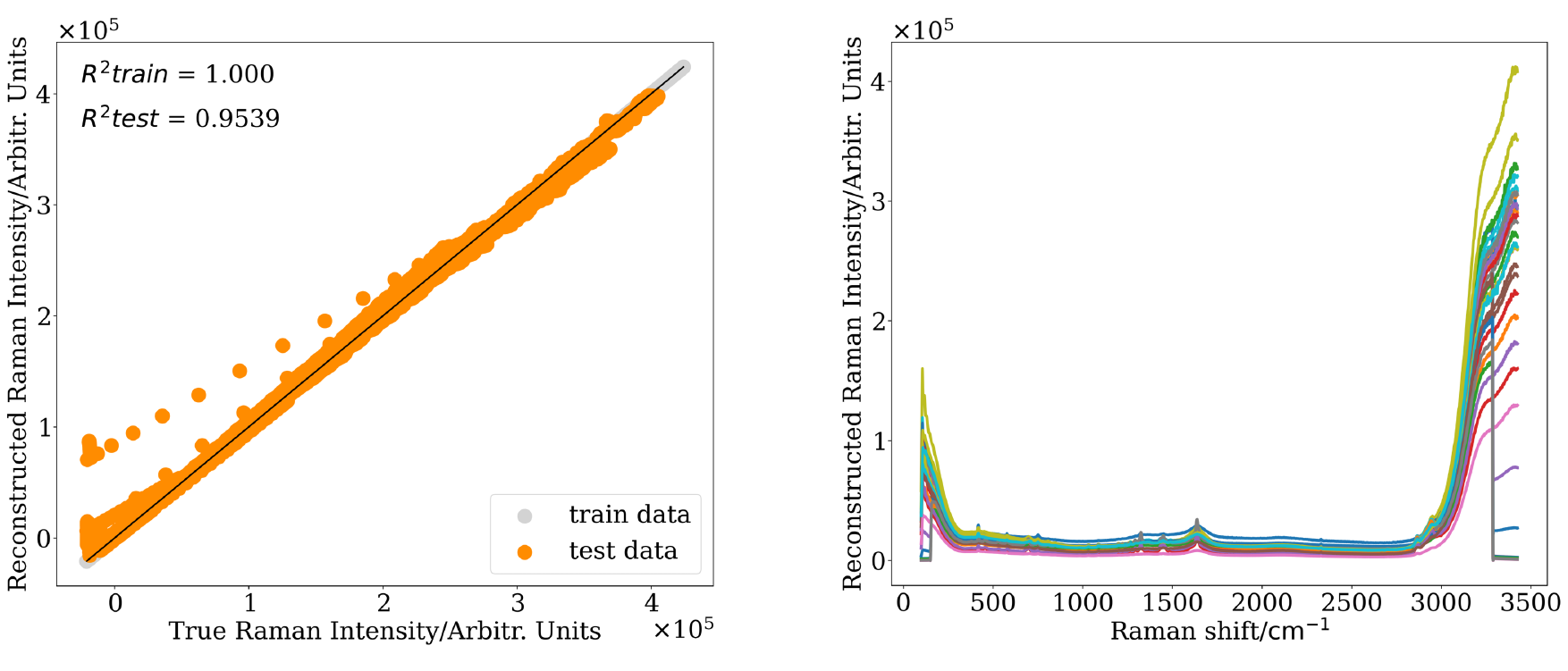}
    \caption{}
    \label{fig:reconstruction_1}
    \end{subfigure}
    
    \begin{subfigure}{1\textwidth}
    \includegraphics[width=1\linewidth]{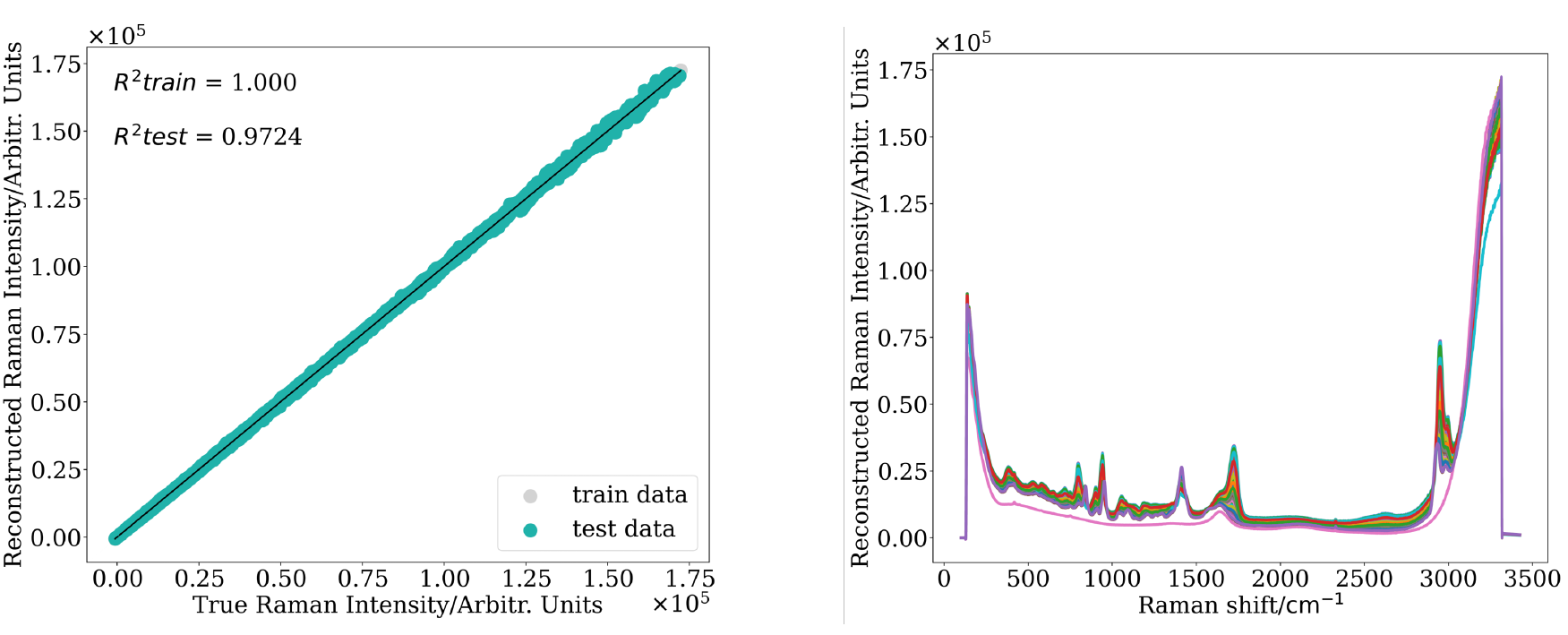}
    \caption{}
    \label{fig:reconstruction_2}
    \end{subfigure}
    
    \caption{Reconstructed spectra representation and parity plot from the selected parametrization, (a) for the first case study  and (b) for the second case study. On the left we see the parity plots, where the gray dots (aligned through the diagonal line) represent the training datapoints for both case studies and the colored ones represent the testing datapoints. On the right we see the reconstructed spectra from the selected parametrization, where each line represents the spectrum of each reconstructed datapoint, for both case studies.}
    \label{fig:dmaps_reconstrucions}
\end{figure}

For the inverse mapping from the reduced to the ambient space with the Laplacian pyramid extension, we split the true spectra dataset and the reduced parametrization dataset and we use 75$\%$ of the datasets for the training and 25$\%$ for the testing. For the first case study we get an $R^2=0.9539$ for the reconstructed test spectra from the selected parametrization and $R^2=0.9724$ from the parametrization for the second case study. The parity plots and the reconstructed spectra are shown in Fig.~\ref{fig:dmaps_reconstrucions}. More details about the manifolds selection are explained on Section~\ref{SI_DMAPs study} of the Supporting Information.

\subsection{Separate concentration prediction results}\label{separate predictions}
In the following, Section~\ref{sep_predictions_case_1} and Section~\ref{sep_predictions_case_2} present the concentration prediction results with Y-AE, ANNs, XGB and PLS, for the first and the second case study respectively.
We use 75$\%$ of the datasets for the training and 25$\%$ for the testing on each prediction method, for both case studies. The value range of the targets on both case studies is summarized in Table~\ref{tab:labels_range}.
The evaluation of the prediction performance of the methods is based on the coefficients of determination (\(R^2\)), Root-Mean-Squared-Error (RMSE) and Mean-Absolute-Error (MAE) metrics for the training and the testing:
\[ R^2\ = 1 - \frac{SS_{res}}{SS_{tot}} = 1 - \frac{\sum_{i=1}^{n} (y_i - \hat{y}_i)^2}{\sum_{i=1}^{n} (y_i - \bar{y})^2} \]

\[ RMSE = \sqrt{\frac{1}{n}\sum_{i=1}^{n} (y_i - \hat{y}_i)^2} \]

\[ MAE = \frac{1}{n}\sum_{i=1}^{n} |y_i - \hat{y}_i| \]
Here, $y_i$ is the true target value, i.e., the data, $\hat{y}_i$ is the predicted target value, $\bar{y}$ is the mean of the true target values and $n$ is the total number of datapoints. In the \(R^2\) formula the $SS_{res}$ is the Residual Sum of Squares and the the $SS_{tot}$ is the Total Sum of Squares. In the MAE formula, $|y_i - \hat{y}_i|$ indicates the absolute error of each datapoint.

As the datasets are small, we perform an uncertainty quantification (UQ) study on the single-target predictions from Y-AE and ANN, by training ten models for each method, and compare the predictions for consistency. The results of the UQ are summarized in Fig.~\ref{fig:UQ_sep Y-AE_case 1} (Supporting Information) and Fig.~\ref{fig:UQ_ANN_case 1} (Supporting Information) for the first case study, for Y-AE and ANN respectively. For all the methods, for all the targets we generally have higher discrepancy for the predicted test datapoints compared to the predicted train datapoints. This is because the test datapoints are new data for the models. Nevertheless, the predictions from ten models for each method are consistent and on the same range with each other and with the predictions presented on Section~\ref{sep_predictions_case_1}.
For the second case study the results from the uncertainty quantification are summarized in Fig.~\ref{fig:UQ_sep Y-AE_case 2} (Supporting Information) and Fig.~\ref{fig:UQ_ANN_case 2} (Supporting Information) for Y-AE and ANN respectively. Similarly to the first case study, we see that there is a higher discrepancy for the predicted test datapoints than for the predicted train datapoints, even though the discrepancy for the predicted train datapoints for some of the targets is of a higher value in this case compared to what we see in the first case study.

\begin{table}[ht]  
    \centering
    \begin{tabular}{|c|c|c|}
    \hline
        Case study & Target & Value range \\
    \hline
        & VCL & \(0.000 - 1.470\) \\
        microgel synthesis & PVCL & \(0.000 - 2.220\) \\
        $(\si{\wtpercent})$ & $\rm{H2O}$ & \(97.52 - 100.0\) \\
    \hline
        & CA & \(0.000 - 0.0211\) \\
        & CA\(^{-}\) & \(0.000 - 0.01110\) \\
        citric acid dissociation & CA\(^{2-}\) & \(0.000 - 0.008300\) \\
        $(\si{mol \per mol})$ & CA\(^{3-}\) & \(0.000 - 0.01010\) \\
        & $\rm{H2O}$ & \(0.9785 - 1.000\) \\
    \hline
    \end{tabular}
    \caption{Targets value range. Water has the highest concentration values on both case studies.}
    \label{tab:labels_range}
\end{table}

\subsubsection{Case study 1: microgel synthesis data}\label{sep_predictions_case_1}
After the dimensionality reduction of the spectra, we compare three alternative machine learning methods for the target predictions, namely the Y-AE, ANNs, and extreme gradient boosting (XGB).
The inputs for the three methods are the DMAP coordinates representing the Raman spectra.
We also predict the targets with PLS, using the spectra measurements as inputs to the method.
The outputs for the four methods are the monomer, polymer and water concentration values.

\begin{figure}[p] 
    \centering
    \begin{subfigure}{0.4\textwidth}
    \includegraphics[width=1\linewidth]{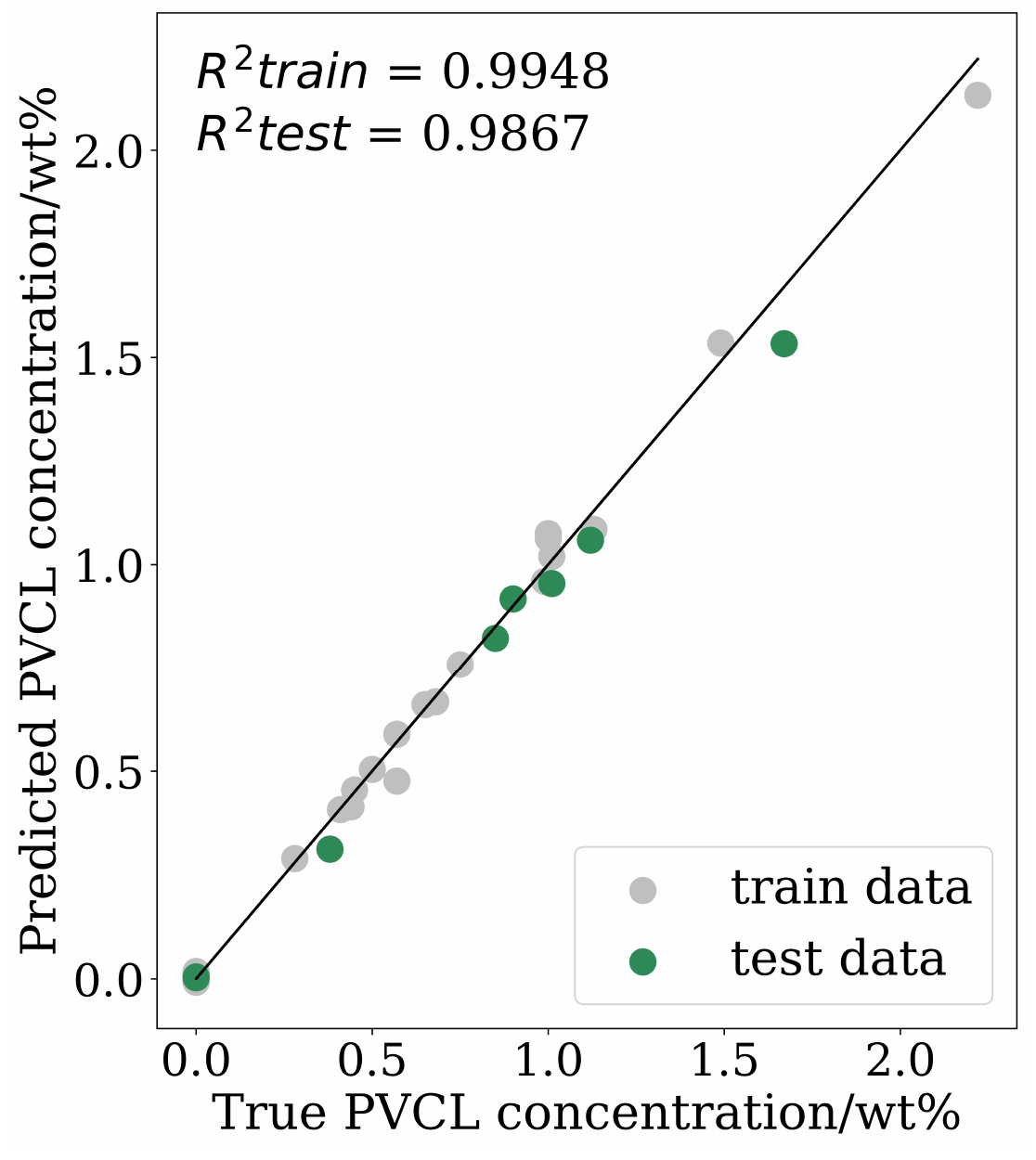}
    \label{fig:sep_Y-AE_case 1}
    \caption{}
    \end{subfigure}
    \hspace{0.05\textwidth}
    \vspace{0.05\textwidth}
    \begin{subfigure}{0.4\textwidth}
    \includegraphics[width=1\linewidth]{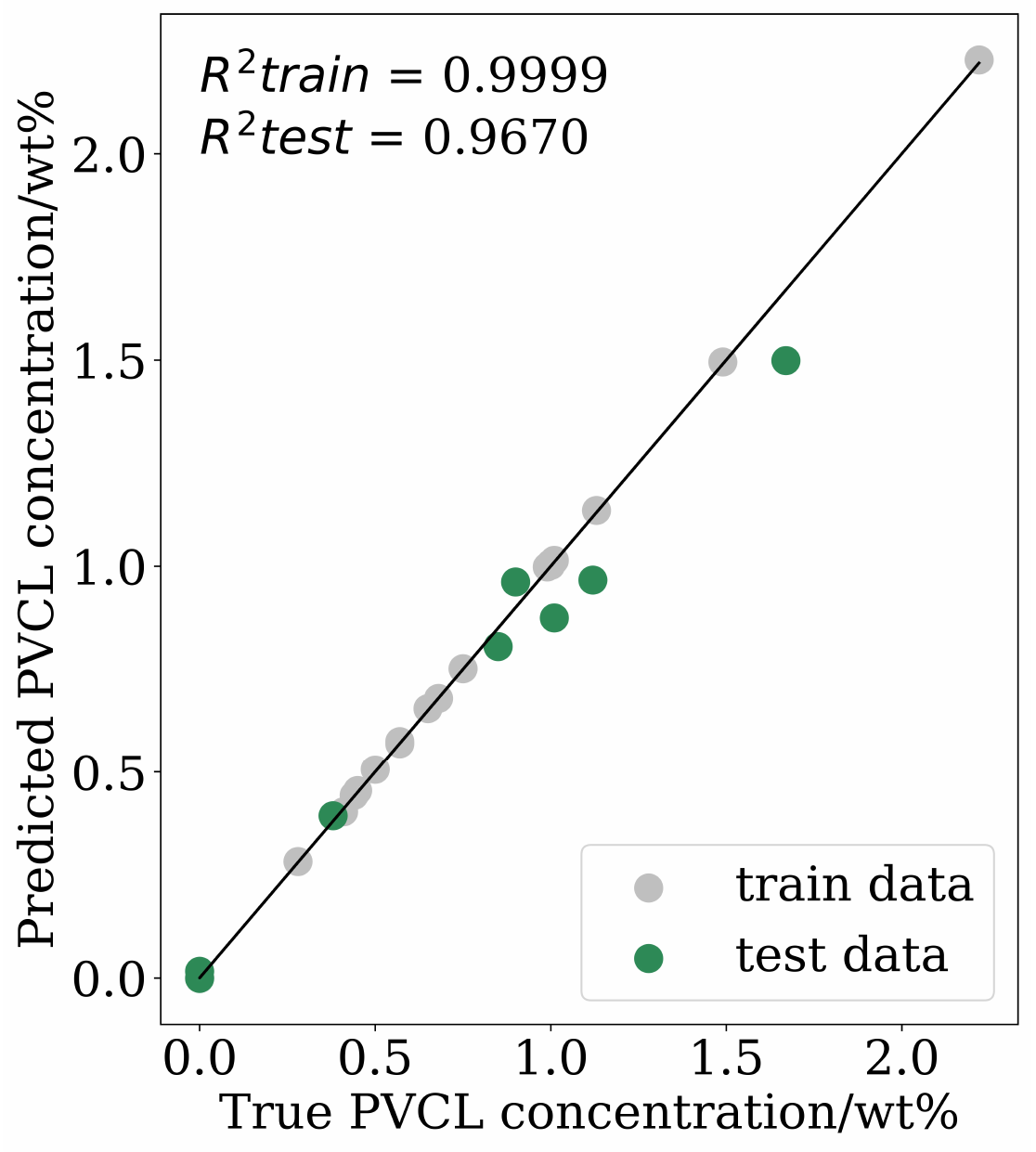}
    \label{fig:ANN_case 1}
    \caption{}
    \end{subfigure}
    \begin{subfigure}{0.4\textwidth}
    \includegraphics[width=1\linewidth]{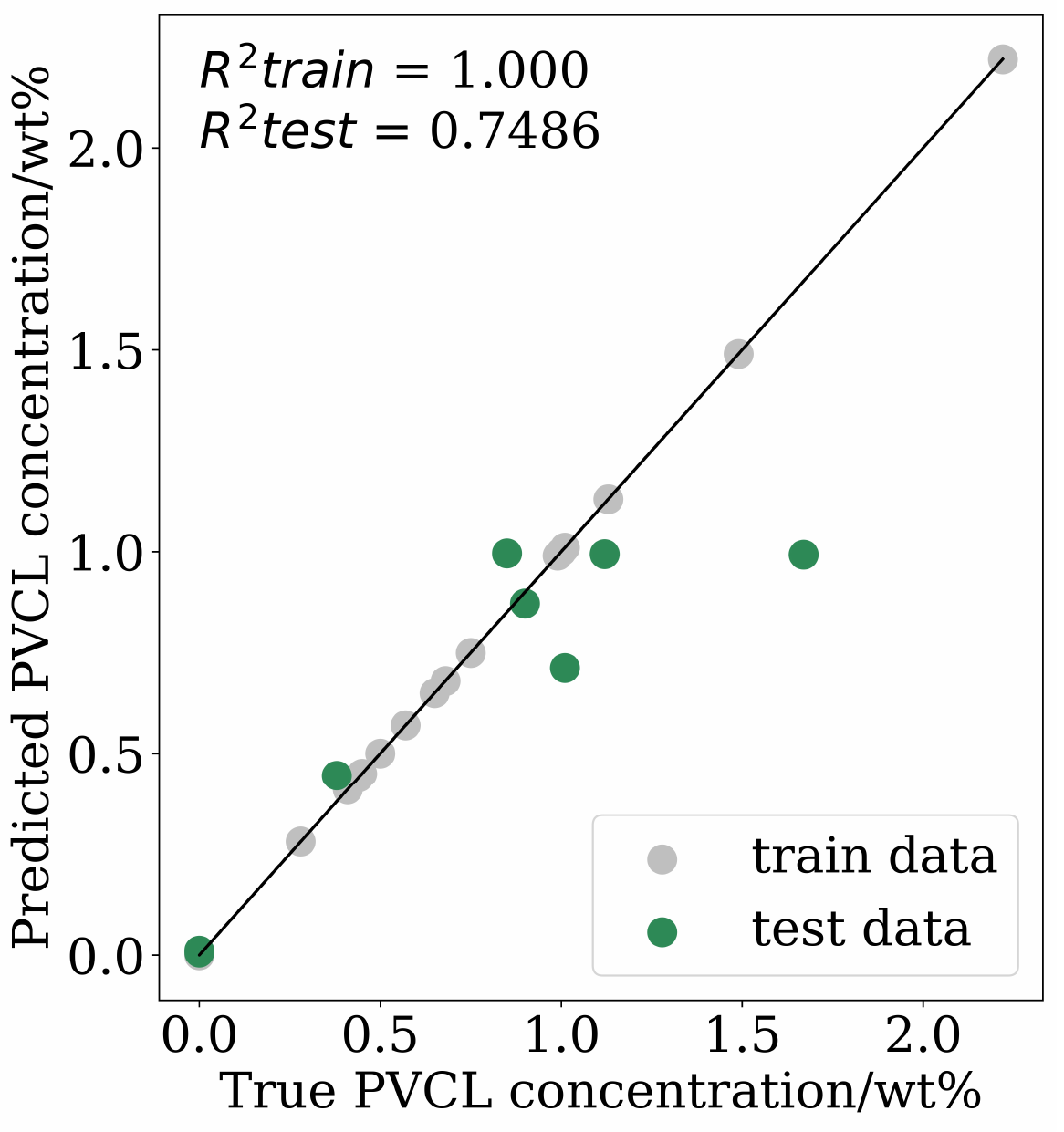}
    \label{fig:XGBOOST_case 1}
    \caption{}
    \end{subfigure}
    \hspace{0.05\textwidth}
    \begin{subfigure}{0.4\textwidth}
    \includegraphics[width=1\linewidth]{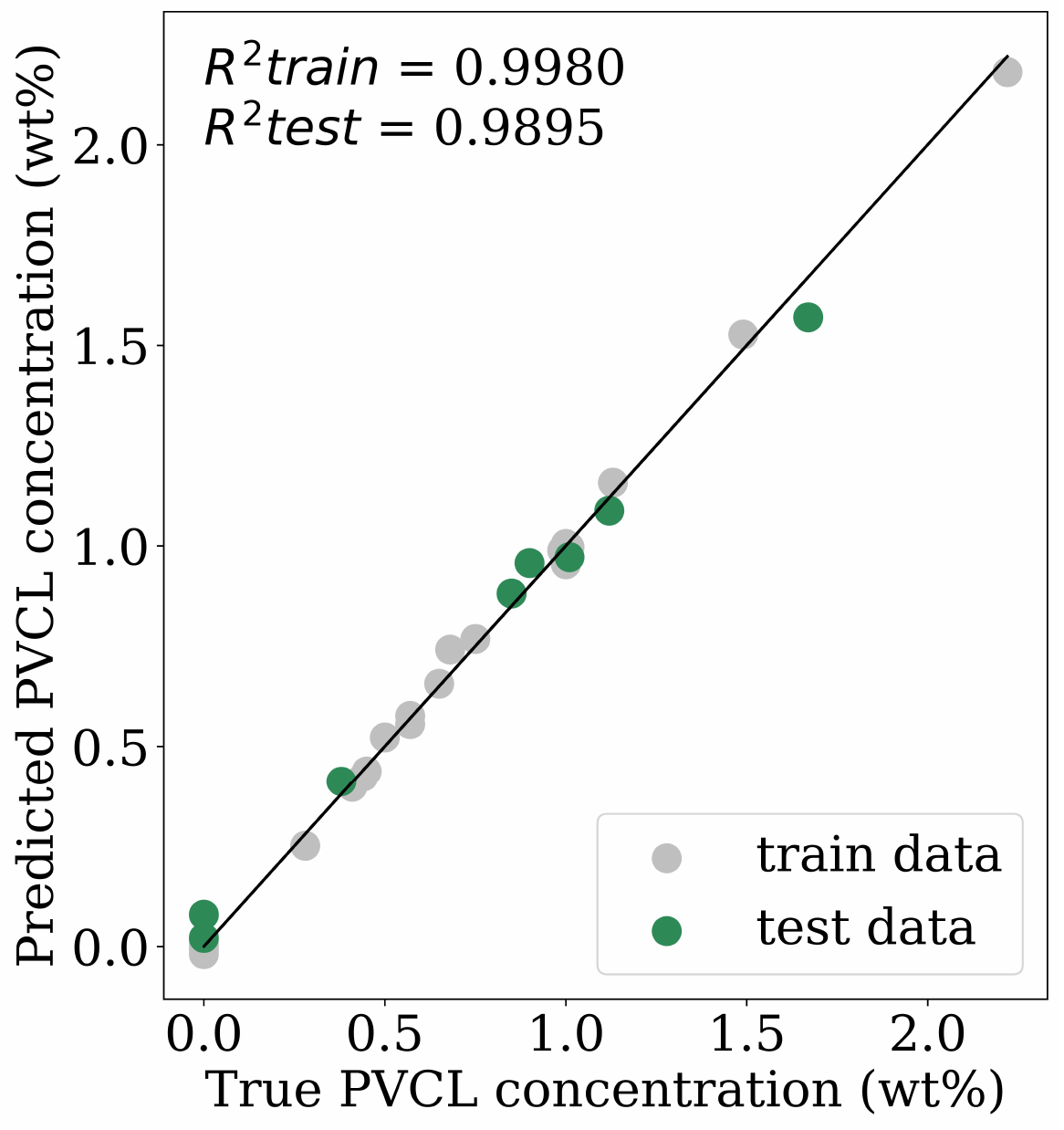}
    \label{fig:PLS_case 1}
    \caption{}
    \end{subfigure}
    \caption{Polymer concentration predictions with Y-AE (a), ANN (b), XGB (c), and PLS (d). The gray dots represent the training datapoints, while the colored ones represent the testing datapoints. PLS gives the best prediction results with very few latent variables: this indicates that the spectroscopic data lie on a linear low-dimensional manifold. The predictions from Y-AE and ANN are on the same range.}
    \label{fig:predictions_sep_case_1}
\end{figure}

First, we predict the targets with the Y-AE. For each target we train a Y-AE described in Section~\ref{Y-AEs}, with the DMAP coordinates as inputs to the encoder and also as outputs of the decoder. We then train the estimator, defining each target as a function of the first latent variable.
For the predictions using ANNs, we train a feed-forward neural network, and for the predictions with the extreme gradient boosting, we use the XGB algorithm.
For the predictions from PLS we find six, seven and eight latent variables for water, monomer and polymer respectively, based on the metrics for the training and testing.
The parity plots with the predictions from each method for polymer are shown in Fig.~\ref{fig:predictions_sep_case_1}. The plots for the remaining targets are shown in Fig.~\ref{fig:predictions_sep_water_case_1} (Supporting Information) and Fig.~\ref{fig:predictions_sep_monomer_case_1} (Supporting Information).

Based on the metrics and the parity plots, we see that Y-shaped autoencoders achieve high prediction performance, even though the predictions from ANN and Y-AE are on the same range, while PLS gives the highest prediction results. Since we are able to get these results with very few latent variables from a linear method, it implies that the spectra lie on a linear underlying manifold.
The model calibration results for polymer from IHM in Meyer-Kirschner et al.~\cite{julian16} are $RMSECV = 0.084 \si{\wtpercent}$ and $R^2 = 0.9835$. 
However, the comparison of our approach with IHM is not straightforward, as in IHM the whole dataset is used for model calibration with Leave-One-Out cross-validation, and only the spectra of a certain range is considered for the study.
Qualitatively speaking, the fit achieved by the proposed method is not quite as good as the IHM method used in Meyer-Kirschner et al.~\cite{julian16} for the model calibration.

\subsubsection{Case study 2: citric acid dissociation data}\label{sep_predictions_case_2}
\begin{figure}[p] 
    \centering
    \begin{subfigure}{0.44\textwidth}
    \includegraphics[width=1\linewidth]{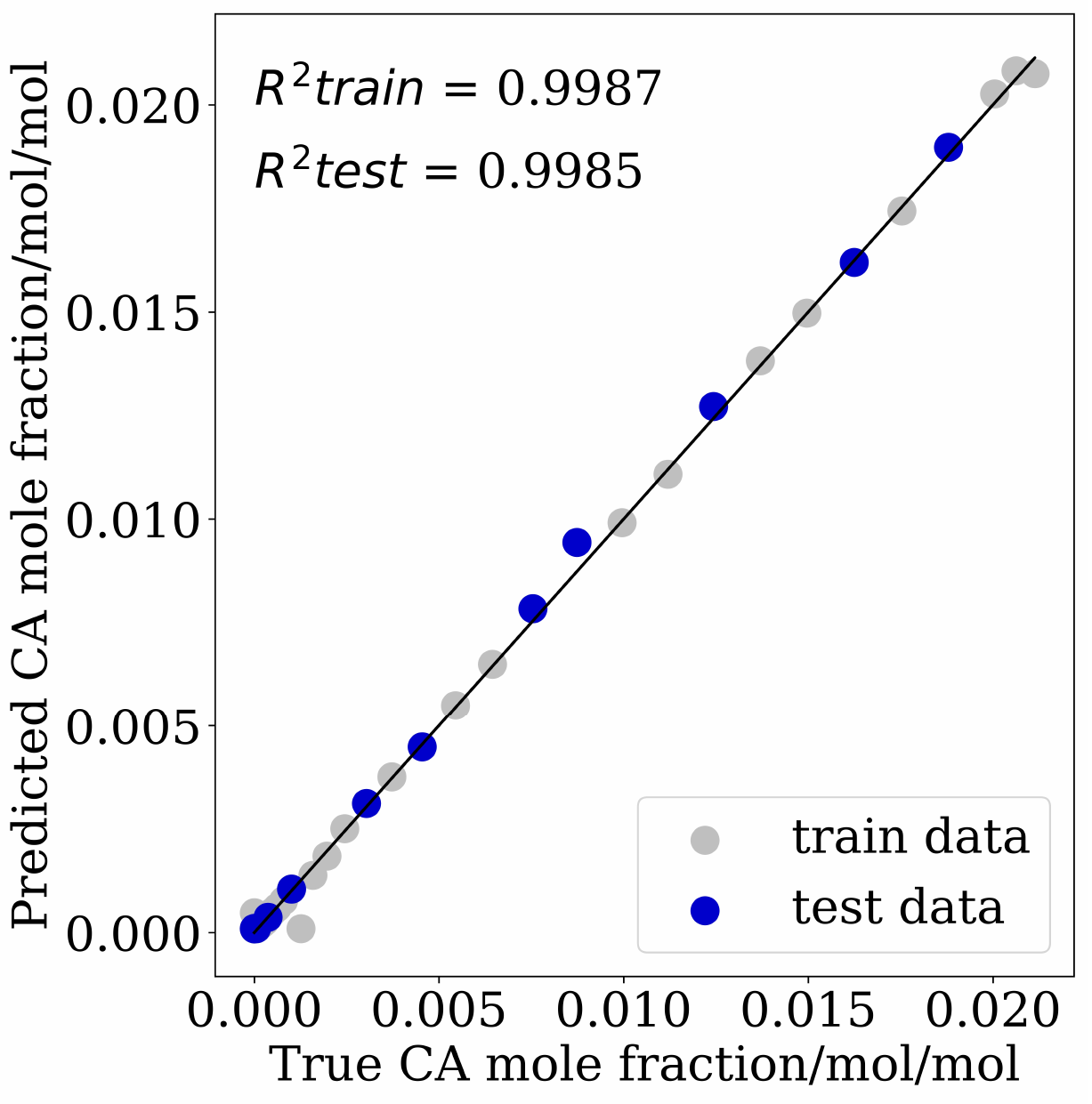}
    \label{fig:sep_Y-AE_case 2}
    \caption{}
    \end{subfigure}
    \hspace{0.1\textwidth}
    \vspace{0.05\textwidth}
    \begin{subfigure}{0.44\textwidth}
    \includegraphics[width=1\linewidth]{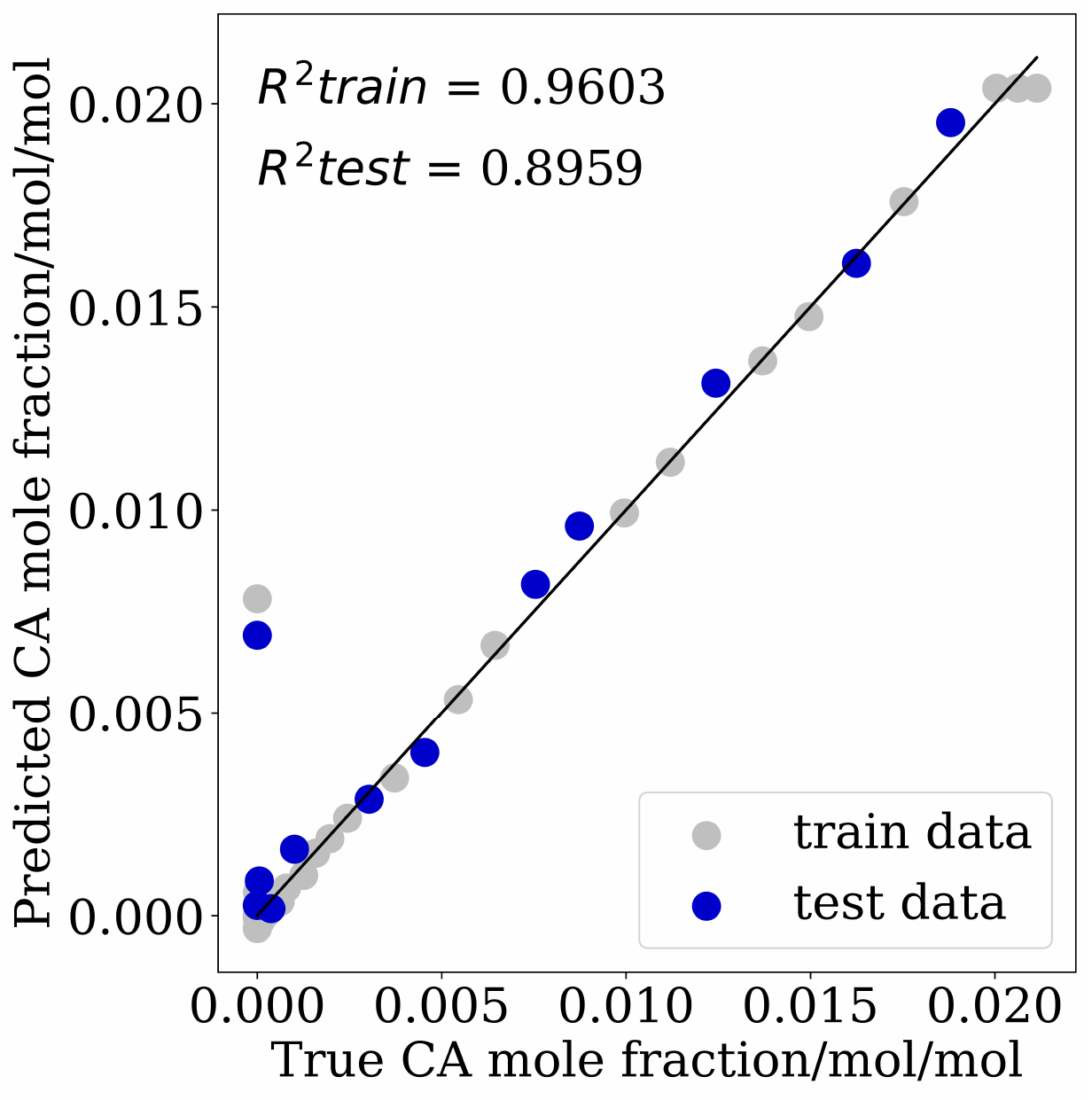}
    \label{fig:ANN_case 2}
    \caption{}
    \end{subfigure}
    \begin{subfigure}{0.44\textwidth}
    \includegraphics[width=1\linewidth]{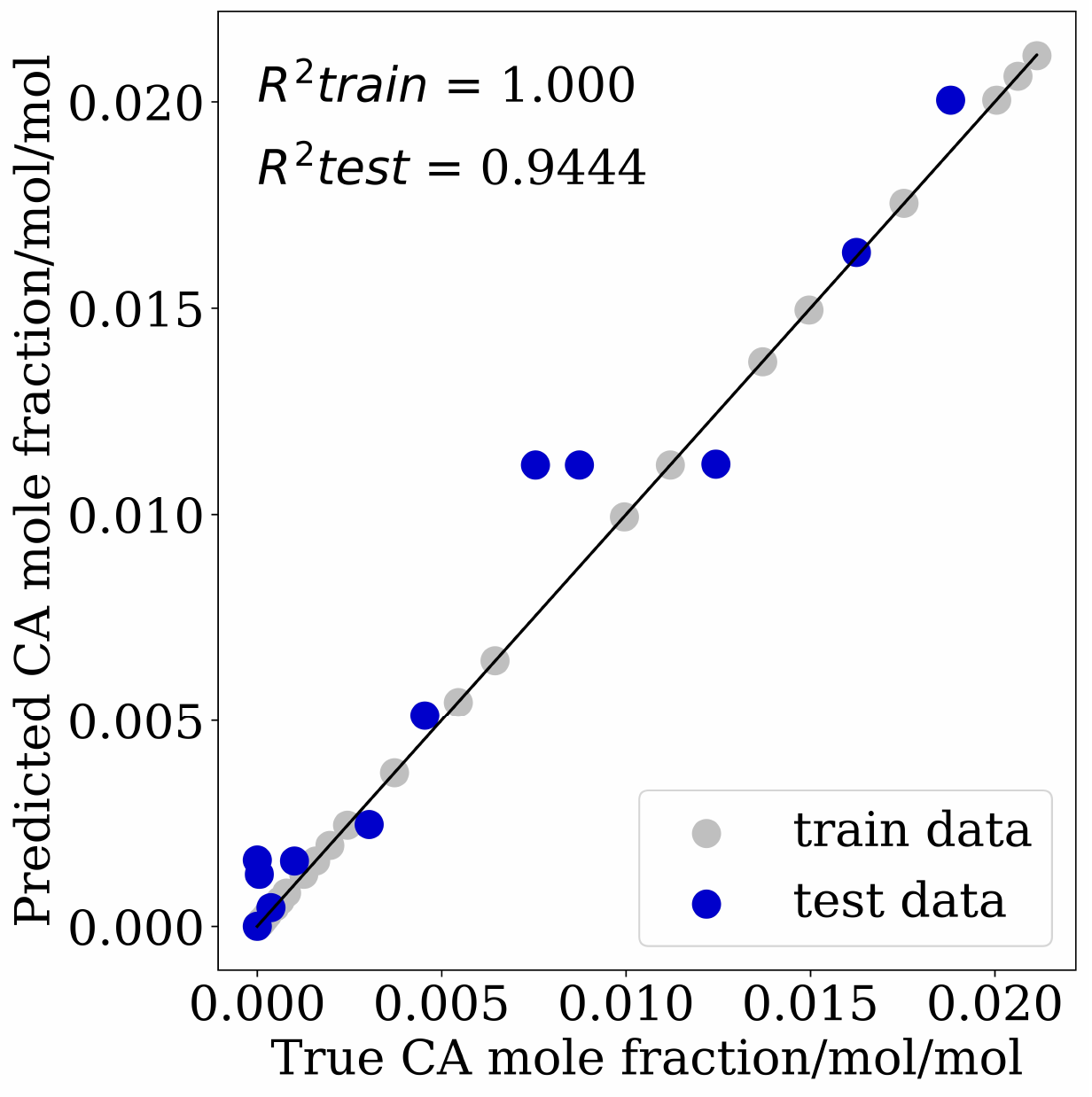}
    \label{fig:XGBOOST_case 2}
    \caption{}
    \end{subfigure}
    \hspace{0.1\textwidth}
    \vspace{0.05\textwidth}
    \begin{subfigure}{0.44\textwidth}
    \includegraphics[width=1\linewidth]{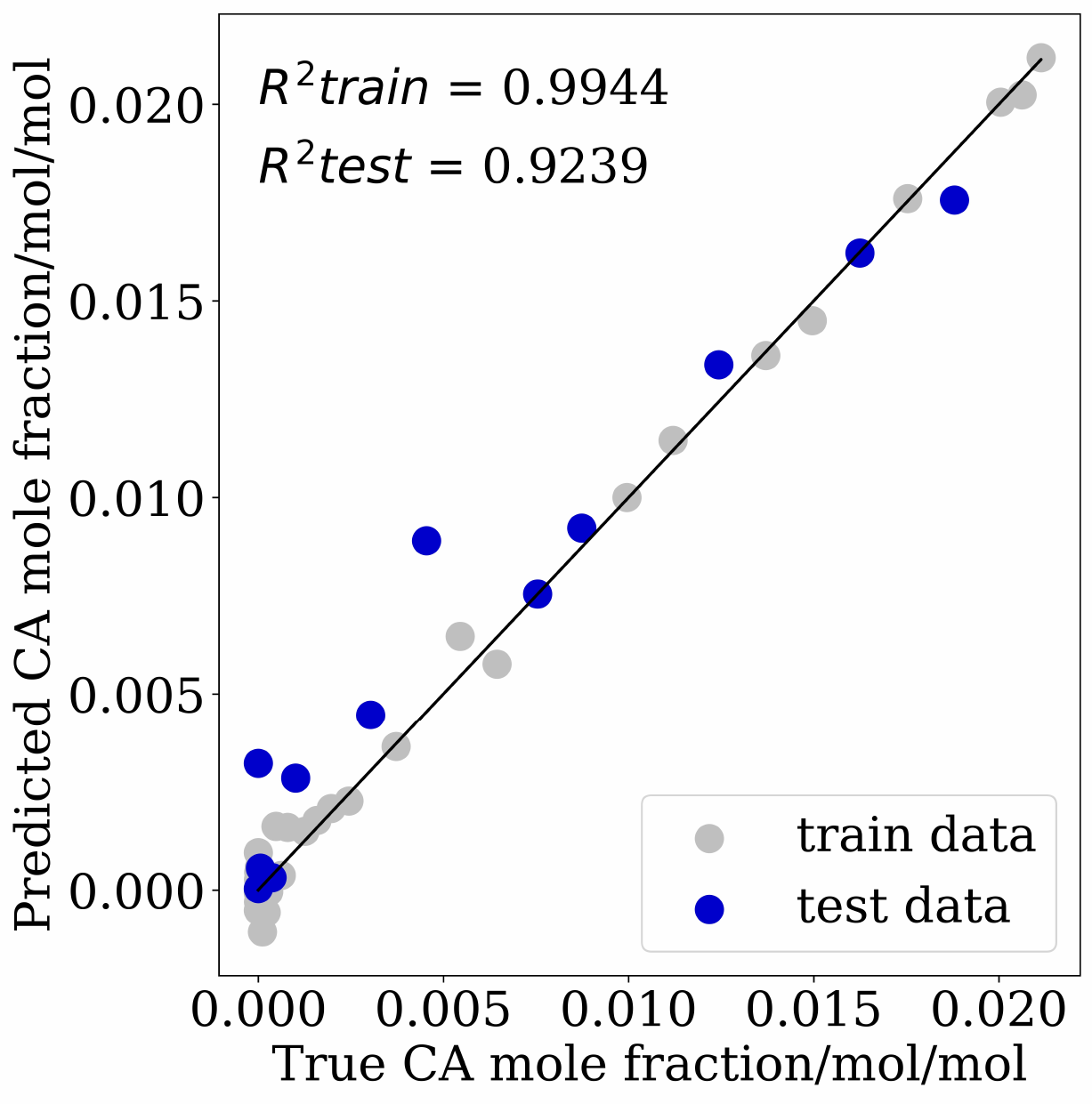}
    \label{fig:PLS_case 2}
    \caption{}
    \end{subfigure}
    \caption{Concentration predictions of citric acid with Y-AE (a), ANN (b),XGB (c), and PLS. The gray dots represent the training datapoints, while the colored ones represent the testing datapoints. Y-AE gives the highest prediction results, while the predictions from PLS and XGB are on the same range.}
    \label{fig:predictions_sep_case_2}
\end{figure}

For this case study we follow the same workflow as for the first case study, described in the section above.
Here, the inputs are the DMAP coordinates representing the Raman spectra, and the outputs are the concentration values of citric acid and its dissociated acid species in water, CA\(^{-}\), CA\(^{2-}\), CA\(^{3-}\).
The parity plots with the predictions from each method for citric acid are shown in Fig.~\ref{fig:predictions_sep_case_2}. The plots for the remaining targets are shown in Fig.~\ref{fig:predictions_sep_CA-_case_2} (Supporting Information), Fig.~\ref{fig:predictions_sep_CA2-_case_2} (Supporting Information), Fig.~\ref{fig:predictions_sep_CA3-_case_2} (Supporting Information) and Fig.~\ref{fig:predictions_sep_water_case_2} (Supporting Information).

From Fig.~\ref{fig:predictions_sep_case_2} we see that Y-shaped autoencoder achieves higher prediction performance than the rest of the methods. It outperforms PLS, which predictions are on the same range as XGB.
The model calibration results for citric acid from IHM in Echtermeyer et al.~\cite{Echtermeyer20} are $RMSECV = 0.0003 \si{mol \per mol}$ and $R^2 = 0.9981$.
The comparison with IHM remains challenging, since again in IHM the whole dataset is used for the model calibration with Leave-$10\%$-Out cross-validation, and only the spectra of a certain range is considered for the study. 
Qualitatively speaking, the fit achieved by the proposed method is not quite as good as the IHM method used in Echtermeyer et al.~\cite{Echtermeyer20} for the model calibration.

\subsection{Simultaneous concentration prediction results}\label{simultaneous predictions} 
In the following, Section~\ref{sim_predictions_case_1} and Section~\ref{sim_predictions_case_2} present the simultaneous concentration prediction results with Y-AE, for the first and the second case study respectively.  We again do an uncertainty quantification study on the multi-target predictions from Y-AE, by training ten models, and the results are shown in Fig.~\ref{fig:UQ_simultaneous Y-AE_case 1} (Supporting Information) and Fig.~\ref{fig:UQ_simultaneous Y-AE_case 2} (Supporting Information) respectively.

\subsubsection{Case study 1: microgel synthesis data}\label{sim_predictions_case_1}
After predicting each of the targets separately with the regression methods, we now predict them simultaneously from the same Y-AE model. The inputs are again the DMAP coordinates representing the Raman spectra, and the outputs are the monomer, polymer and water concentration values. 
When training the Y-AE we study three different approaches, regarding the number of latent variables we consider as input to the estimator. First, we define the targets to be functions of three latent variables, then of two and at the end of one latent variable.
The prediction results for each approach are summarized in Table~\ref{tab:simultaneous_case 1}. 

\begin{table}[ht] 
    \centering
    \makebox[\linewidth]{
        \begin{tabular}{|c|c|c|c|c|c|} 
        \hline
            \multicolumn{3}{|c|}{} & \multicolumn{3}{c|}{Approach}\\
            \cline{4-6}
            \multicolumn{3}{|c|}{Target} & $z_1, z_2, z_3 \mapsto \hat{y}_1, \hat{y}_2, \hat{y}_3$ & $z_1, z_2 \mapsto \hat{y}_1, \hat{y}_2, \hat{y}_3$ & $z_1 \mapsto \hat{y}_1, \hat{y}_2, \hat{y}_3$\\
        \hline
            \multirow{6}{*}{$\rm{H_2O}$} & \multirow{3}{*}{Train} & $\rm{R^2}$ & \cellcolor{lightgray} 0.9998 &  0.9979 & 0.8486 \\
            \cline{3-6}
            & & RMSE & 0.0001 & 0.0003 & 0.0026 \\
            \cline{3-6}
            & & MAE & 0.0001 & 0.0002 & 0.0016 \\
            \cline{2-6}
            & \multirow{3}{*}{Test}& $\rm{R^2}$ & 0.9843 & \cellcolor{LimeGreen} 0.9914 & 0.7097 \\
            \cline{3-6}
            & & RMSE & 0.0009 & 0.0007 & 0.0039 \\
            \cline{3-6}
            & & MAE & 0.0006 & 0.0005 & 0.0025 \\
        \hline
            \multirow{6}{*}{VCL} & \multirow{3}{*}{Train} & $\rm{R^2}$ & \cellcolor{lightgray} 0.9986 & 0.9886 & 0.8841 \\
             \cline{3-6}
            & & RMSE & 0.0002 & 0.0005 & 0.0016 \\
            \cline{3-6}
            & & MAE & 0.0001 & 0.0004 & 0.0013 \\
            \cline{2-6}
            & \multirow{3}{*}{Test}& $\rm{R^2}$ & \cellcolor{LimeGreen} 0.9575 & 0.9544 & 0.3145\\
             \cline{3-6}
            & & RMSE & 0.0006 & 0.0006 & 0.0023 \\
            \cline{3-6}
            & & MAE & 0.0005 & 0.0005 & 0.0019 \\
        \hline
            \multirow{6}{*}{PVCL} & \multirow{3}{*}{Train} & $\rm{R^2}$ & \cellcolor{lightgray} 0.9990 & 0.9840 & 0.6386 \\
            \cline{3-6}
            & & RMSE & 0.0002 & 0.0007 & 0.0032 \\
            \cline{3-6}
            & & MAE & 0.0001 & 0.0004 & 0.0023 \\
            \cline{2-6}
            & \multirow{3}{*}{Test}& $\rm{R^2}$ & 0.9363 & \cellcolor{LimeGreen} 0.9716 & 0.0373 \\
            \cline{3-6}
            & & RMSE & 0.0014 & 0.0009 & 0.0053 \\
            \cline{3-6}
            & & MAE & 0.0011 & 0.0007 & 0.0037 \\
        \hline
        \end{tabular}
    }
    \caption{Prediction performance of the simultaneous target predictions with Y-AE for the first case study. We consider three different approaches depending on the number of latent variables considered for the predictions. The gray shade indicates the highest $R^2$ value achieved on training and the green shade on testing for each target.
    The results from the first and the second approach are on the same range.}
    \label{tab:simultaneous_case 1}
\end{table}

The gray shade indicates the highest $R^2$ value achieved on training and the green shade on testing for each target for all the approaches. We see that the predictions from the first and the second approach are similar. Compared to the single-target predictions from Y-AE in Section~\ref{sep_predictions_case_1}, the multi-target prediction results are in the same range, but slightly lower. Thus, the simultaneous predictions do not bring any improvement to the results, for this case study.
We predict the targets simultaneously also with PLS, with seven latent variables. The inputs are the spectra measurements and the outputs are species concentrations. The multi-target predictions from PLS have lower accuracy than the single-target predictions from PLS, but higher accuracy than the multi-target predictions from Y-AE. The results are summarized in Table~\ref{tab:summary_case 1}, on Section~\ref{summary}.

\subsubsection{Case study 2: citric acid dissociation data}\label{sim_predictions_case_2}
For this case study we follow the same workflow for the simultaneous target prediction as for the fist case study. The inputs here are again the DMAP coordinates representing the Raman spectra and the outputs are the the concentration values of citric acid and its dissociated acid species in water, CA\(^{-}\), CA\(^{2-}\), CA\(^{3-}\).
Given that the low dimensional representation of the Raman spectra consists of two dimensions, we are not not able to determine five targets simultaneously from only two latent variables with the Y-AE. Therefore we define four more parametrizations of the manifold from the initial two-dimensional parametrization, consisting of four, five and six dimensions. We select the representations based on how precisely the original spectrum can be reconstructed from those coordinates. For every parametrization we allow for lower accuracy, considering that the initial two-dimensional representation gave the finest spectra reconstruction.

When training the Y-AE we study two different approaches, regarding the number of latent variables we consider as input to the estimator, depending on the parametrization of the manifold. We define the targets to be functions of four latent variables (for the four-dimensional representation) and then of five latent variables (for the five and six-dimensional representation). We consider either four or five latent variables based on the mass balance. From the predictions we see that one of the five dimensional representations gives the highest prediction results, using five latent variables to predict five targets. However, those results are worse than the single-target predictions from the two-dimensional representation, on Section~\ref{sep_predictions_case_2}. We also predict the targets separately from the five-dimensional parametrization, but the results are again worse than the ones from the two-dimensional one. This could be due to the fact that the five-dimensional parametrization has lower reconstruction accuracy of the spectra, compared to the initial two-dimensional representation.
We predict the targets simultaneously also with PLS, with eight latent variables. The inputs are the spectra measurements and the outputs are species concentrations. The multi-target predictions from PLS have slightly higher accuracy than the single-target predictions from PLS and the multi-target predictions from Y-AE.
The multi-target prediction results from Y-AE and from PLS are summarized in Table~\ref{tab:summary_case 2}, on Section~\ref{summary}.

\subsection{Summary of the results for both case studies}\label{summary}
In this section we summarize the prediction results from all the methods for both case studies (Table~\ref{tab:summary_case 1} for the first case study and Table~\ref{tab:summary_case 2} for the second case study). The gray shade indicates the highest $R^2$ value achieved on training and the green shade on testing for each target for all the approaches.

\begin{table}[p] 
    \centering
    \makebox[\linewidth]{
        \begin{tabular}{|c|c|c|c|c|c|c|c|c|} 
        \hline
            \multicolumn{3}{|c|}{} & \multicolumn{6}{c|}{Method}\\
            \cline{4-9}
            \multicolumn{3}{|c|}{Target} & \multicolumn{4}{c|}{Single-target} & \multicolumn{2}{|c|}{Multi-target}\\
            \cline{4-9}
            \multicolumn{3}{|c|}{} &  Y-AE & ANN & XGB & PLS & Y-AE & PLS\\
        \hline
            \multirow{6}{*}{$\rm{H_2O}$} & \multirow{3}{*}{Train} & $\rm{R^2}$ &  0.9991 & 0.9957 & \cellcolor{lightgray} 0.9999 & 0.9972 & 0.9979 & 0.9949 \\
            \cline{3-9}
            & & RMSE & 0.0002 & 0.0004 & 0.0000 & 0.0003 & 0.0003 & 0.0005 \\
            \cline{3-9}
            & & MAE & 0.0002 & 0.0002 & 0.0000 & 0.0003  & 0.0002 & 0.0004 \\
            \cline{2-9}
            & \multirow{3}{*}{Test}& $\rm{R^2}$ & \cellcolor{LimeGreen} 0.9956 & 0.9819 & 0.8699 & 0.9933 & 0.9914 & 0.9923 \\
             \cline{3-9}
            & & RMSE & 0.0005 & 0.0010 & 0.0026 & 0.0006 & 0.0007 & 0.0006\\
            \cline{3-9}
            & & MAE & 0.0004 & 0.0006 & 0.0023 & 0.0005  & 0.0005 & 0.0005 \\
        \hline
            \multirow{6}{*}{VCL} & \multirow{3}{*}{Train} & $\rm{R^2}$ & 0.9885 & 0.9864 & \cellcolor{lightgray} 0.9999 & 0.9976 & 0.9886 & 0.9967\\
             \cline{3-9}
            & & RMSE & 0.0005 & 0.0005 & 0.0000 & 0.0002 & 0.0005 & 0.0003 \\
            \cline{3-9}
            & & MAE & 0.0003 & 0.0002 & 0.0000 & 0.0002  & 0.0004 & 0.0002 \\
            \cline{2-9}
            & \multirow{3}{*}{Test}& $\rm{R^2}$ & 0.9346 & 0.9187 & 0.5969 & \cellcolor{LimeGreen} 0.9907 & 0.9544 & 0.9843 \\
             \cline{3-9}
            & & RMSE & 0.0007 & 0.0008 & 0.0018 & 0.0003 & 0.0006 & 0.0003 \\
            \cline{3-9}
            & & MAE & 0.0006 & 0.0007 & 0.0010 & 0.0002  & 0.0005 & 0.0003 \\
        \hline
            \multirow{6}{*}{PVCL} & \multirow{3}{*}{Train} & $\rm{R^2}$ & 0.9948 & 0.9999 & \cellcolor{lightgray} 0.9999 & 0.9980 & 0.9840 & 0.9937 \\
            \cline{3-9}
            & & RMSE & 0.0004 & 0.0000 & 0.0000 & 0.0002 & 0.0007 & 0.0004\\
            \cline{3-9}
            & & MAE & 0.0003 & 0.0000 & 0.0000 & 0.0002  & 0.0004 & 0.0003 \\
            \cline{2-9}
            & \multirow{3}{*}{Test}& $\rm{R^2}$ & 0.9867 & 0.9670 & 0.7486 & \cellcolor{LimeGreen} 0.9895 & 0.9716 & 0.9875\\
             \cline{3-9}
            & & RMSE & 0.0006 & 0.0010 & 0.0027 & 0.0006 & 0.0009 & 0.0006\\
            \cline{3-9}
            & & MAE & 0.0005 & 0.0007 & 0.0017 & 0.0005  & 0.0007 & 0.0005 \\
        \hline
        \end{tabular}
    }
    \caption{Summary of the prediction results for the first case study. Gray shade indicates the highest $R^2$ value achieved on training and green shade on testing for each target. The single-target predictions from PLS have higher accuracy than the predictions from the remaining methods.}
    \label{tab:summary_case 1}
\end{table}

For the multi-target predictions for the first case study, we see the results from the second approach, predicting three targets from two latent variables. We see that XGB achieve the highest training prediction performance for all the targets (with slight difference from Y-AE, ANN and PLS), Y-AE achieve the highest testing prediction performance for water, and PLS achieves the highest testing prediction performance for monomer and polymer. We see that for this case study, the simultaneous target prediction results from Y-AE are in general in the same range with the single-target predictions from Y-AE, but slightly lower. While the multi-target predictions from PLS, are slightly lower than the single-target predictions from PLS, but higher than the multi-target predictions from Y-AE.

\begin{table}[p] 
    \centering
    \makebox[\linewidth]{
        \begin{tabular}{|c|c|c|c|c|c|c|c|c|} 
        \hline
            \multicolumn{3}{|c|}{} & \multicolumn{6}{c|}{Method}\\
            \cline{4-9}
            \multicolumn{3}{|c|}{Target} & \multicolumn{4}{c|}{Single-target} & \multicolumn{2}{|c|}{Multi-target}\\
            \cline{4-9}
            \multicolumn{3}{|c|}{} &  Y-AE & ANN & XGB & PLS & Y-AE & PLS\\
        \hline
            \multirow{6}{*}{CA} & \multirow{3}{*}{Train} & $\rm{R^2}$ & 0.9987 & 0.9603 & \cellcolor{lightgray} 0.9999 & 0.9944 & 0.9808 & 0.9882  \\
             \cline{3-9}
            & & RMSE & 0.0002 & 0.0014 & 0.0000 & 0.0005  & 0.0010 & 0.0008 \\
            \cline{3-9}
            & & MAE & 0.0001 & 0.0004 & 0.0000 & 0.0004  & 0.0006 & 0.0006 \\
            \cline{2-9}
            & \multirow{3}{*}{Test}& $\rm{R^2}$ & \cellcolor{LimeGreen} 0.9985 & 0.8959 & 0.9444 & 0.9239 & 0.7848 & 0.9272 \\
             \cline{3-9}
            & & RMSE & 0.0002 & 0.0021 & 0.0015 & 0.0018  & 0.0030 & 0.0017 \\
            \cline{3-9}
            & & MAE & 0.0002 & 0.0010 & 0.0011 & 0.0012  & 0.0016 & 0.00013 \\
        \hline
            \multirow{6}{*}{CA\(^{-}\)} & \multirow{3}{*}{Train} & $\rm{R^2}$ & 0.9590 & 0.9907 & \cellcolor{lightgray} 0.9999 & 0.9926  & 0.9594 & 0.9565 \\
             \cline{3-9}
            & & RMSE & 0.0008 & 0.0004 & 0.0000 & 0.0003  & 0.0008 & 0.0008 \\
            \cline{3-9}
            & & MAE & 0.0004 & 0.0001 & 0.0000 & 0.0002  & 0.0004 & 0.0007 \\
            \cline{2-9}
            & \multirow{3}{*}{Test}& $\rm{R^2}$ & \cellcolor{LimeGreen} 0.9585 & 0.9082 & 0.7240 & 0.9057 & 0.9553 & 0.8264 \\
             \cline{3-9}
            & & RMSE & 0.0008 & 0.0012 & 0.0020 & 0.0012  & 0.0008 & 0.0016 \\
            \cline{3-9}
            & & MAE & 0.0004 & 0.0005 & 0.0012 & 0.0010  & 0.0005 & 0.0013 \\
        \hline
            \multirow{6}{*}{CA\(^{2-}\)} & \multirow{3}{*}{Train} & $\rm{R^2}$ & 0.9951 & 0.9629 & \cellcolor{lightgray} 0.9999 & 0.9997  & 0.9087 & 0.9730 \\
                \cline{3-9}
            & & RMSE & 0.0002 & 0.0006 & 0.0000 & 0.0001  & 0.0009 & 0.0005 \\
            \cline{3-9}
            & & MAE & 0.0002 & 0.0004 & 0.0000 & 0.0000  & 0.0004 & 0.0004 \\
            \cline{2-9}
            & \multirow{3}{*}{Test}& $\rm{R^2}$ & \cellcolor{LimeGreen} 0.9621 & 0.9485 & 0.6142 & 0.7194 & 0.9054 & 0.7300 \\
                \cline{3-9}
            & & RMSE & 0.0005 & 0.0006 & 0.0017 & 0.0015  & 0.0009 & 0.0014 \\
            \cline{3-9}
            & & MAE & 0.0003 & 0.0004 & 0.0009 & 0.0011  & 0.0005 & 0.0010 \\
        \hline
            \multirow{6}{*}{CA\(^{3-}\)} & \multirow{3}{*}{Train} & $\rm{R^2}$ & 0.9984 & 0.9297 & \cellcolor{lightgray} 0.9999 & 0.9639  & 0.8241 & 0.8348 \\
                \cline{3-9}
             & & RMSE & 0.0001 & 0.0004 & 0.0000 & 0.0003  & 0.0006 & 0.0006 \\
             \cline{3-9}
            & & MAE & 0.0000 & 0.0000 & 0.0000 & 0.0002  & 0.0003 & 0.0004 \\
            \cline{2-9}
            & \multirow{3}{*}{Test}& $\rm{R^2}$ & \cellcolor{LimeGreen} 0.9642 & 0.3637 & 0.8979 & 0.5192  & 0.5227 & 0.5353 \\
                \cline{3-9}
            & & RMSE & 0.0005 & 0.0022 & 0.0009 & 0.0019  & 0.0019 & 0.0019 \\
            \cline{3-9}
            & & MAE & 0.0002 & 0.0007 & 0.0003 & 0.0010  & 0.0008 & 0.0010 \\
        \hline
            \multirow{6}{*}{$\rm{H_2O}$} & \multirow{3}{*}{Train} & $\rm{R^2}$ & 0.9993 & 0.9997 & \cellcolor{lightgray} 0.9999 & 0.9945  & 0.8093 & 0.9925 \\
                \cline{3-9}
            & & RMSE & 0.0001 & 0.0001 & 0.0000 & 0.0003  & 0.0018 & 0.0004 \\
            \cline{3-9}
            & & MAE & 0.0001 & 0.0000 & 0.0000 & 0.0002  & 0.0008 & 0.0003 \\
            \cline{2-9}
            & \multirow{3}{*}{Test}& $\rm{R^2}$ & \cellcolor{LimeGreen} 0.9941 & 0.9749 & 0.8010 & 0.9020  & 0.7883 & 0.9163 \\
                \cline{3-9}
            & & RMSE & 0.0002 & 0.0005 & 0.0014 & 0.0010  & 0.0015 & 0.0009 \\
            \cline{3-9}
            & & MAE & 0.0002 & 0.0003 & 0.0008 & 0.0007  & 0.0008 & 0.0006 \\
        \hline
        \end{tabular}
    }
    \caption{Summary of the results for the second case study. Gray shade indicates the highest $R^2$ value achieved on training and green shade on testing for each target. The single-target predictions from Y-AE have higher accuracy.}
    \label{tab:summary_case 2}
\end{table}

For the multi-target predictions for the second case study, we see the results from the five-dimensional manifold, derived from the initial two-dimensional manifold. We see that XGB achieve the highest training prediction performance for all the targets (with slight difference from Y-AE, ANN and PLS), while Y-AE achieve the highest testing performance for all the targets.
For this case study, the multi-target prediction approach brings no improvement to the predictions, since the predictions for the targets are worse than those of the single-target prediction approach.
This again could come as a result of the lower reconstruction accuracy of the five-dimensional manifold.

\section{Conclusion} \label{conclusions}
In this contribution, we build up on the DMAPs and Y-shaped conformal autoencoder workflow, recently introduced for in-line particle size determination, to predict concentrations. We apply this approach on two case studies considering Raman measurements and concentration measurements.

By implementing DMAPs, we are able to determine the low-dimensional underlying manifold of spectra, reducing the dimensionality of the initial dataset.
Then, we use the reduced representation of the spectra as input to a Y-AE, ANN and XGB model to predict the target concentration values. We also use PLS to predict the targets.
The comparison shows that Y-AE achieve the highest prediction performance for the second case study.
For the first case study, the linear method PLS outperformed, indicating that the spectra lie on a linear low-dimensional manifold.
Comparing our proposed workflow  with IHM is not straightforward: in IHM, the whole dataset is used for the model calibration, with cross-validation, while in our workflow we use 75$\%$ of the dataset for the training and 25$\%$ for the testing. Furthermore, we use the entire spectrum, while in IHM only the fingerprint region is considered.
Nevertheless, the fit achieved by the proposed method is not quite as good as the fit from IHM.

We also predict the targets simultaneously from the same Y-AE model, with the results showing similar trend with the single-target predictions for the first case study and worse trend for the second case study.
These results demonstrate that the DMAPs and Y-AE workflow can also be used for concentration determination and multi-target predictions.

Future work should consider the simultaneous determination of particle size and liquid concentration in polymerization processes, as well as the application of the method to other systems.

\section*{Author contributions}
\textbf{Anna Taka}.: writing - original draft, methodology, implementation, investigation, visualization.
\textbf{Eleni D. Koronaki}: writing - review and editing, methodology, investigation.
\textbf{Ioannis G. Kevrekidis}: writing - review and editing, conceptualization, methodology, investigation.
\textbf{Eike Cramer}: writing - review and editing, methodology, investigation, supervision.
\textbf{Joern Viell}: review and editing.
\textbf{Alexander Mitsos}: writing - review and editing, supervision, conceptualization, methodology, funding acquisition.

\section*{Acknowledgments} \label{acknowledgements}
This work is conducted as part of the CINEMA project. The project (GA number 101072732) has received funding from the HORIZON-MSCA-2021-DN-01 call of the research and innovation programme of Horizon Europe 2021 under the Marie Sklodowska-Curie actions. We also acknowledge the support from the Werner Siemens Foundation in the frame of the WSS Research Center ``catalaix''.
We thank Lennard Schulte and Alexandros Giannakopoulos for the scientific discussions and Daniel Jungen for visualization suggestions.

\section*{Code availability statement} \label{git}
The code is available in the GitLab repository: http://permalink.avt.rwth-aachen.de/?id=894395.

\nomenclature[A]{\(PAT\)}{Process Analytical technologies}
\nomenclature[A]{\(GC\)}{Gas Chromatography}
\nomenclature[A]{\(NMR\)}{Nuclear Magnetic Resonance spectroscopy}
\nomenclature[A]{\(NIR\)}{Near-infrared spectroscopy}
\nomenclature[A]{\(PLS\)}{Partial Least Squares}
\nomenclature[A]{\(PI\)}{Peak Integration}
\nomenclature[A]{\(MCR-ALS\)}{Multivariate Curve Resolution-Alternating Least Squares}
\nomenclature[A]{\(PCA\)}{Principal Component Analysis}
\nomenclature[A]{\(DMAPs\)}{Diffusion Maps}
\nomenclature[A]{\(Y-AE\)}{Y-shaped Conformal Autoencoder}
\nomenclature[A]{\(IHM\)}{Indirect Hard Modeling}
\nomenclature[A]{XGB}{Extreme Gradient Boosting}
\nomenclature[A]{ANNs}{Artificial Neural Networks}
\nomenclature[A]{UQ}{Uncertainty Quantification}
\nomenclature[A]{NaOH}{sodium hydroxide}
\nomenclature[A]{VCL}{N-vinylcaprolactam (monomer)}
\nomenclature[A]{PVCL}{poly-N-vinylcaprolactam (microgels)}
\nomenclature[A]{CA}{citric acid}
\nomenclature[A]{$\rm{CA}^-$}{dihydrogen citrate ion}
\nomenclature[A]{$\rm{CA}^{2-}$}{mono-hydrogen citrate ion}
\nomenclature[A]{$\rm{CA}^{3-}$}{citrate ion}
\nomenclature[A]{$\rm{H_2O}$}{water}
\nomenclature[A]{$MSE$}{Mean-Squared-Error}
\nomenclature[A]{$RMSE$}{Root-Mean-Squared-Error}
\nomenclature[A]{$RMSECV$}{Root-Mean-Squared-Error of Cross-validation}
\nomenclature[A]{$MAE$}{Mean-Absolute-Error}
\nomenclature[A]{$R^2$}{coefficient of determination}
\nomenclature[A]{$SS_{res}$}{Residual Sum of Squares}
\nomenclature[A]{$SS_{tot}$}{Total Sum of Squares}
\nomenclature[S]{$\textbf{x}$}{vector of the input data}
\nomenclature[S]{$\hat{\textbf{x}}$}{vector of the reconstructed inputs from the Y-AE}
\nomenclature[S]{$\textbf{z}$}{vector of the Y-AE latent variables}
\nomenclature[S]{$\textbf{y}$}{vector of the true output values}
\nomenclature[S]{$\hat{\textbf{y}}$}{vector of the predicted output values from the Y-AE}
\nomenclature[S]{$y_i$}{true target value}
\nomenclature[S]{$\hat{y}_i$}{predicted target value}
\nomenclature[S]{$\bar{y}$}{mean of the true target values}
\nomenclature[S]{$n$}{total number of datapoints}
\nomenclature[S]{$\mathcal{L}$}{loss function}
\nomenclature[G]{$\lambda$}{eigenvalues of the scaled affinity matrix from DMAPs}  
\nomenclature[G]{$\alpha$}{hyperparameter to scale the importance of the orthogonality constraint}
\nomenclature[G]{$\psi_i$}{DMAP coordinates (eigenvectors of the affinity matrix)}

\printnomenclature
\putbib 
\end{bibunit}  

\clearpage

\setcounter{section}{0}
\setcounter{subsection}{0}
\setcounter{figure}{0}
\setcounter{page}{1}

\renewcommand{\figurename}{Fig.} 
\renewcommand{\thefigure}{S\arabic{figure}} 

\begin{center}
    \LARGE{Supplementary Information to Determining Concentration from Raman Spectroscopy with Manifold Learning and Y-shaped Conformal Autoencoders}
    \vspace{1em}
    
    \large{Anna Taka\textsuperscript{1},}
    \large{Eleni D. Koronaki\textsuperscript{2},}
    \large{Ioannis G. Kevrekidis\textsuperscript{3},}
    \large{Eike Cramer\textsuperscript{4},}
    \large{Joern Viell\textsuperscript{1}, and}
    \large{Alexander Mitsos\textsuperscript{5,1,6*}}
    \vspace{0.5em}
    
    \large{\textsuperscript{1}Process Systems Engineering (AVT.SVT), RWTH Aachen University, Forckenbeckstr. 51, 52074 Aachen, Germany}
    
    \large{\textsuperscript{2}Luxembourg Institute of Science and Technology (LIST), 5 Av. des Hauts-Fourneaux, 4362 Esch-Belval Esch-sur-Alzette, Luxembourg}
    
    \large{\textsuperscript{3}Department of Chemical and Biomolecular Engineering and Department of Applied Mathematics and Statistics, Whiting School of Engineering, John Hopkins University, 3400 North Charles Street, Baltimore, MD 21218, USA}
    
    \large{\textsuperscript{4}Department of Chemical Engineering, University College London,  Gower St, London WC1E 6BT, United Kingdom}
    
    \large{\textsuperscript{5}JARA-CSD, 52056 Aachen, Germany}
    
    \large{\textsuperscript{6}Institute of Climate and Energy Systems, Energy Systems Engineering (ICE-1), Forschungszentrum Jülich GmbH, 52425 Jülich, Germany}
    
    \large{\textsuperscript{*}Corresponding author: amitsos@alum.mit.edu}
\end{center}
\vspace{1em}

\begin{bibunit} 
\renewcommand{\refname}{Supporting References} 

\section*{Structure}\label{structure}
This document provides supplementary information to the ``Determining Concentration from Raman Spectroscopy with Manifold Learning and Y-shaped Conformal Autoencoders'' manuscript, and it is structured as follows: Section~\ref{separate_remaining_targets} presents the single-target prediction results for the first and the second case study, for the remaining targets not presented in the manuscript.
Section~\ref{SI_UQ} presents the results of the uncertainty quantification study (UQ) on the single-target predictions from the Y-shaped conformal autoencoder (Y-AE) and the artificial neural network (ANN), and the multi-target predictions from the Y-AE, for the first and the second case study.
Finally, Section~\ref{SI_DMAPs study} presents alternative representations of the underlying manifold of the Raman spectra for the first and the second case study found with DMAPs, and the eigenfunction extension on the data performed with the Nyström extension, which uses some functions defined on the dataset, called Geometric Harmonics.

\section{Separate concentration prediction results}\label{separate_remaining_targets}
In the following, Section~\ref{separate_remaining_targets_case_1} and Section~\ref{separate_remaining_targets_case_2} present the concentration prediction results with Y-AE, ANNs, XGB and PLS, for the first and the second case study respectively, for the remaining targets not presented in the manuscript.

\subsection{Case study 1: microgel synthesis data}\label{separate_remaining_targets_case_1}
\begin{figure}[p] 
    \centering
    \begin{subfigure}{0.4\textwidth}
    \includegraphics[width=1\linewidth]{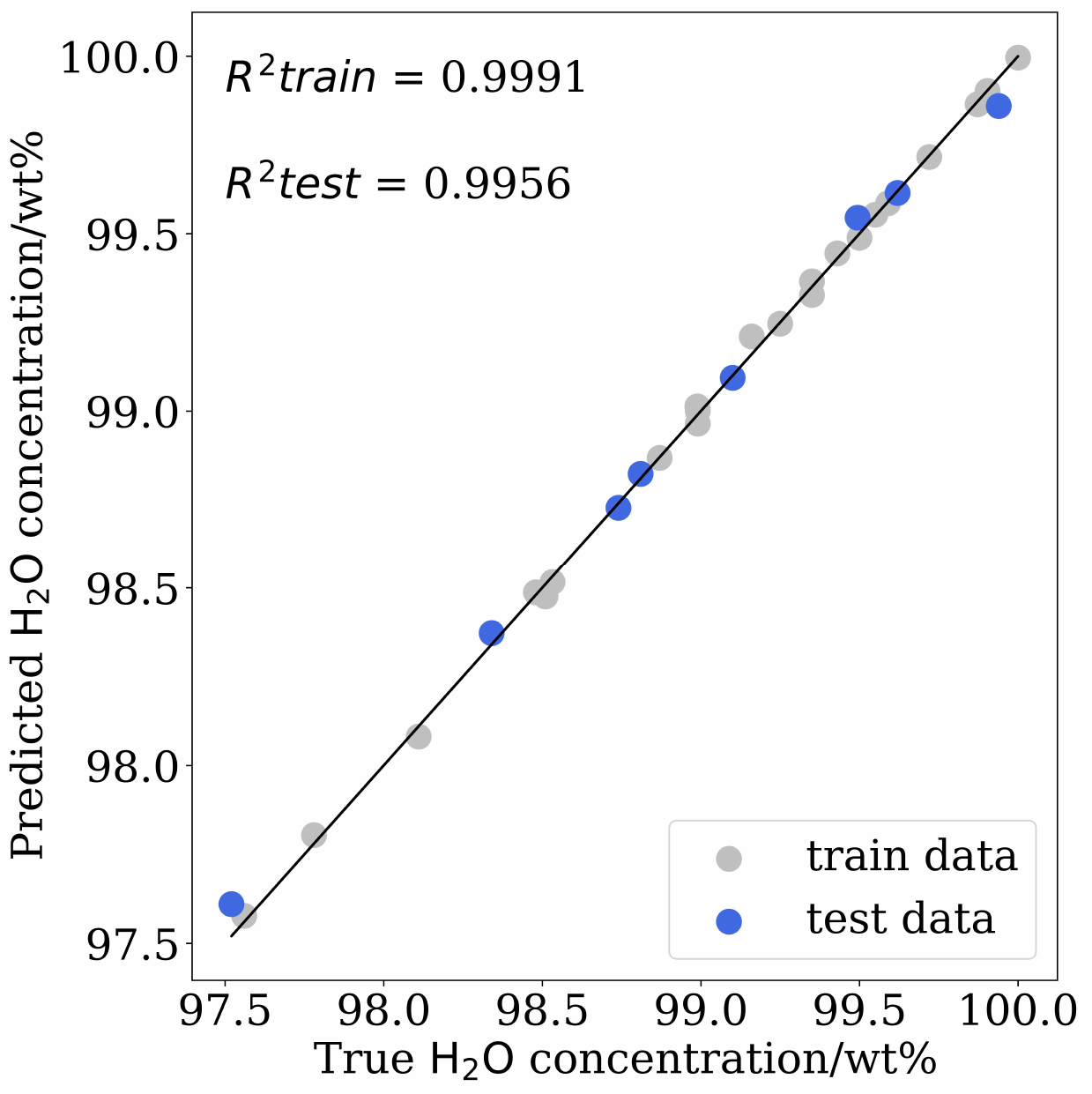}
    \caption{}
    \label{fig:sep_Y-AE_case 1_water}
    \end{subfigure}
    \hspace{0.05\textwidth}
    \vspace{0.05\textwidth}
    \begin{subfigure}{0.4\textwidth}
    \includegraphics[width=1\linewidth]{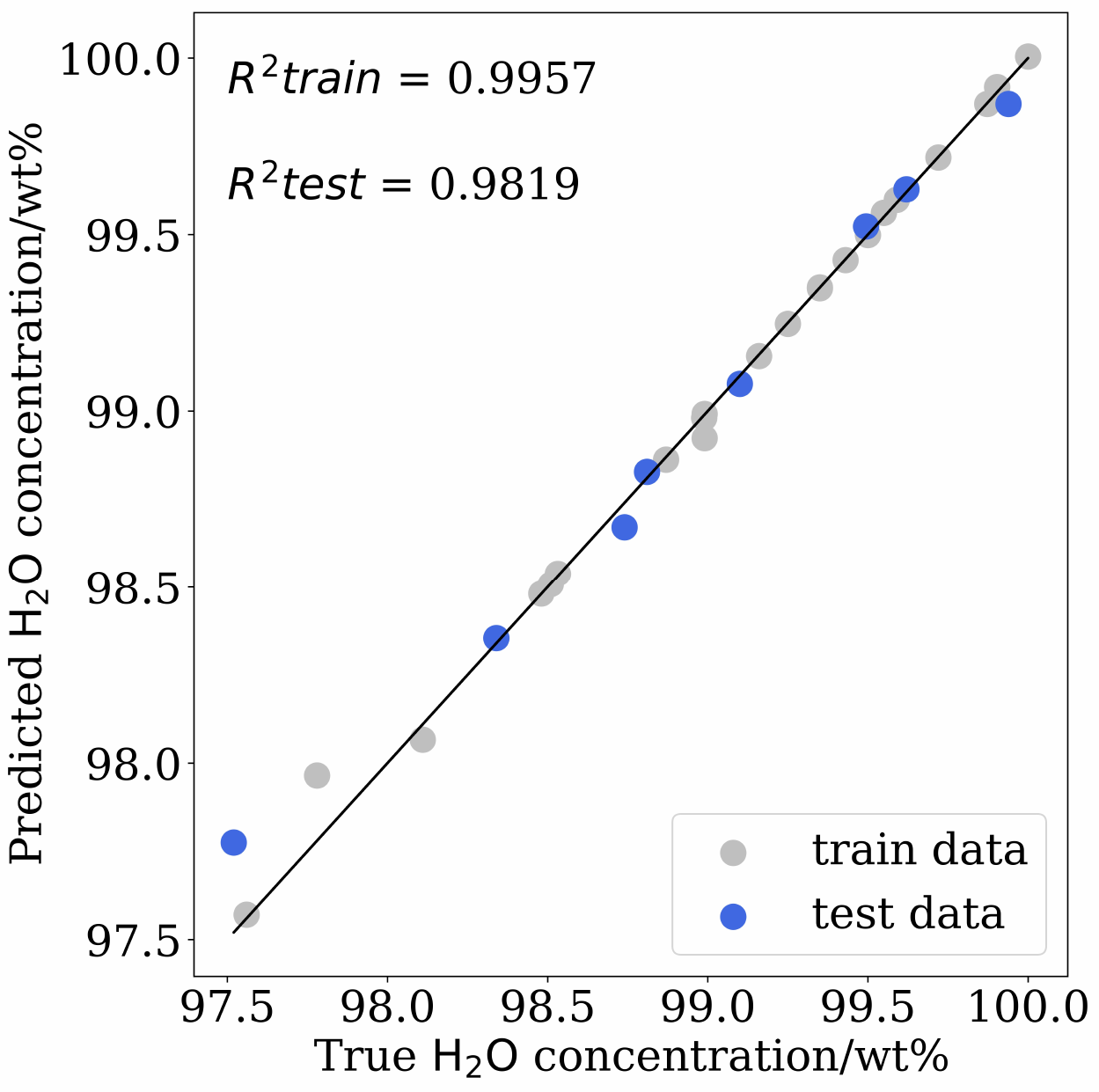}
    \caption{}
    \label{fig:ANN_case 1_water}
    \end{subfigure}
    \begin{subfigure}{0.4\textwidth}
    \includegraphics[width=1\linewidth]{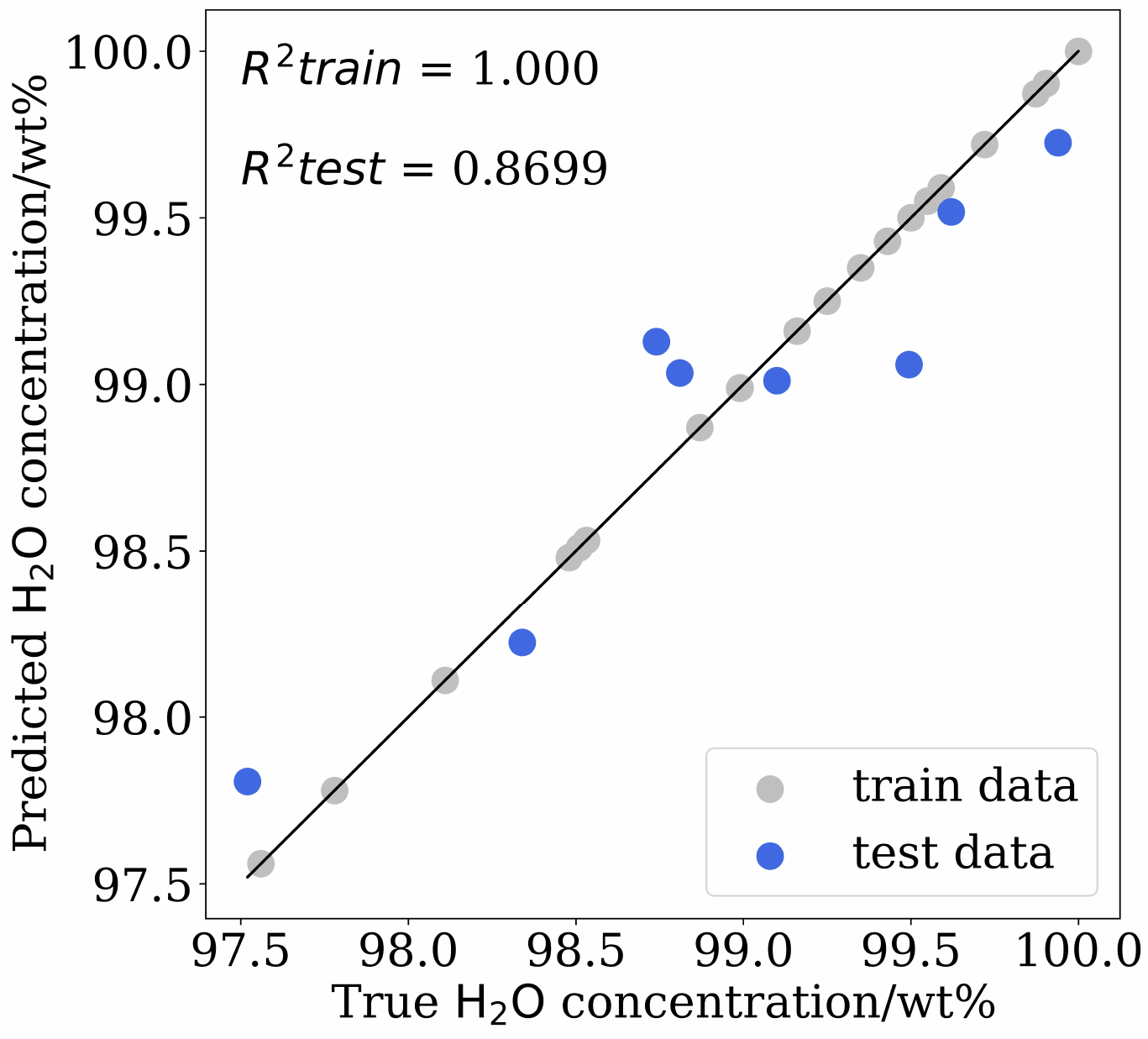}
    \caption{}
    \label{fig:XGBOOST_case 1_water}
    \end{subfigure}
    \hspace{0.05\textwidth}
    \begin{subfigure}{0.4\textwidth}
    \includegraphics[width=1\linewidth]{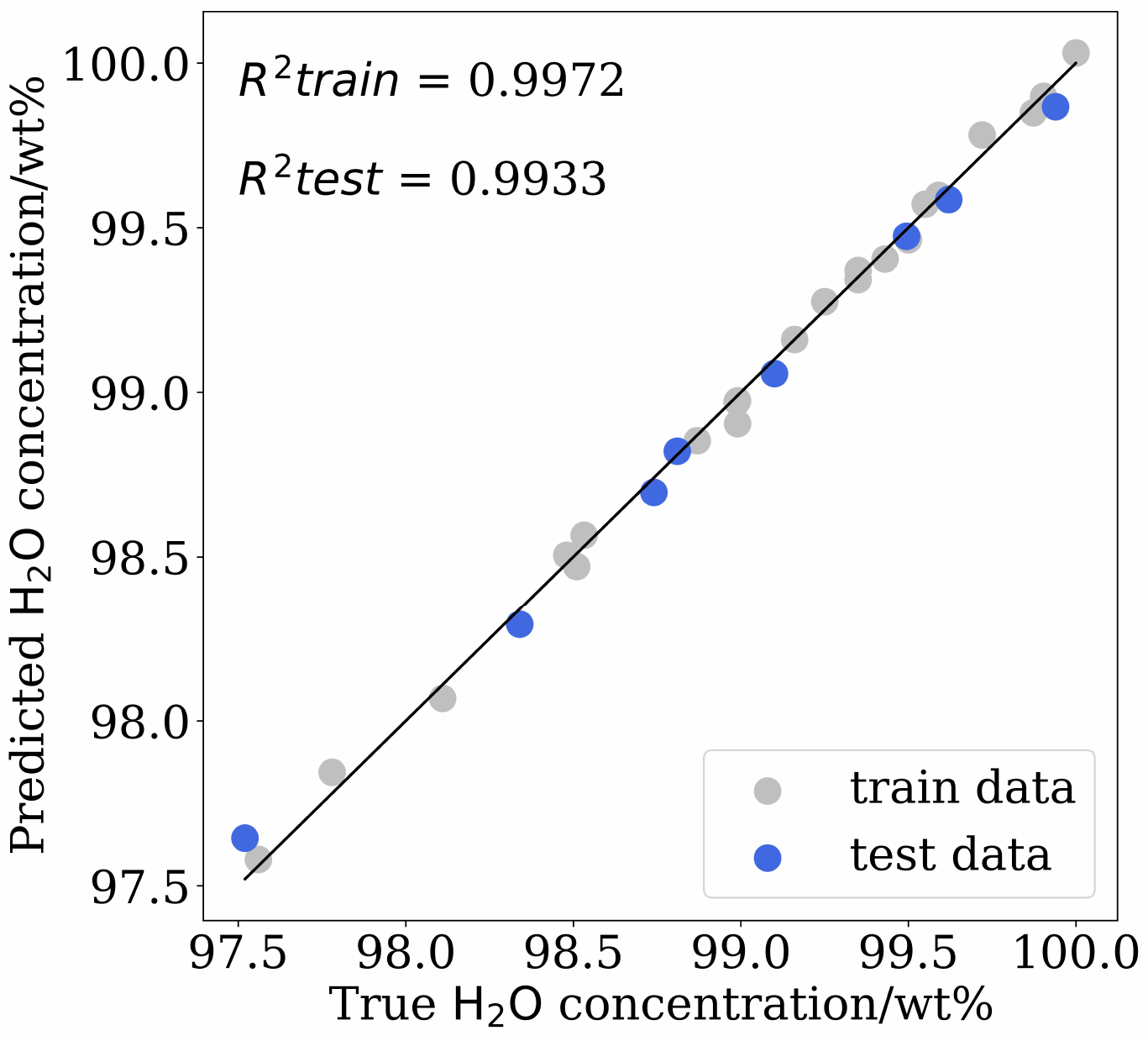}
    \caption{}
    \label{fig:PLS_case 1_water}
    \end{subfigure}
    \caption{Water concentration predictions with Y-AE (a), ANN (b), XGB (c), and PLS (d). The gray dots represent the training datapoints, while the colored ones represent the testing datapoints. PLS gives the highest prediction results with very few latent variables: this indicates that the spectroscopic data lie on a linear low-dimensional manifold. The predictions from Y-AE and ANN are on the same range.}
    \label{fig:predictions_sep_water_case_1}
\end{figure}

\begin{figure}[p] 
    \centering
    \begin{subfigure}{0.4\textwidth}
    \includegraphics[width=1\linewidth]{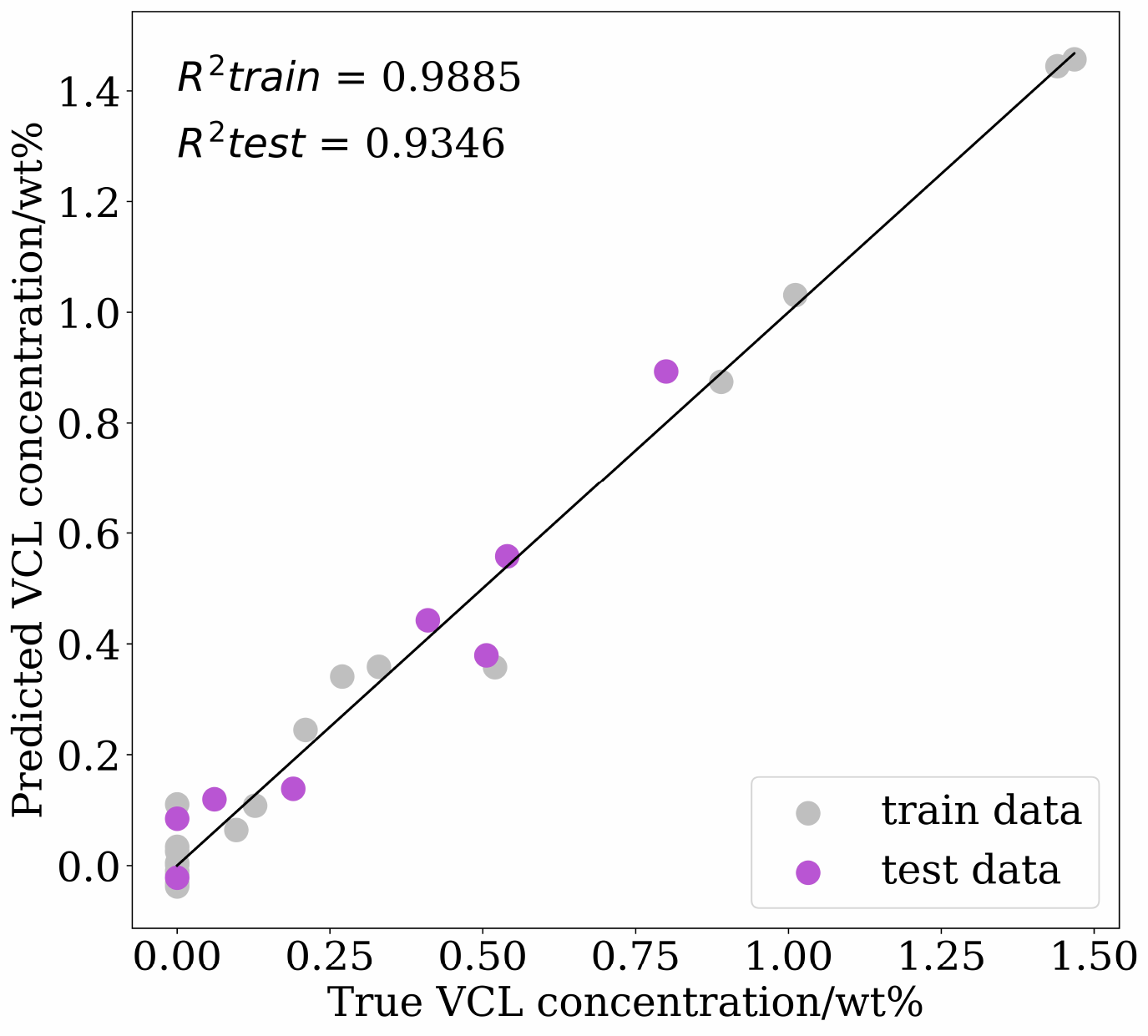}
    \caption{}
    \label{fig:sep_Y-AE_case 1_monomer}
    \end{subfigure}
    \hspace{0.05\textwidth}
    \vspace{0.05\textwidth}
    \begin{subfigure}{0.4\textwidth}
    \includegraphics[width=1\linewidth]{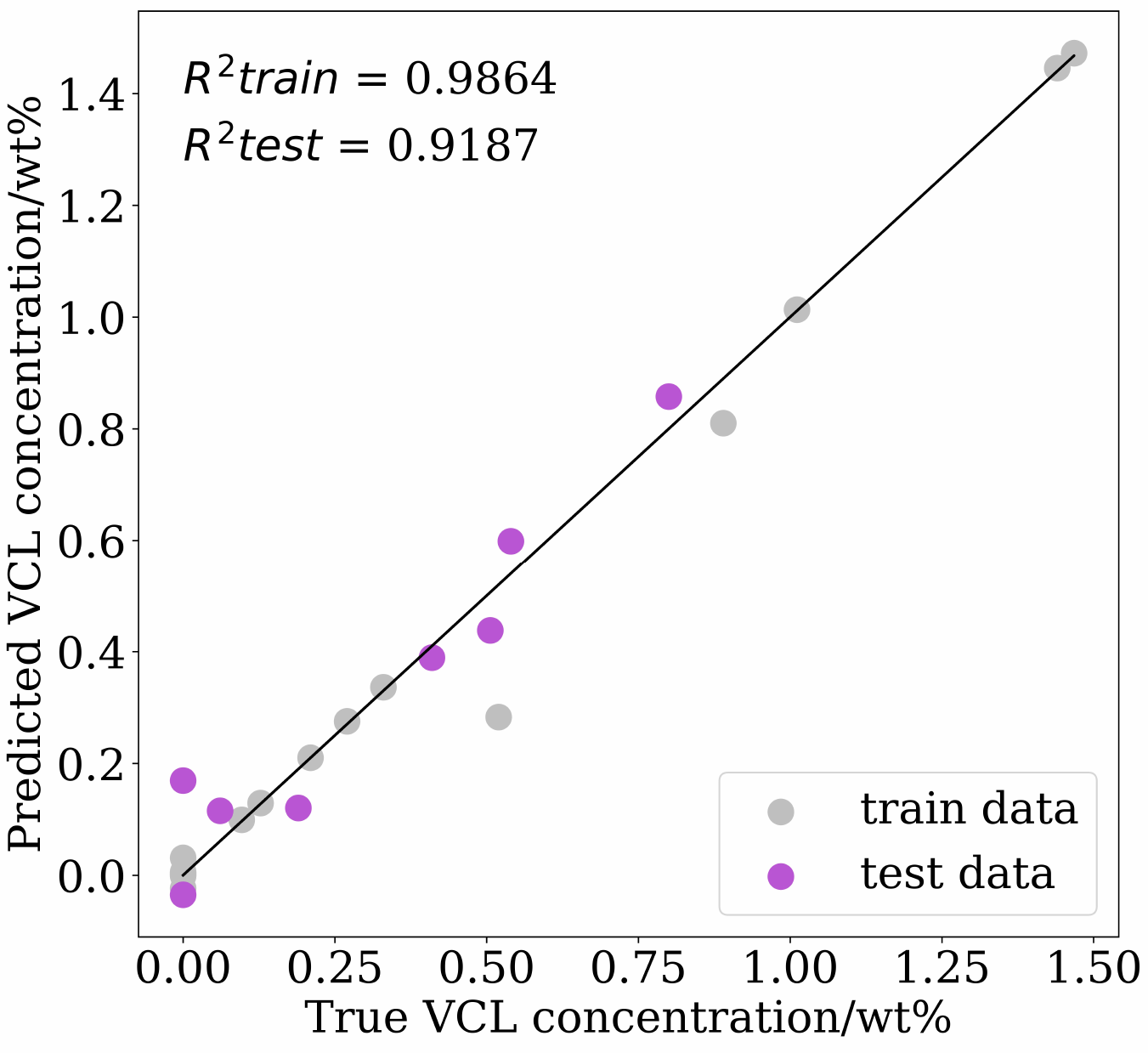}
    \caption{}
    \label{fig:ANN_case 1_monomer}
    \end{subfigure}
    \begin{subfigure}{0.4\textwidth}
    \includegraphics[width=1\linewidth]{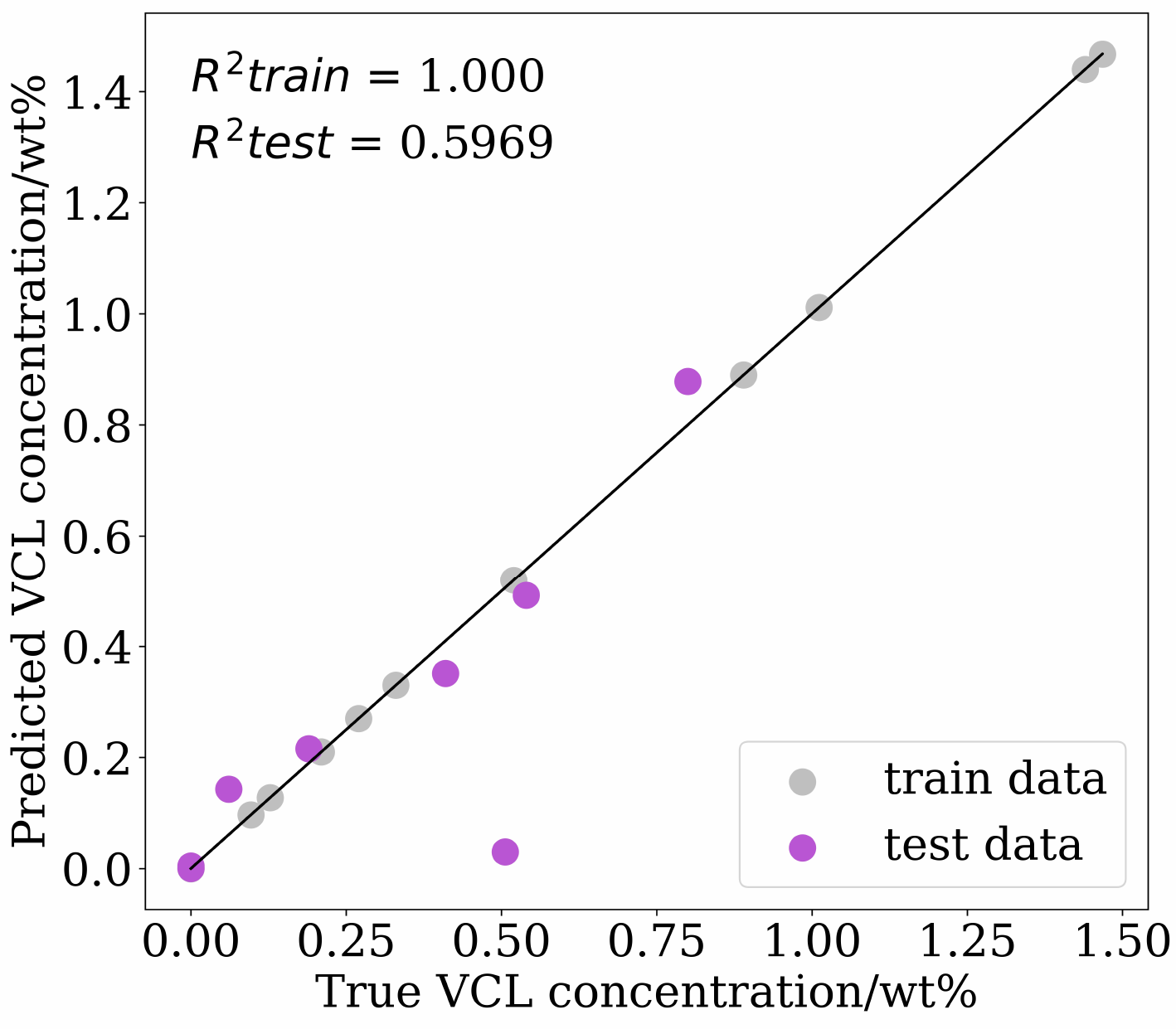}
    \caption{}
    \label{fig:XGBOOST_case 1_monomer}
    \end{subfigure}
    \hspace{0.05\textwidth}
    \begin{subfigure}{0.4\textwidth}
    \includegraphics[width=1\linewidth]{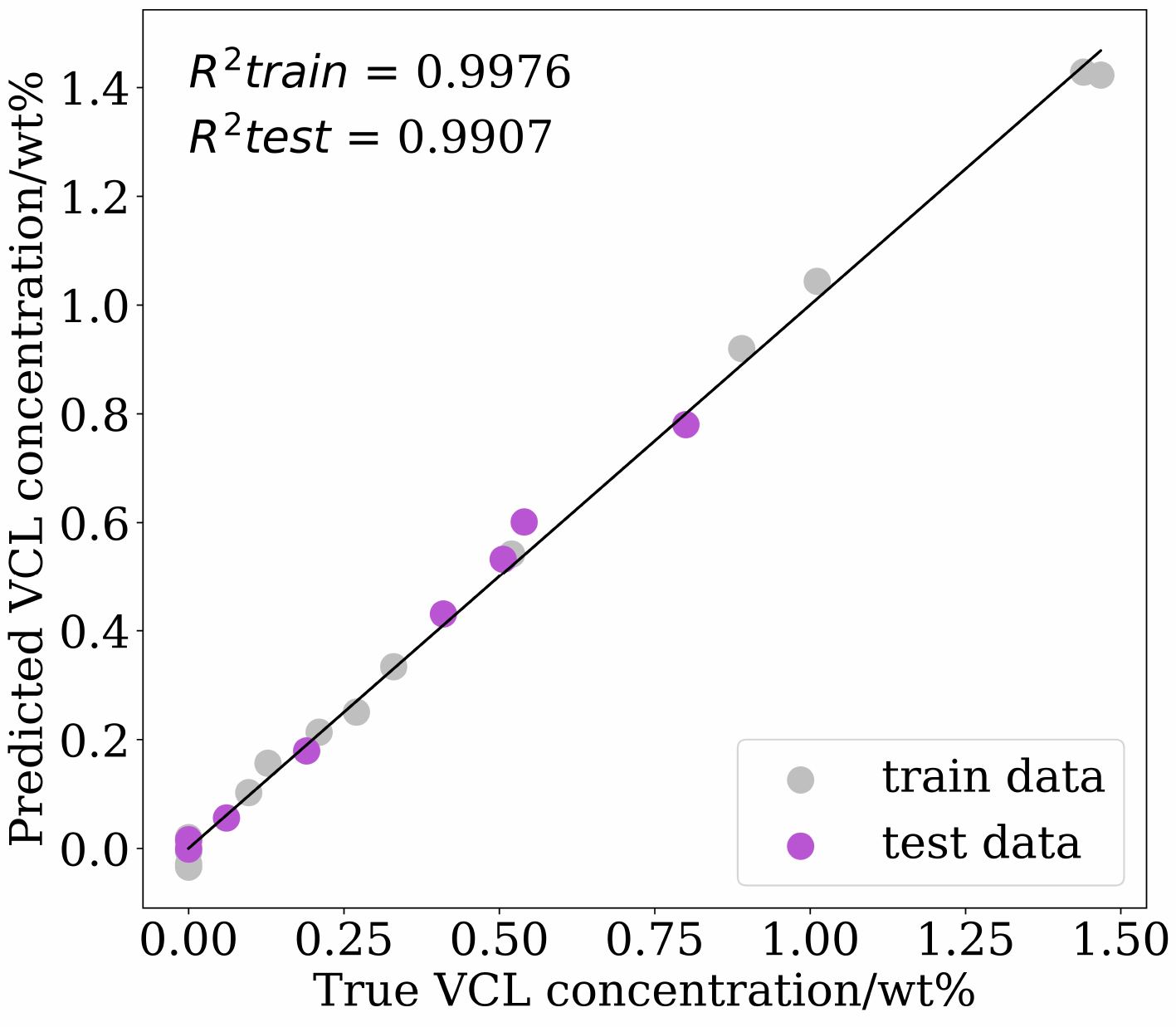}
    \caption{}
    \label{fig:PLS_case 1_monomer}
    \end{subfigure}
    \caption{Monomer concentration predictions with Y-AE (a), ANN (b), XGB (c), and PLS (d). The gray dots represent the training datapoints, while the colored ones represent the testing datapoints. PLS gives the highest prediction results with very few latent variables: this indicates that the spectroscopic data lie on a linear low-dimensional manifold. The predictions from Y-AE and ANN are on the same range.}
    \label{fig:predictions_sep_monomer_case_1}
\end{figure}

After the dimensionality reduction of the spectra, we compare three alternative machine learning methods for the target predictions, namely the Y-AE, ANNs, and extreme gradient boosting (XGB).
The inputs for the three methods are the DMAP coordinates representing the Raman spectra.
We also predict the targets with PLS, using the spectra measurements as inputs to the method.
The outputs for the four methods are the monomer, polymer and water concentration values.
The parity plots for each method, for polymer are shown in Fig.~\ref{fig:predictions_sep_case_1} in the manuscript. For water and monomer the results are shown in Fig.~\ref{fig:predictions_sep_water_case_1} and Fig.~\ref{fig:predictions_sep_monomer_case_1} respectively.

Based on the metrics and the parity plots, we see that Y-shaped autoencoders achieve high prediction performance, even though the predictions from ANN and Y-AE are on the same range. While PLS gives the highest prediction results for monomer. Since we were able to get these results with very few latent variables from a linear method, it implies that the spectra lie on a linear underlying manifold.

\subsection{Case study 2: citric acid dissociation data}\label{separate_remaining_targets_case_2}
\begin{figure}[p] 
    \centering
    \begin{subfigure}{0.44\textwidth}
    \includegraphics[width=1\linewidth]{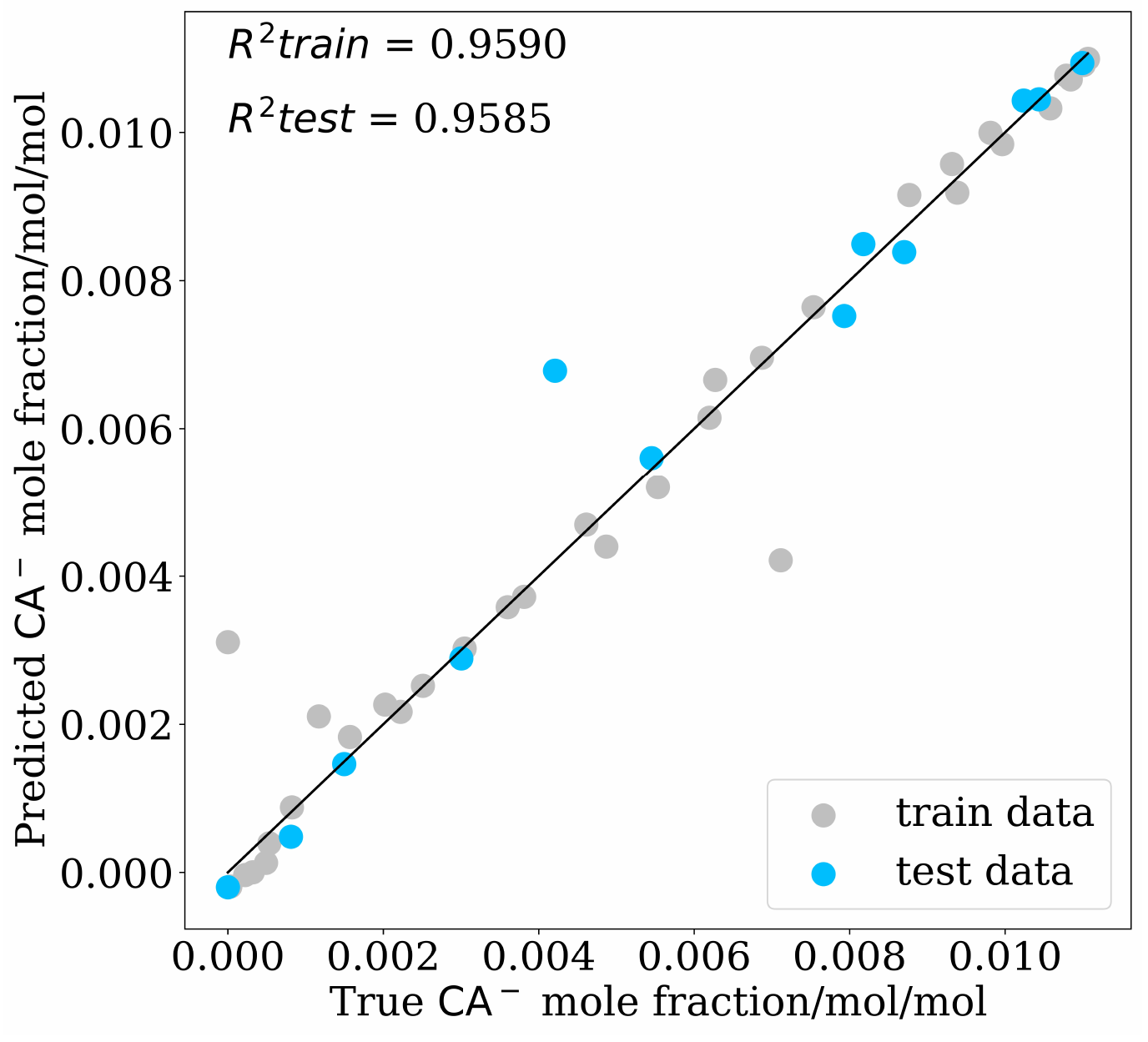}
    \caption{}
    \label{fig:sep_Y-AE_case 2_CA-}
    \end{subfigure}
    \hspace{0.1\textwidth}
    \vspace{0.05\textwidth}
    \begin{subfigure}{0.44\textwidth}
    \includegraphics[width=1\linewidth]{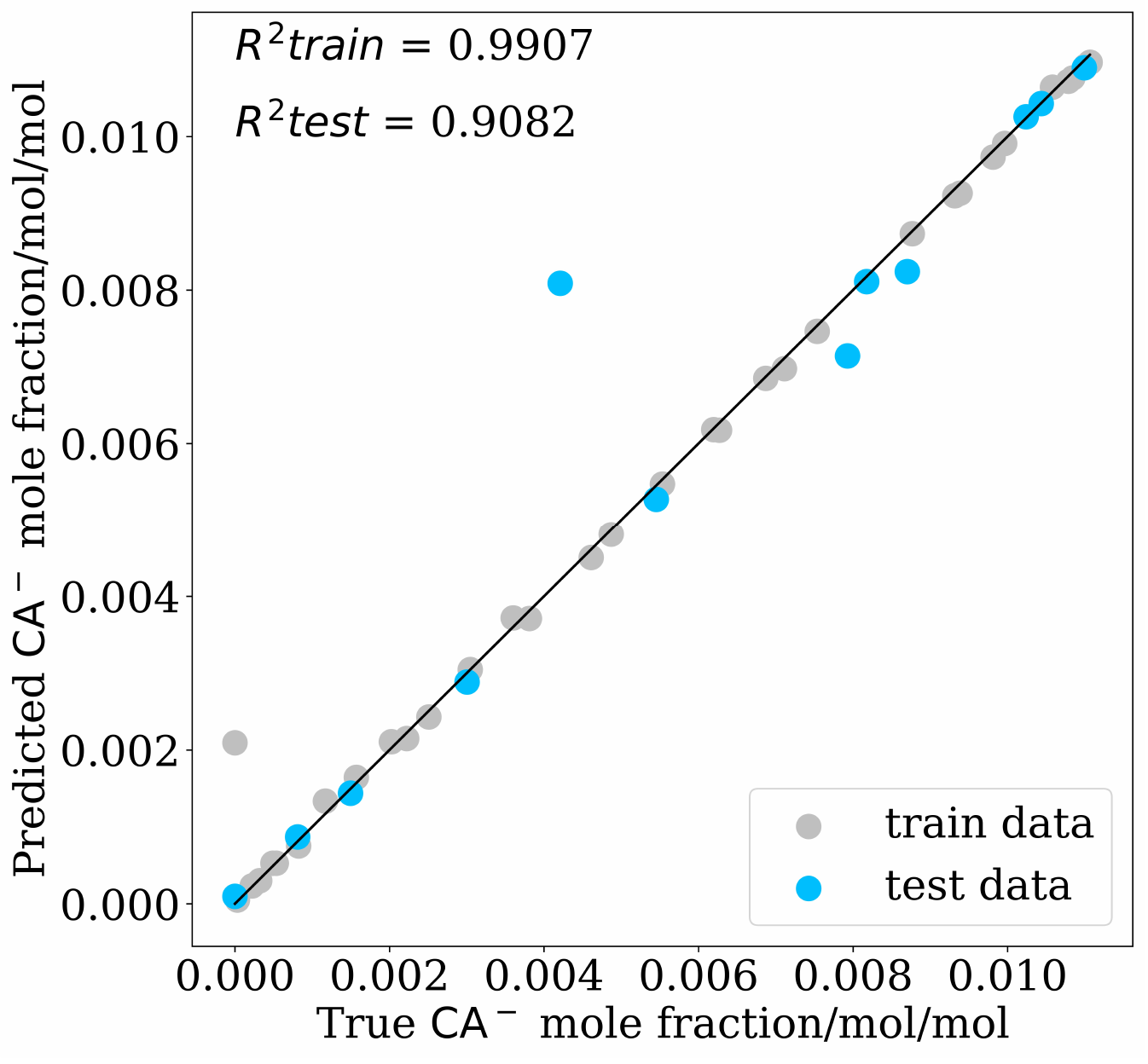}
    \caption{}
    \label{fig:ANN_case 2_CA-}
    \end{subfigure}
    \begin{subfigure}{0.44\textwidth}
    \includegraphics[width=1\linewidth]{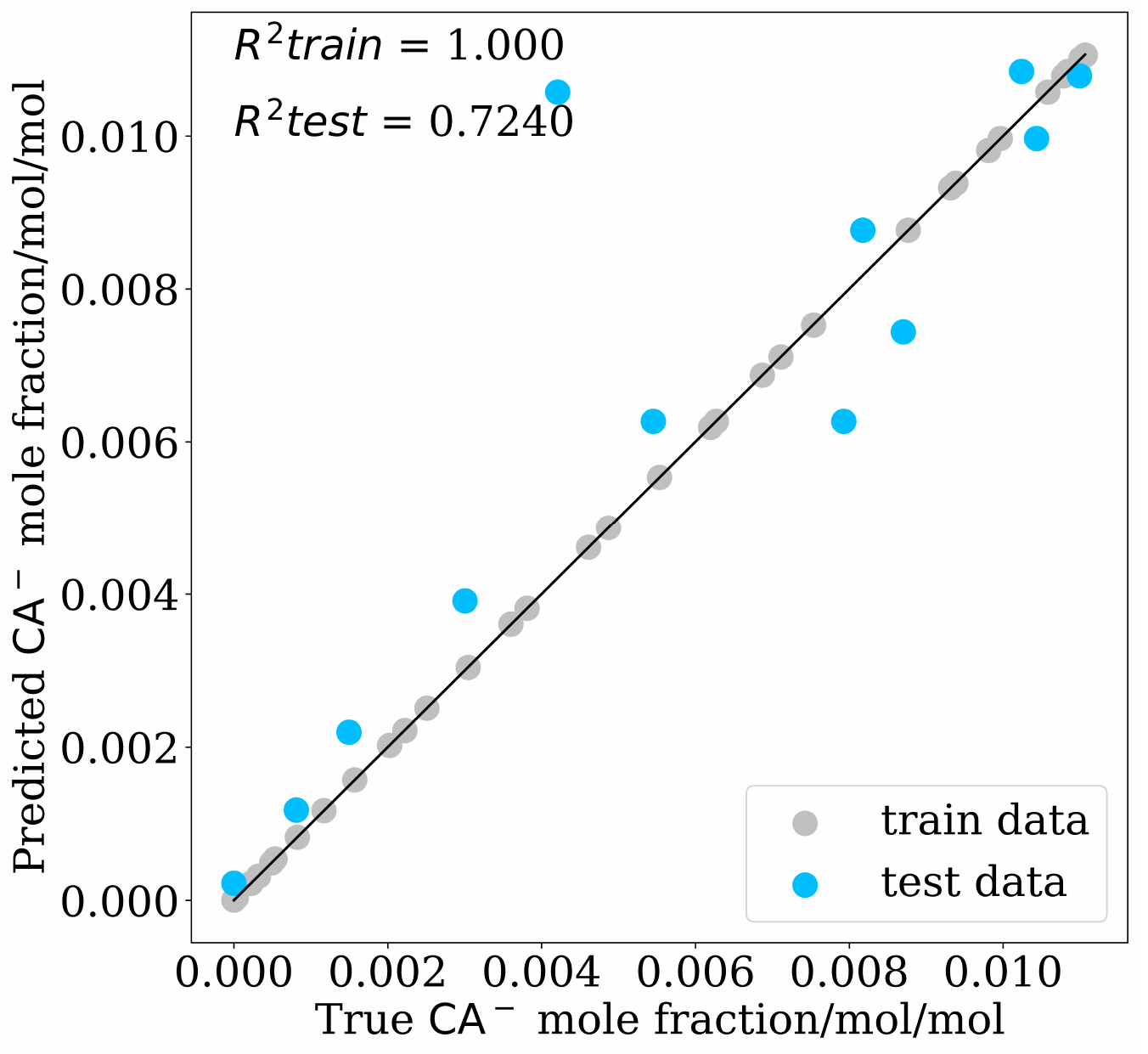}
    \caption{}
    \label{fig:XGBOOST_case 2_CA-}
    \end{subfigure}
    \hspace{0.1\textwidth}
    \vspace{0.05\textwidth}
    \begin{subfigure}{0.44\textwidth}
    \includegraphics[width=1\linewidth]{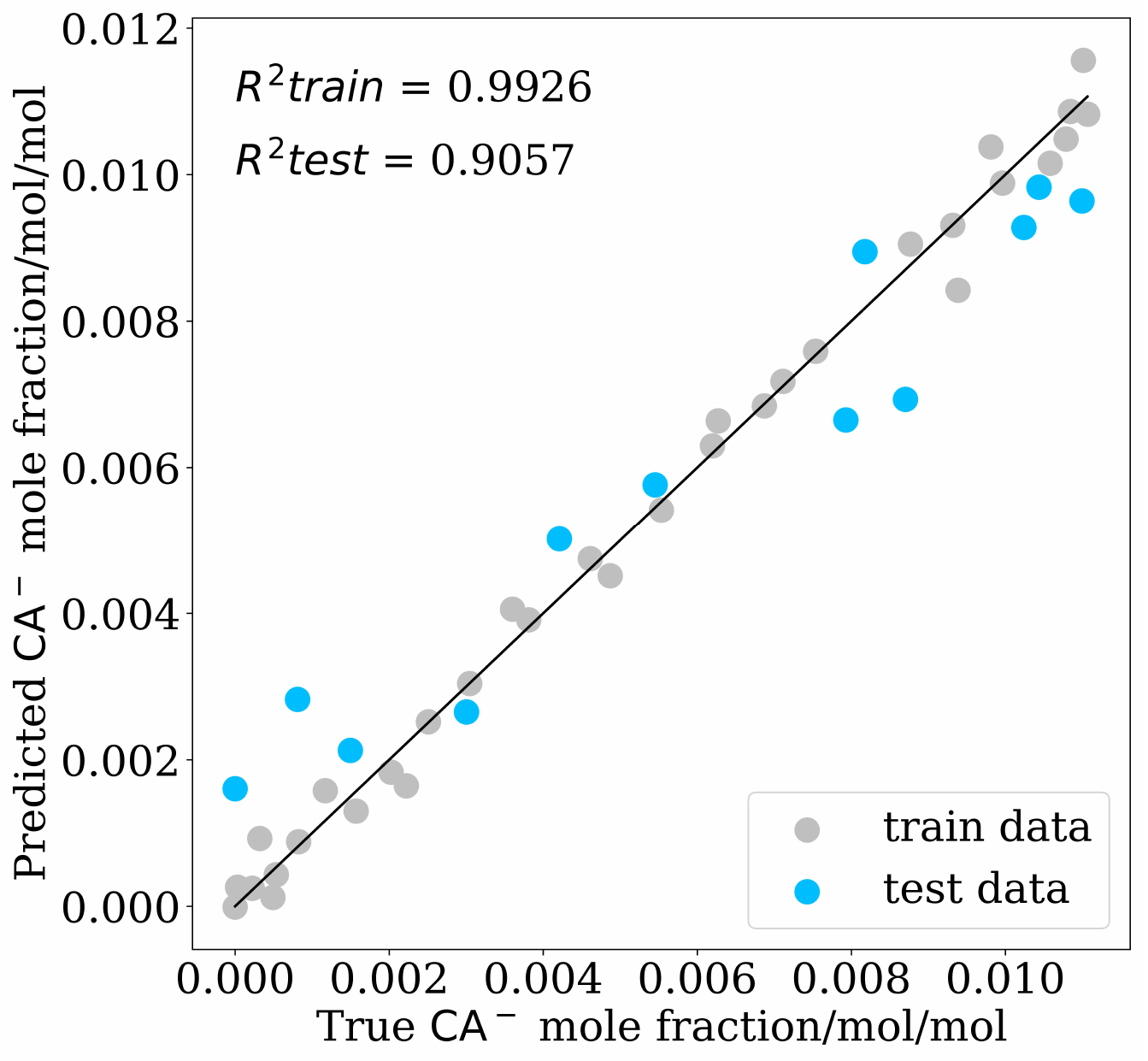}
    \caption{}
    \label{fig:PLS_case 2_CA-}
    \end{subfigure}
    \caption{Concentration predictions of CA\(^{-}\) with Y-AE (a), ANN (b),XGB (c), and PLS. The gray dots represent the training datapoints, while the colored ones represent the testing datapoints. Y-AE gives the highest prediction results, while the predictions from PLS and XGB are on the same range.}
    \label{fig:predictions_sep_CA-_case_2}
\end{figure}

\begin{figure}[p]
    \centering
    \begin{subfigure}{0.44\textwidth}
    \includegraphics[width=1\linewidth]{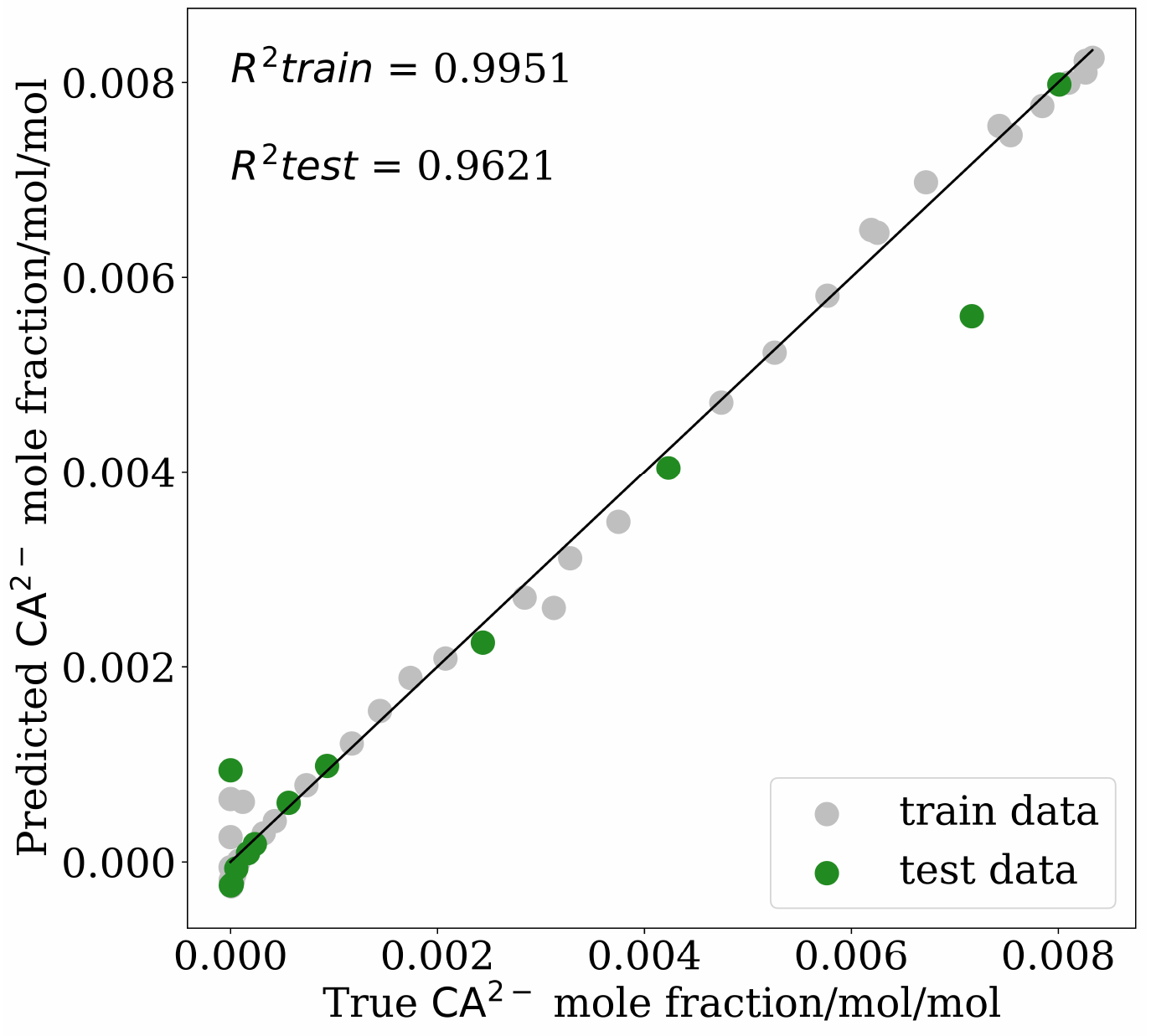}
    \caption{}
    \label{fig:sep_Y-AE_case 2_CA2-}
    \end{subfigure}
    \hspace{0.1\textwidth}
    \vspace{0.05\textwidth}
    \begin{subfigure}{0.44\textwidth}
    \includegraphics[width=1\linewidth]{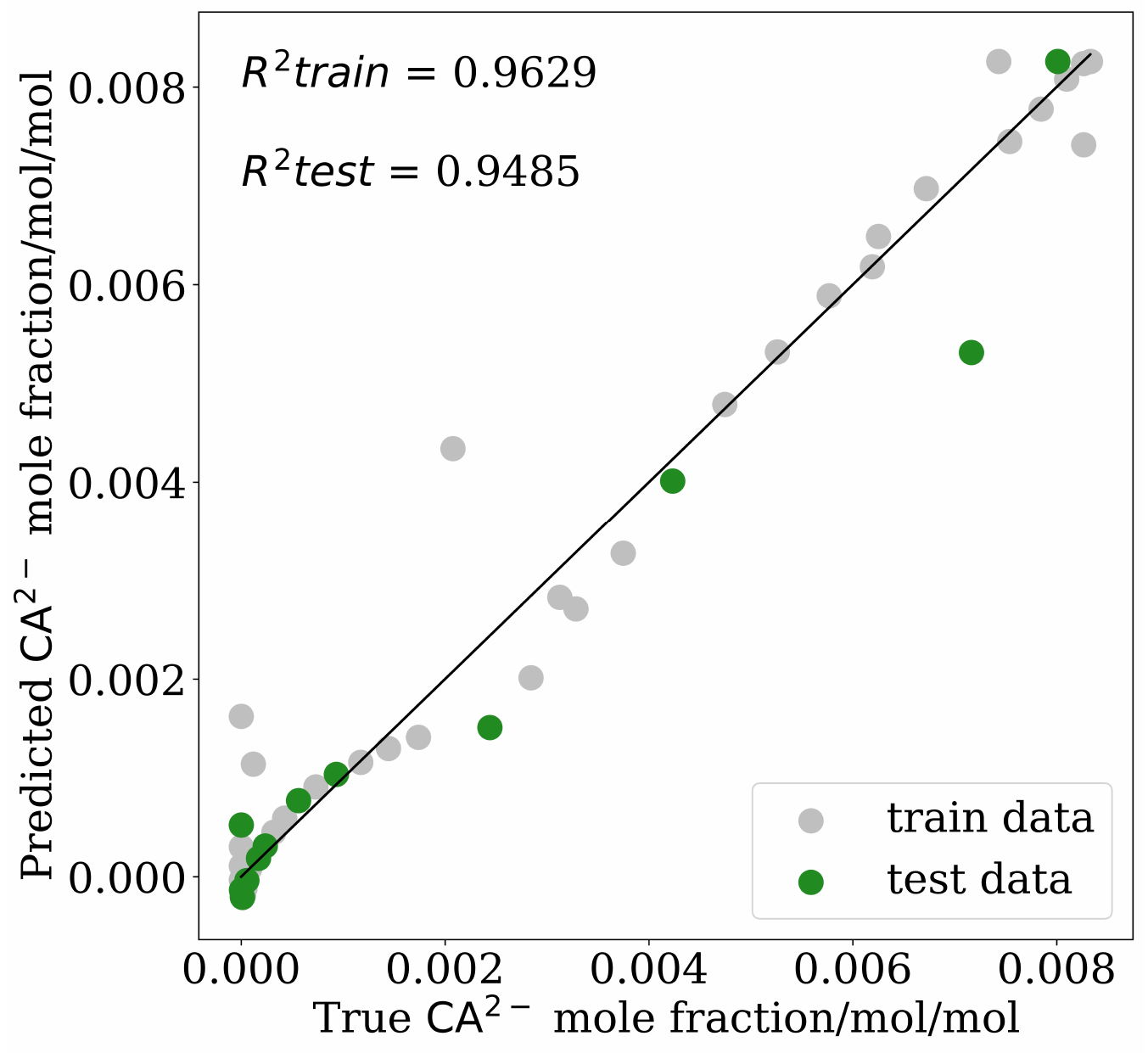}
    \caption{}
    \label{fig:ANN_case 2_CA2-}
    \end{subfigure}
    \begin{subfigure}{0.44\textwidth}
    \includegraphics[width=1\linewidth]{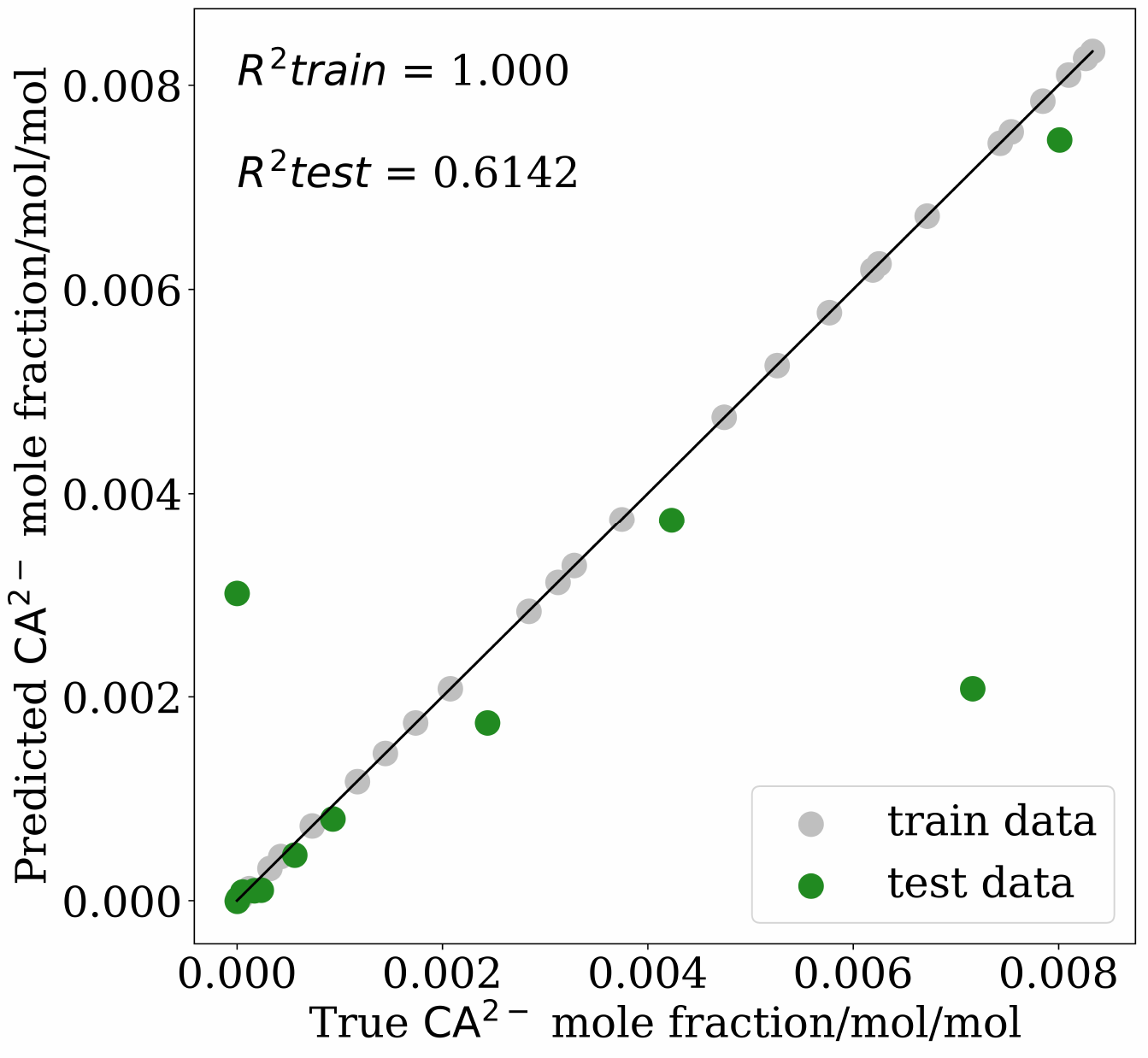}
    \caption{}
    \label{fig:XGBOOST_case 2_CA2-}
    \end{subfigure}
    \hspace{0.1\textwidth}
    \vspace{0.05\textwidth}
    \begin{subfigure}{0.44\textwidth}
    \includegraphics[width=1\linewidth]{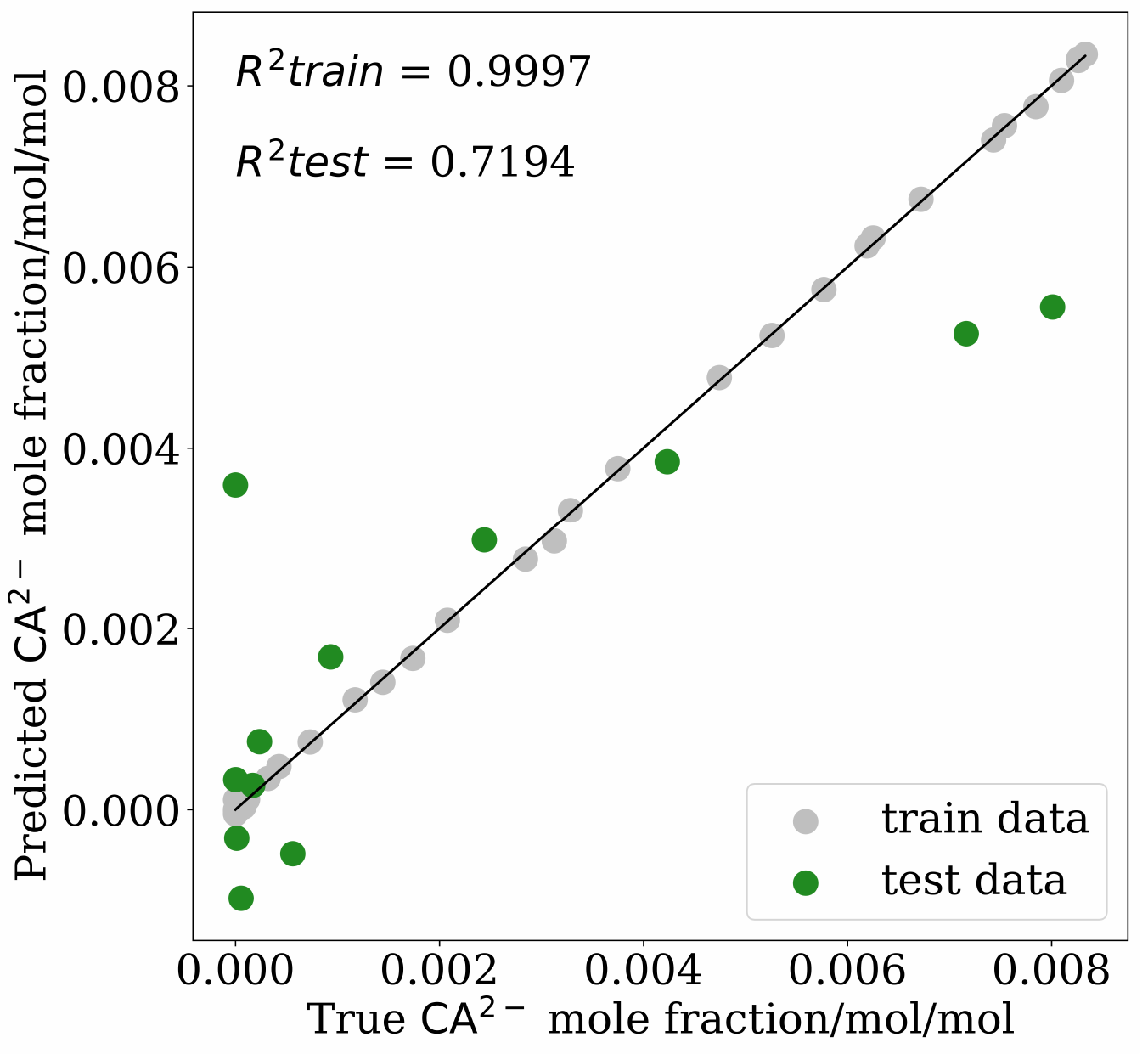}
    \caption{}
    \label{fig:PLS_case 2_CA2-}
    \end{subfigure}
    \caption{Concentration predictions of CA\(^{2-}\) with Y-AE (a), ANN (b),XGB (c), and PLS. The gray dots represent the training datapoints, while the colored ones represent the testing datapoints. Y-AE gives the highest prediction results, while the predictions from PLS and XGB are on the same range.}
    \label{fig:predictions_sep_CA2-_case_2}
\end{figure}

\begin{figure}[p]
    \centering
    \begin{subfigure}{0.44\textwidth}
    \includegraphics[width=1\linewidth]{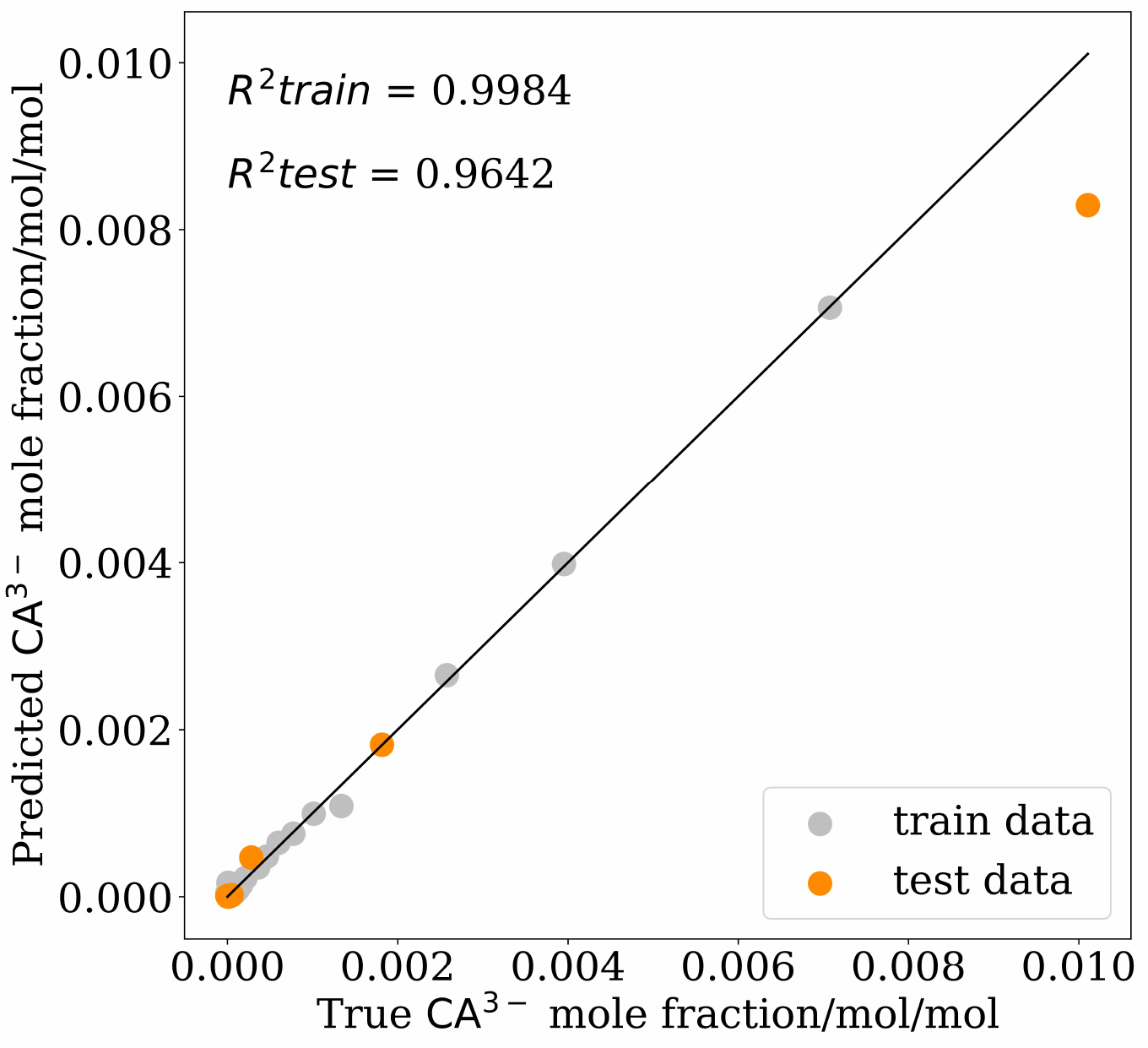}
    \caption{}
    \label{fig:sep_Y-AE_case 2_CA3-}
    \end{subfigure}
    \hspace{0.1\textwidth}
    \vspace{0.05\textwidth}
    \begin{subfigure}{0.44\textwidth}
    \includegraphics[width=1\linewidth]{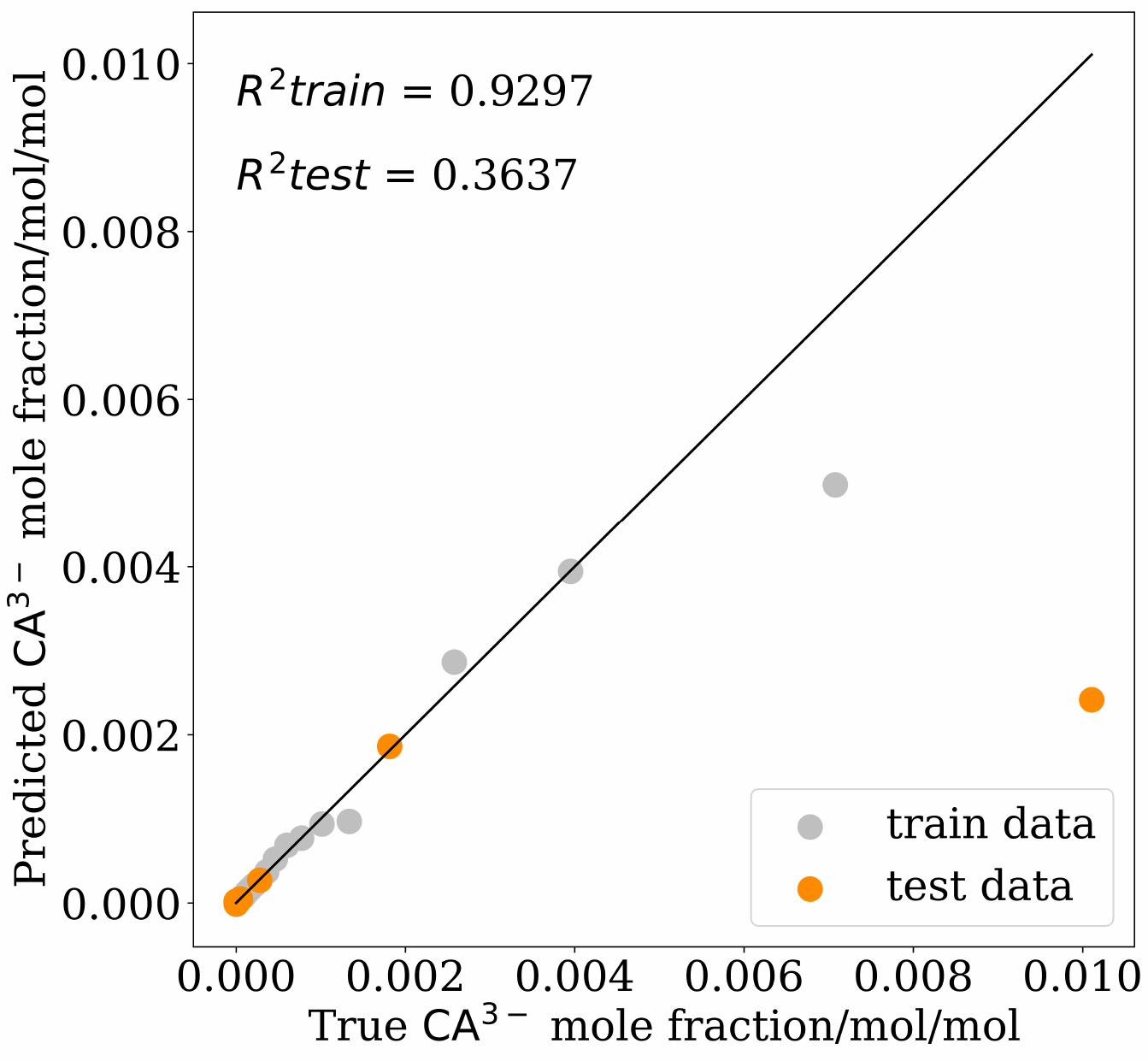}
    \caption{}
    \label{fig:ANN_case 2_CA3-}
    \end{subfigure}
    \begin{subfigure}{0.44\textwidth}
    \includegraphics[width=1\linewidth]{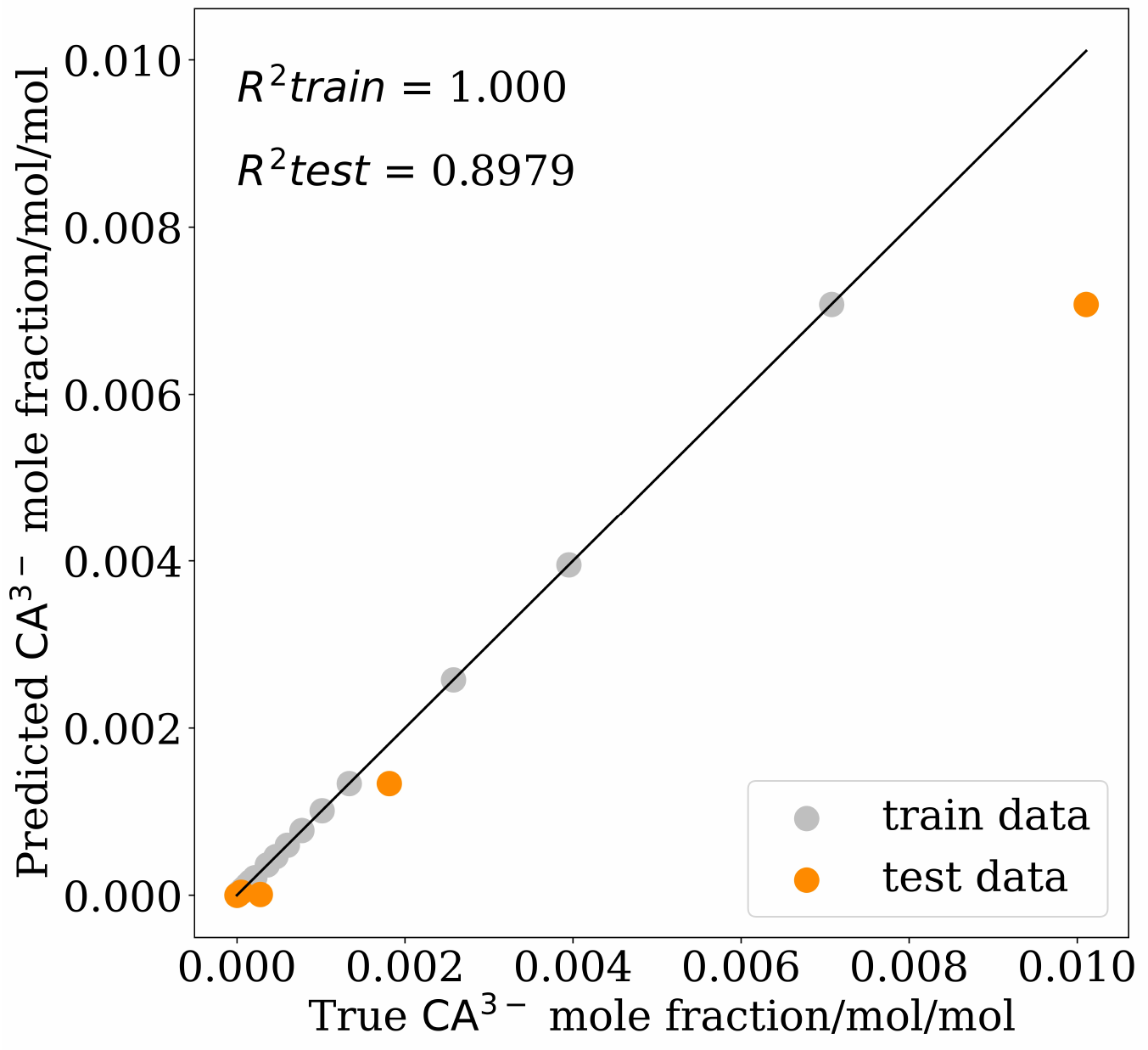}
    \caption{}
    \label{fig:XGBOOST_case 2_CA3-}
    \end{subfigure}
    \hspace{0.1\textwidth}
    \vspace{0.05\textwidth}
    \begin{subfigure}{0.44\textwidth}
    \includegraphics[width=1\linewidth]{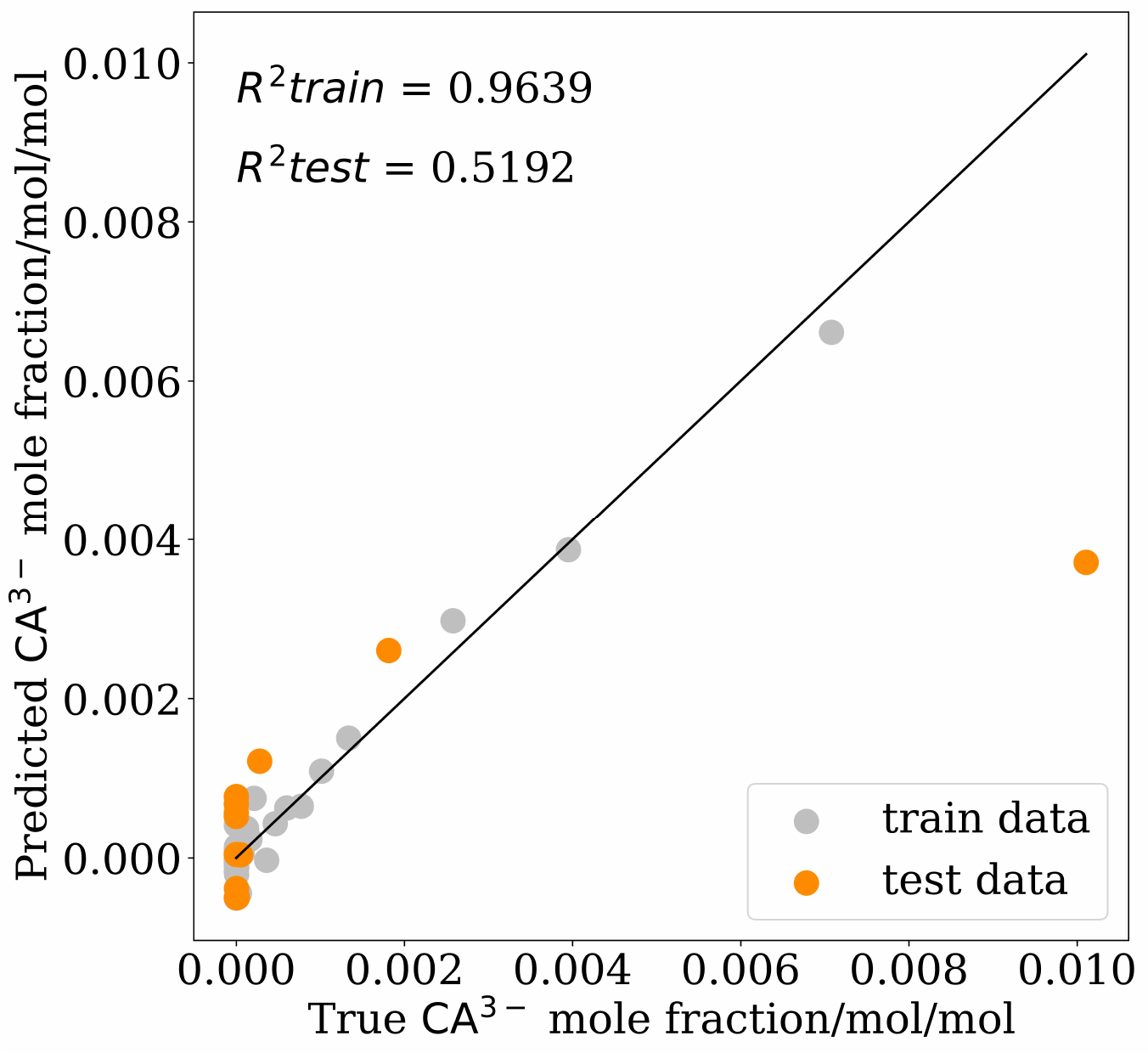}
    \caption{}
    \label{fig:PLS_case 2_CA3-}
    \end{subfigure}
    \caption{Concentration predictions of CA\(^{3-}\) with Y-AE (a), ANN (b),XGB (c), and PLS. The gray dots represent the training datapoints, while the colored ones represent the testing datapoints. Y-AE gives the highest prediction results, while the predictions from PLS and XGB are on the same range.}
    \label{fig:predictions_sep_CA3-_case_2}
\end{figure}

\begin{figure}[p]
    \centering
    \begin{subfigure}{0.44\textwidth}
    \includegraphics[width=1\linewidth]{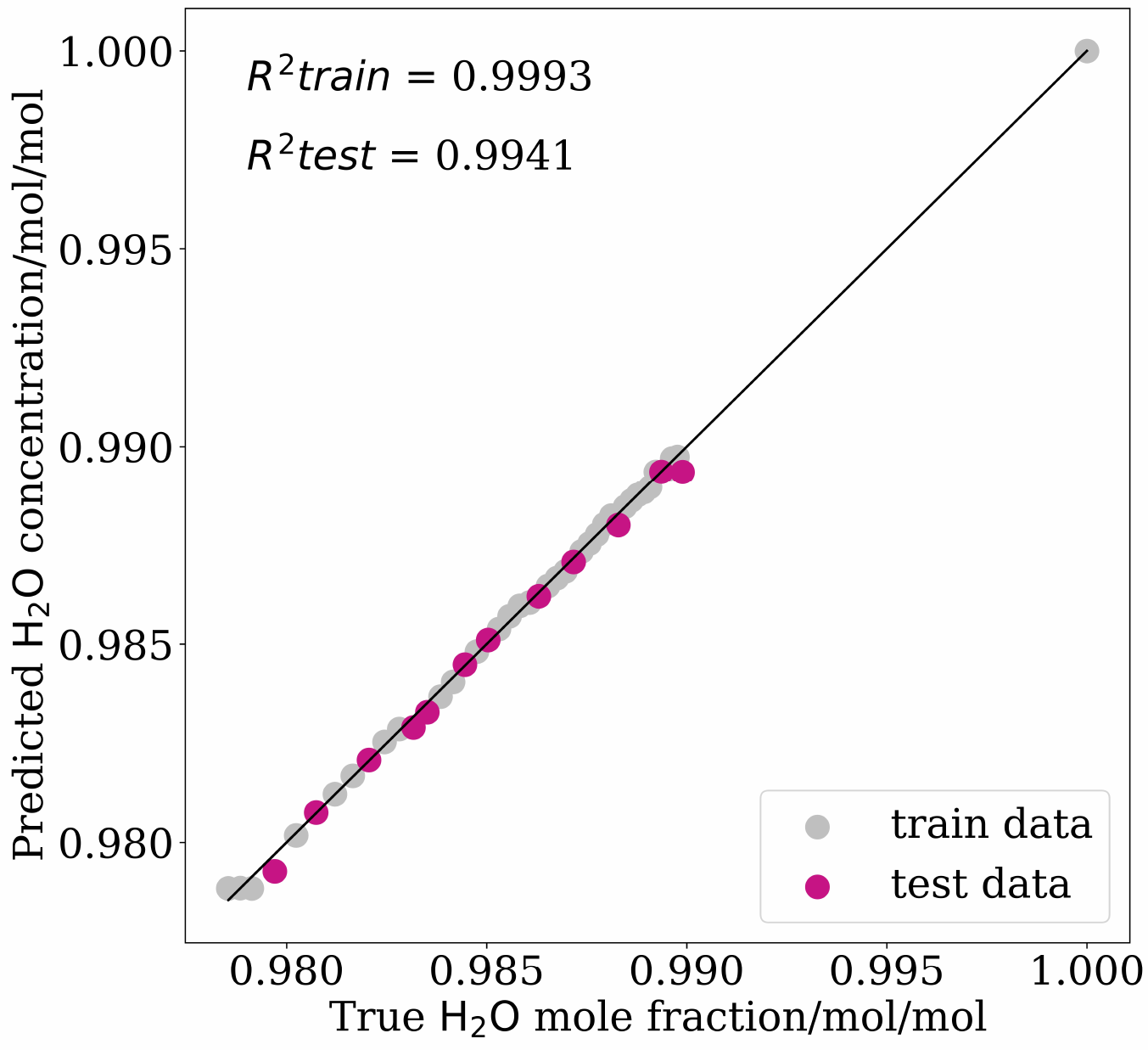}
    \caption{}
    \label{fig:sep_Y-AE_case 2_water}
    \end{subfigure}
    \hspace{0.1\textwidth}
    \vspace{0.05\textwidth}
    \begin{subfigure}{0.44\textwidth}
    \includegraphics[width=1\linewidth]{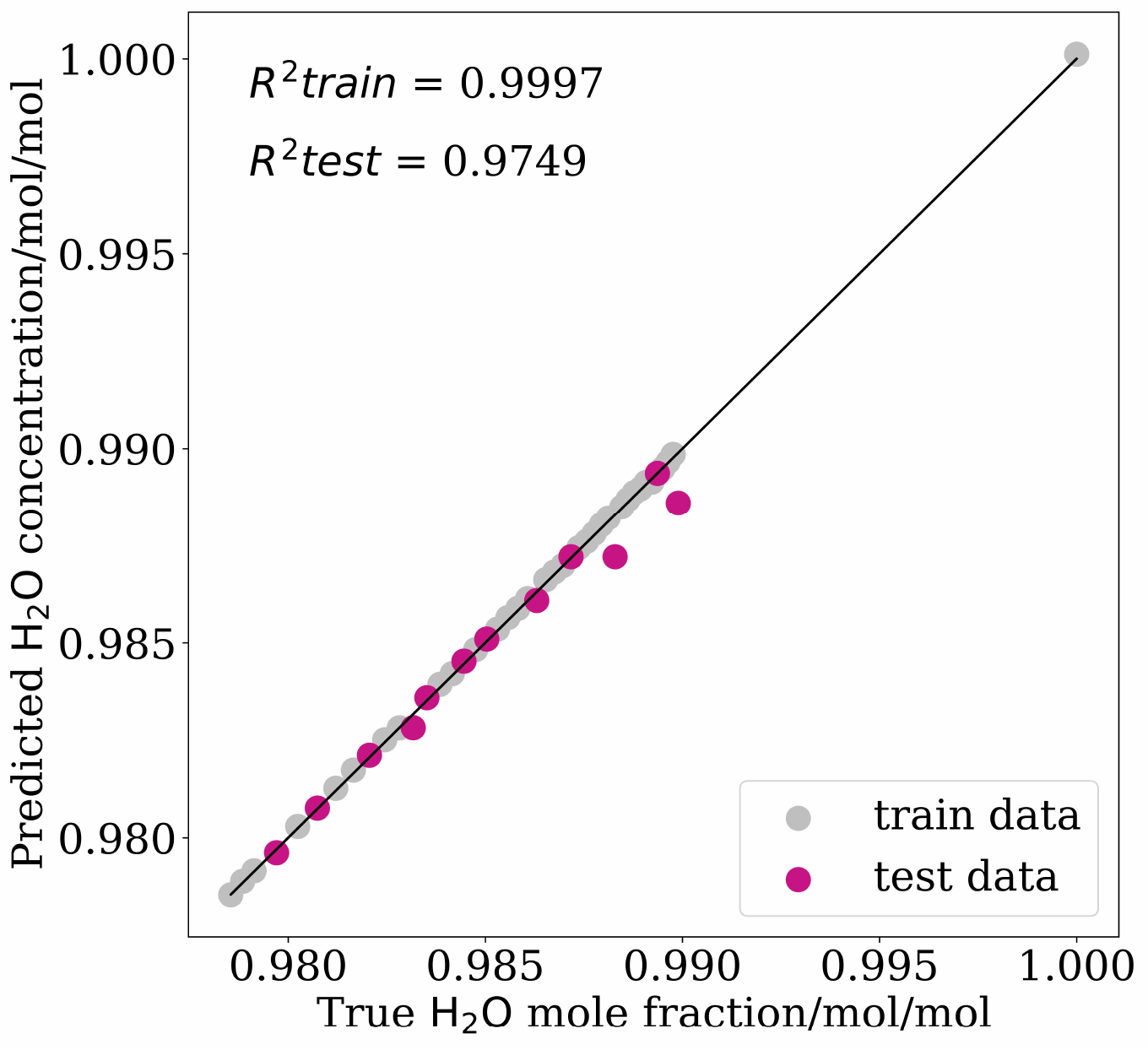}
    \caption{}
    \label{fig:ANN_case 2_water}
    \end{subfigure}
    \begin{subfigure}{0.44\textwidth}
    \includegraphics[width=1\linewidth]{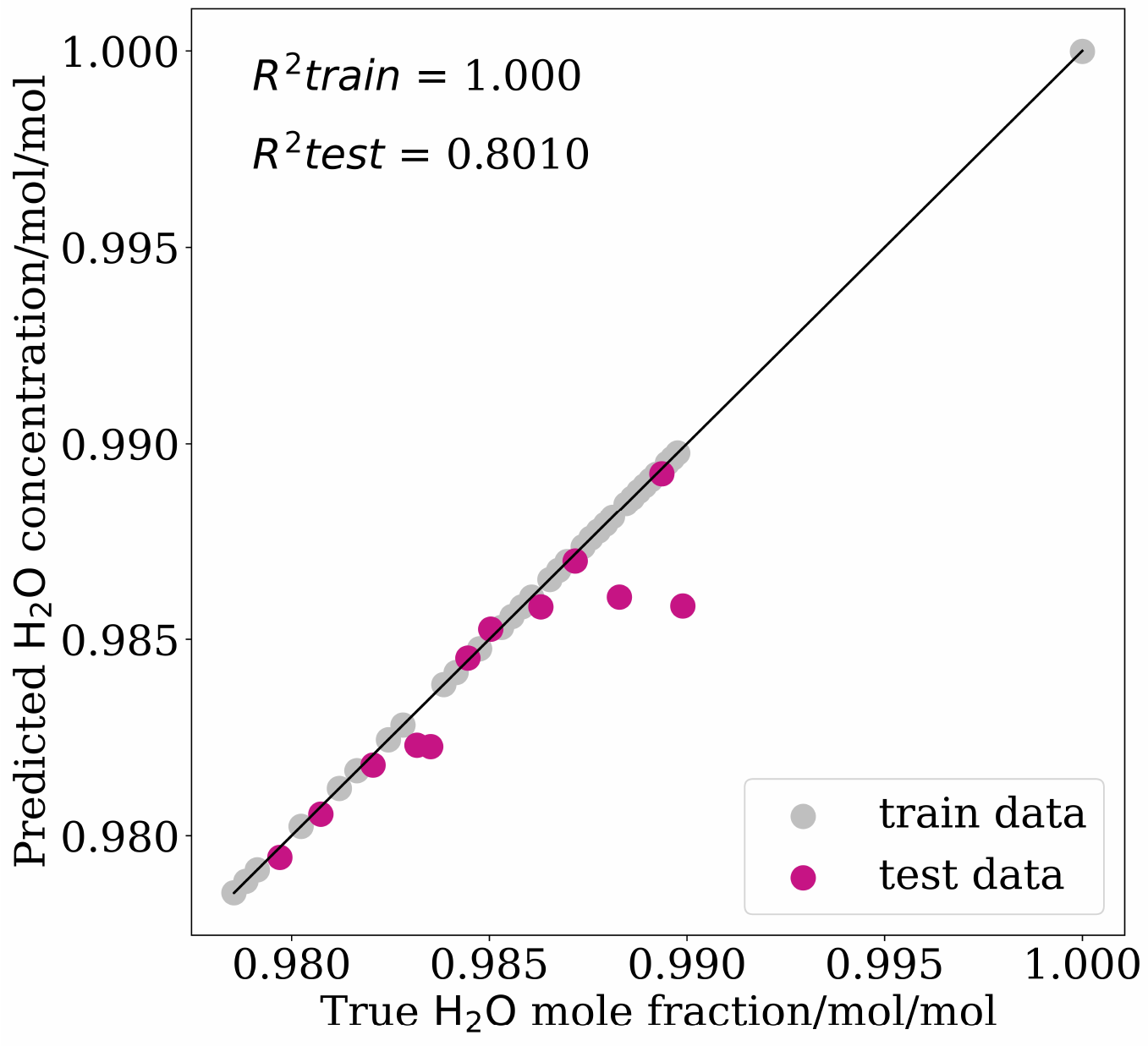}
    \caption{}
    \label{fig:XGBOOST_case 2_water}
    \end{subfigure}
    \hspace{0.1\textwidth}
    \vspace{0.05\textwidth}
    \begin{subfigure}{0.44\textwidth}
    \includegraphics[width=1\linewidth]{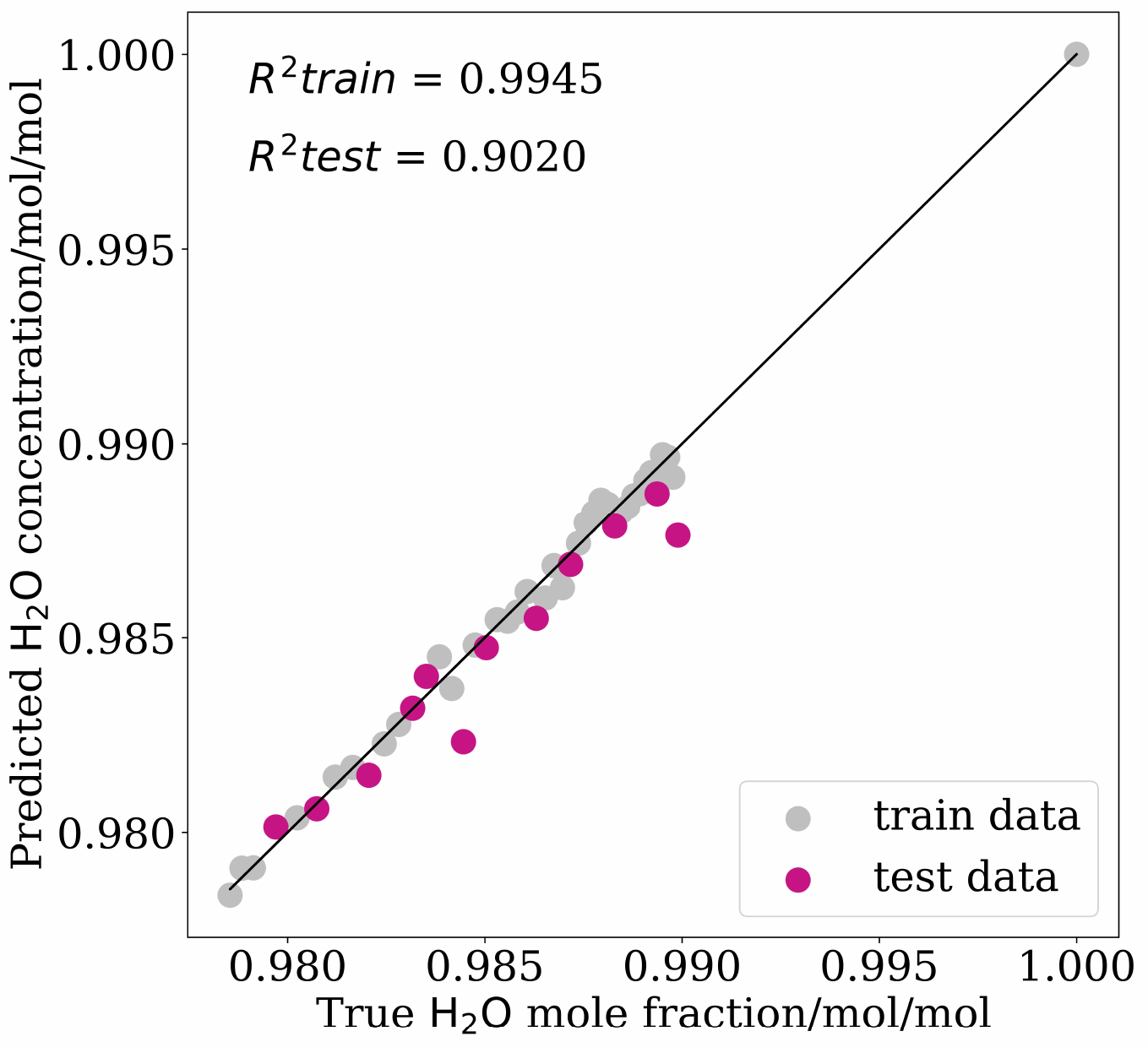}
    \caption{}
    \label{fig:PLS_case 2_water}
    \end{subfigure}
    \caption{Concentration predictions of water with Y-AE (a), ANN (b),XGB (c), and PLS. The gray dots represent the training datapoints, while the colored ones represent the testing datapoints. Y-AE gives the highest prediction results, while the predictions from PLS and XGB are on the same range.}
    \label{fig:predictions_sep_water_case_2}
\end{figure}

For this case study we follow the same workflow as above. Here, the inputs are the DMAP coordinates representing the Raman spectra, and the outputs are the concentration values of citric acid, its dissociated acid species in water, CA\(^{-}\), CA\(^{2-}\), CA\(^{3-}\), and water.
The parity plots for each method, for citric acid are shown in Fig.~\ref{fig:predictions_sep_case_2} in the manuscript.
The parity plots with the prediction results for each method, for the remaining targets are shown in Fig.~\ref{fig:predictions_sep_CA-_case_2}, Fig.~\ref{fig:predictions_sep_CA2-_case_2}, Fig.~\ref{fig:predictions_sep_CA3-_case_2}, and Fig.~\ref{fig:predictions_sep_water_case_2}.

Here we see that Y-shaped autoencoders achieve higher prediction performance than the rest of the methods. The predictions from PLS are on the same range as XGB.

\section{Uncertainty quantification study}\label{SI_UQ}
In the following, Section~\ref{SI_UQ_case_1} and Section~\ref{SI_UQ_case_2} present the results of the uncertainty quantification study (UQ) on the single-target predictions from the Y-shaped conformal autoencoder (Y-AE) and the artificial neural network (ANN), and the multi-target predictions from the Y-AE, for the first and the second case study respectively.

\subsection{Case study 1: microgel synthesis data}\label{SI_UQ_case_1}
\begin{figure}[ht] 
    \centering
    \begin{subfigure}{0.4\textwidth}
    \includegraphics[width=1\linewidth]{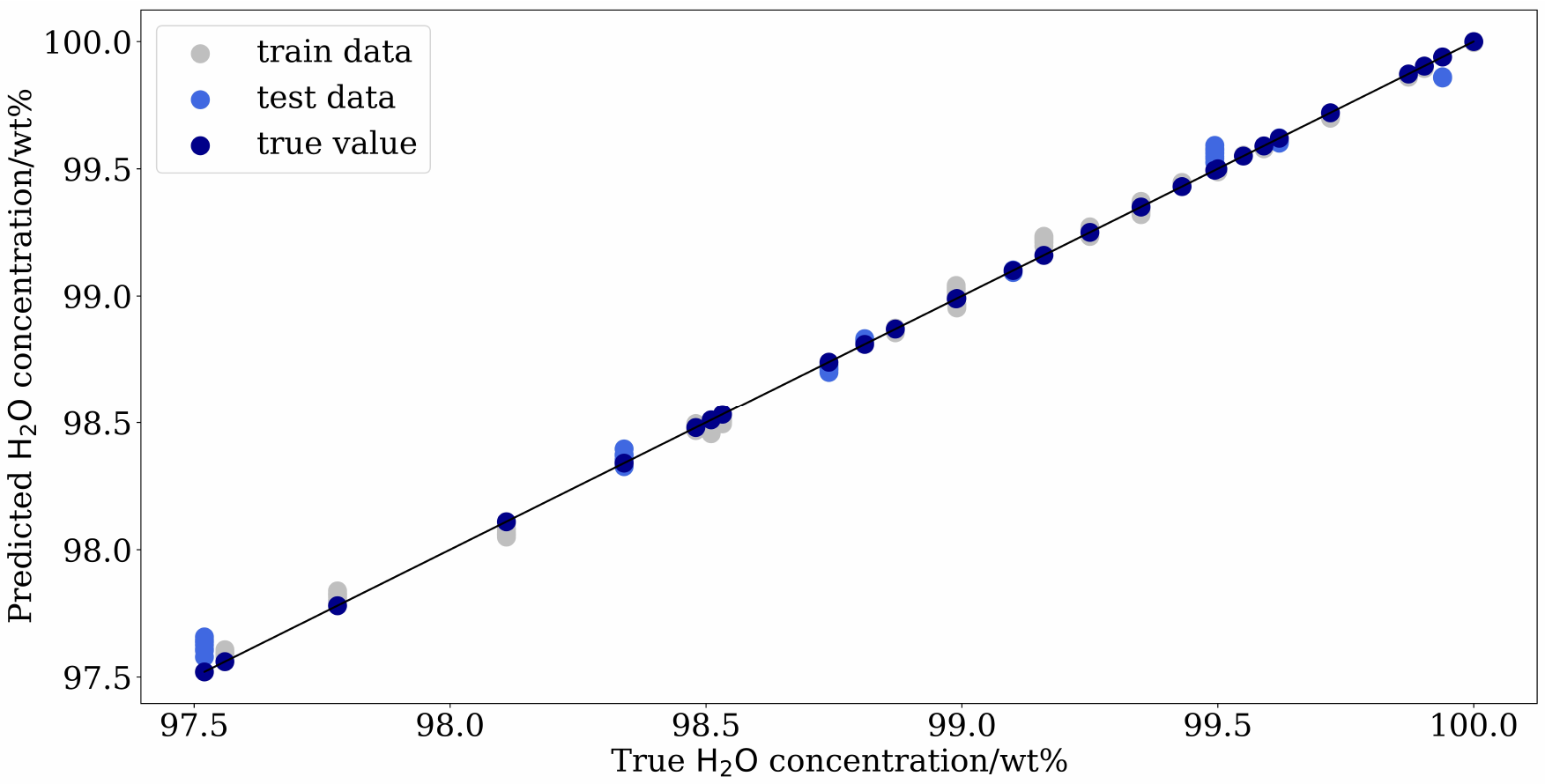}
    \caption{}
    \label{fig:UQ_sep_Y-AE_H2O_case 1}
    \end{subfigure}
    \hspace{0.1\textwidth}
    \vspace{0.05\textwidth}
    \begin{subfigure}{0.4\textwidth}
    \includegraphics[width=1\linewidth]{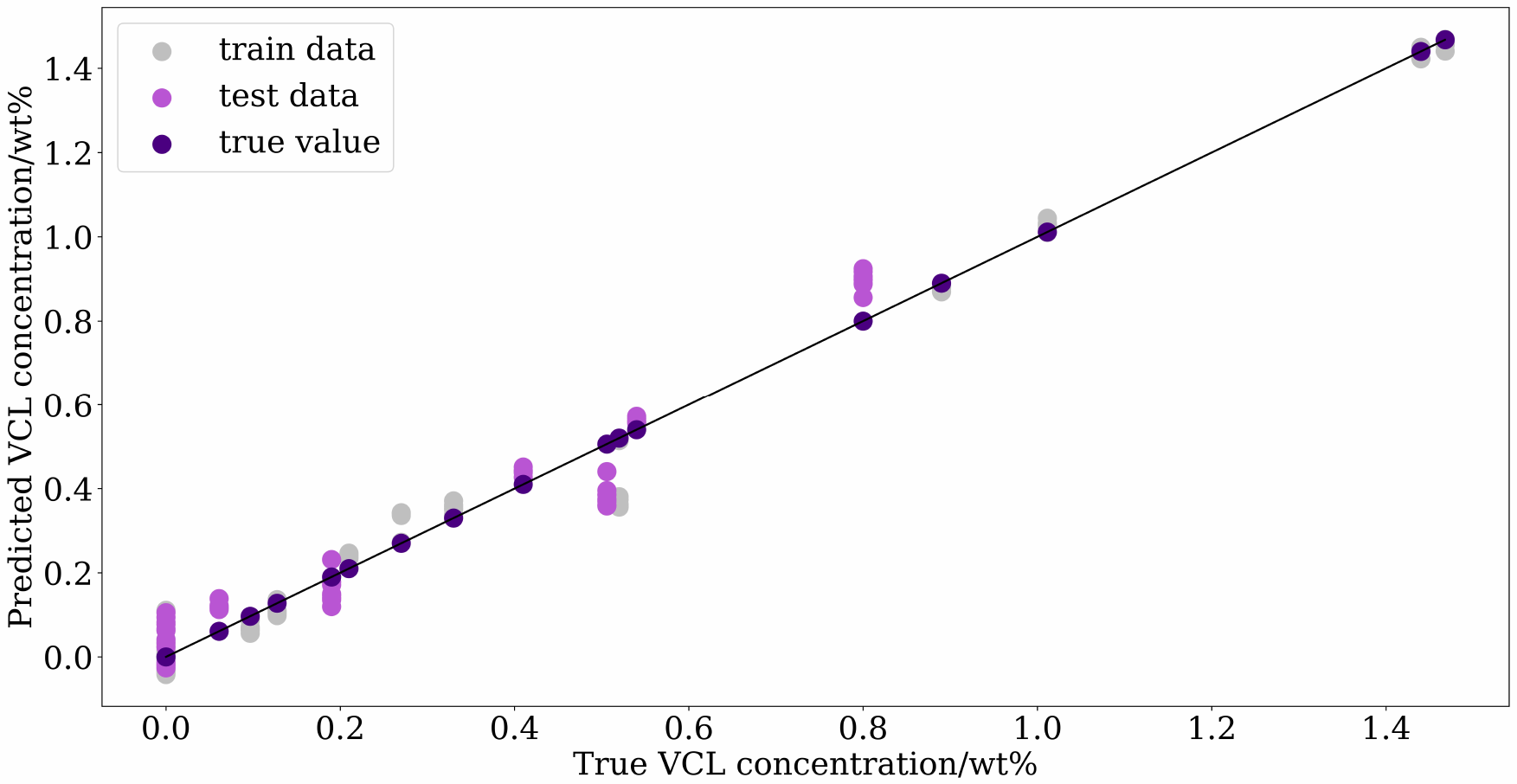}
    \caption{}
    \label{fig:UQ_sep_Y-AE_mon_case 1}
    \end{subfigure}
    \begin{subfigure}{0.4\textwidth}
    \includegraphics[width=1\linewidth]{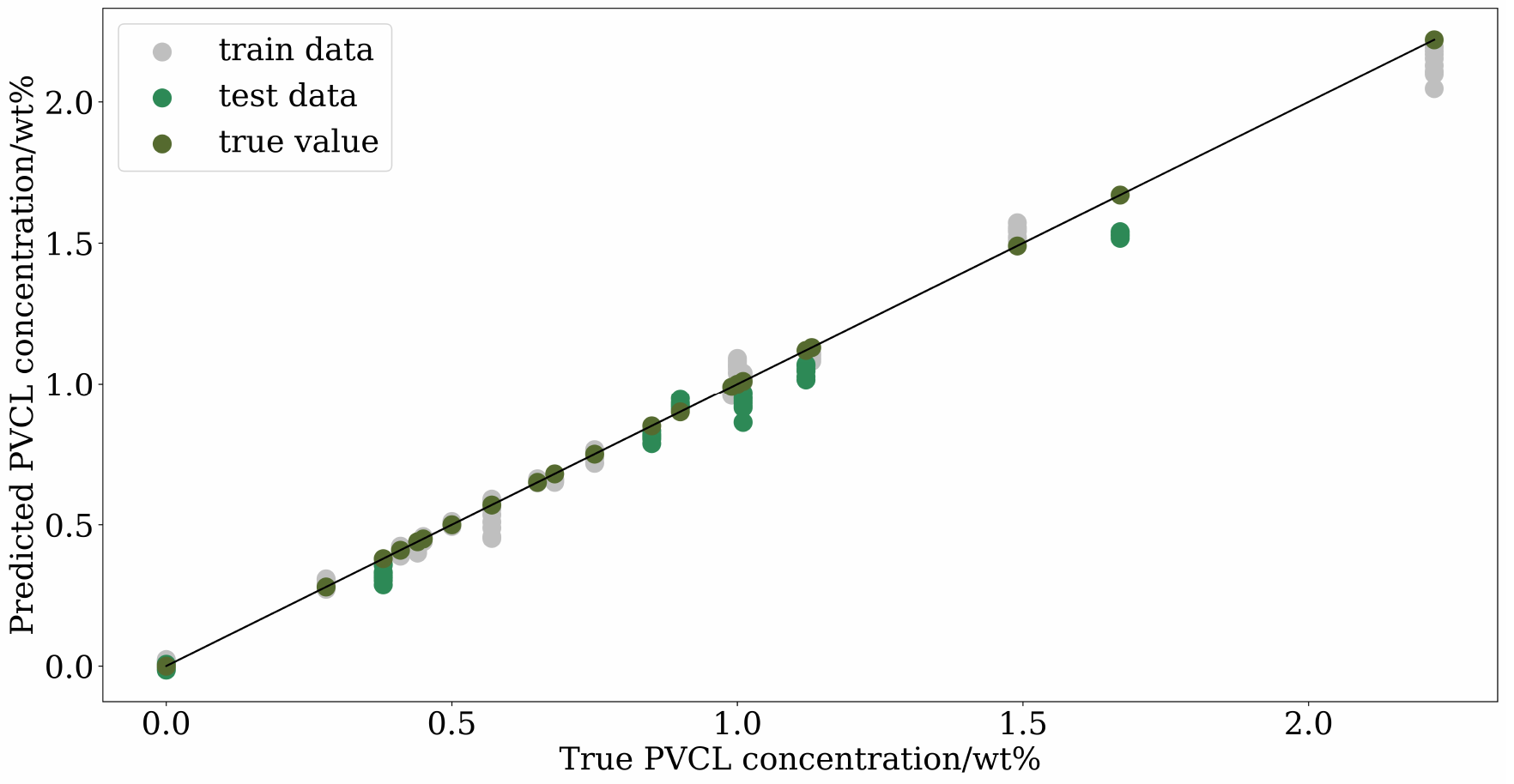}
    \caption{}
    \label{fig:UQ_sep_Y-AE_pol_case 1}
    \end{subfigure}
    \caption{Uncertainty quantification study plots from the separate target prediction with Y-AE for the first case study, for water (a), monomer (b), and polymer (c). There is a higher discrepancy for the predicted test datapoints compared to the predicted train datapoints, because the test datapoints are new data for the models. The predictions from the ten models are on the same range.}
    \label{fig:UQ_sep Y-AE_case 1}
\end{figure}

\begin{figure}[ht] 
    \centering
    \begin{subfigure}{0.4\textwidth}
    \includegraphics[width=1\linewidth]{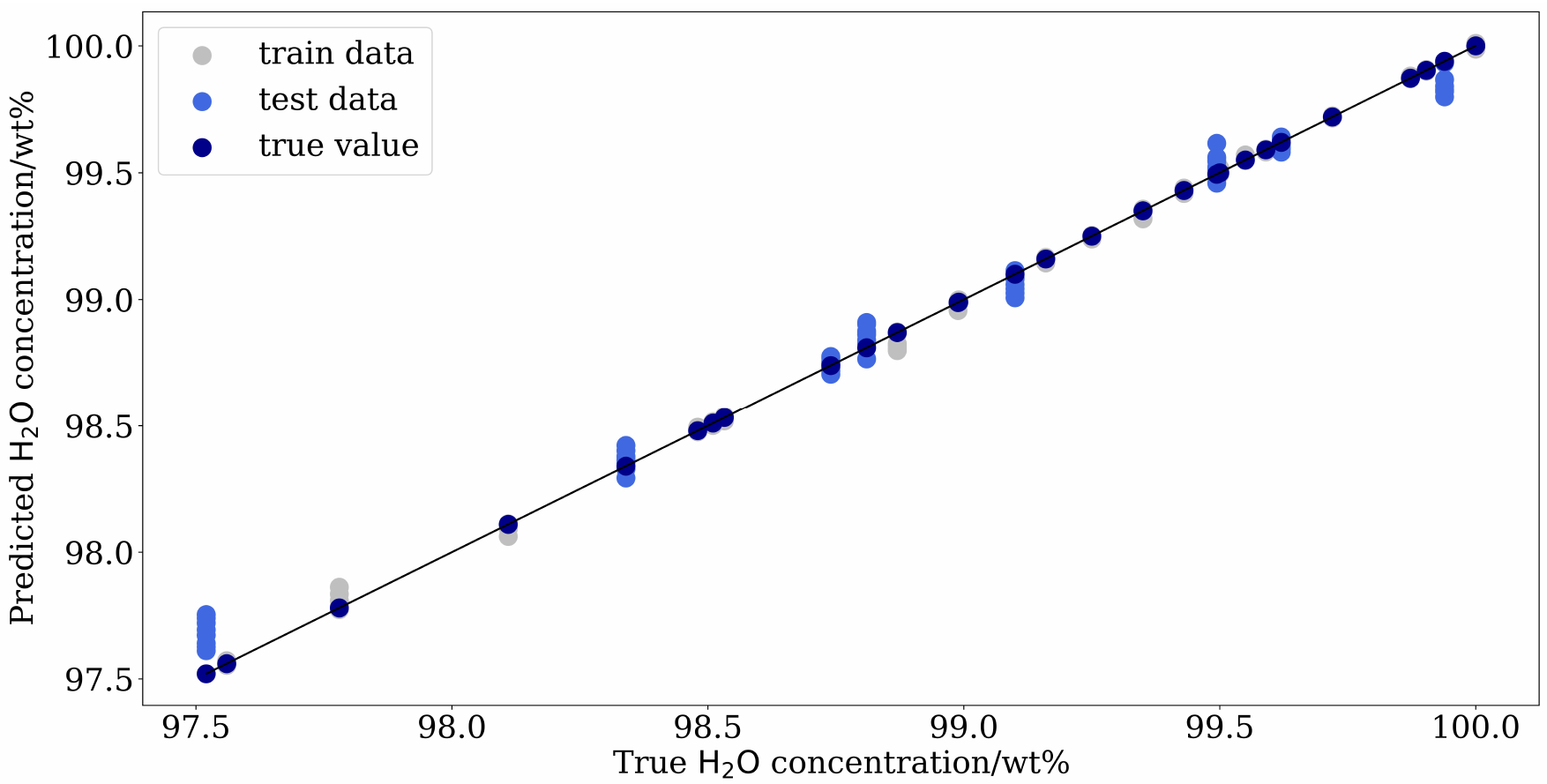}
    \caption{}
    \label{fig:UQ_ANN_H2O_case 1}
    \end{subfigure}
    \hspace{0.1\textwidth}  
    \vspace{0.05\textwidth}
    \begin{subfigure}{0.4\textwidth}
    \includegraphics[width=1\linewidth]{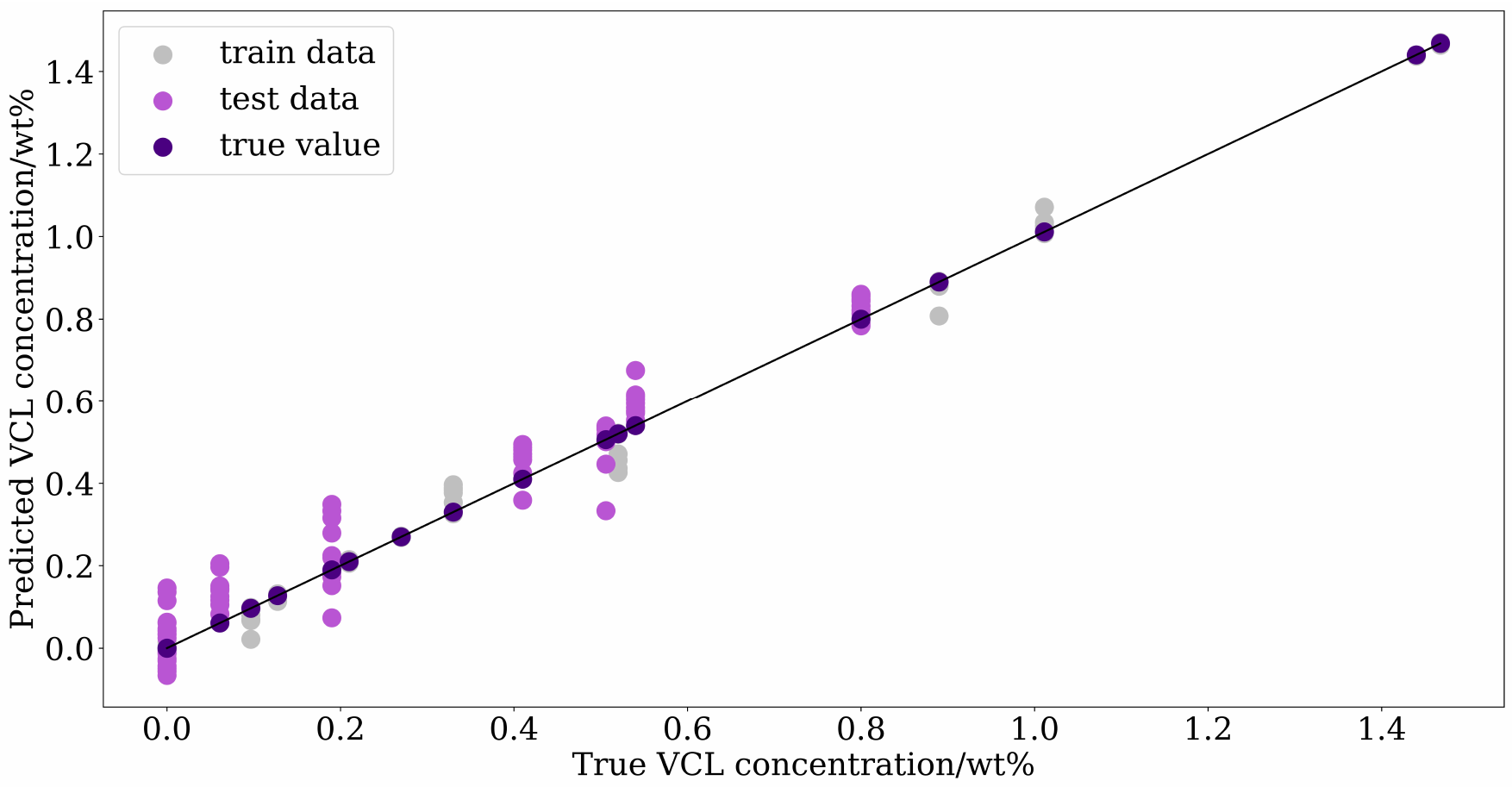}
    \caption{}
    \label{fig:UQ_ANN_mon_case 1}
    \end{subfigure}
    \begin{subfigure}{0.4\textwidth}
    \includegraphics[width=1\linewidth]{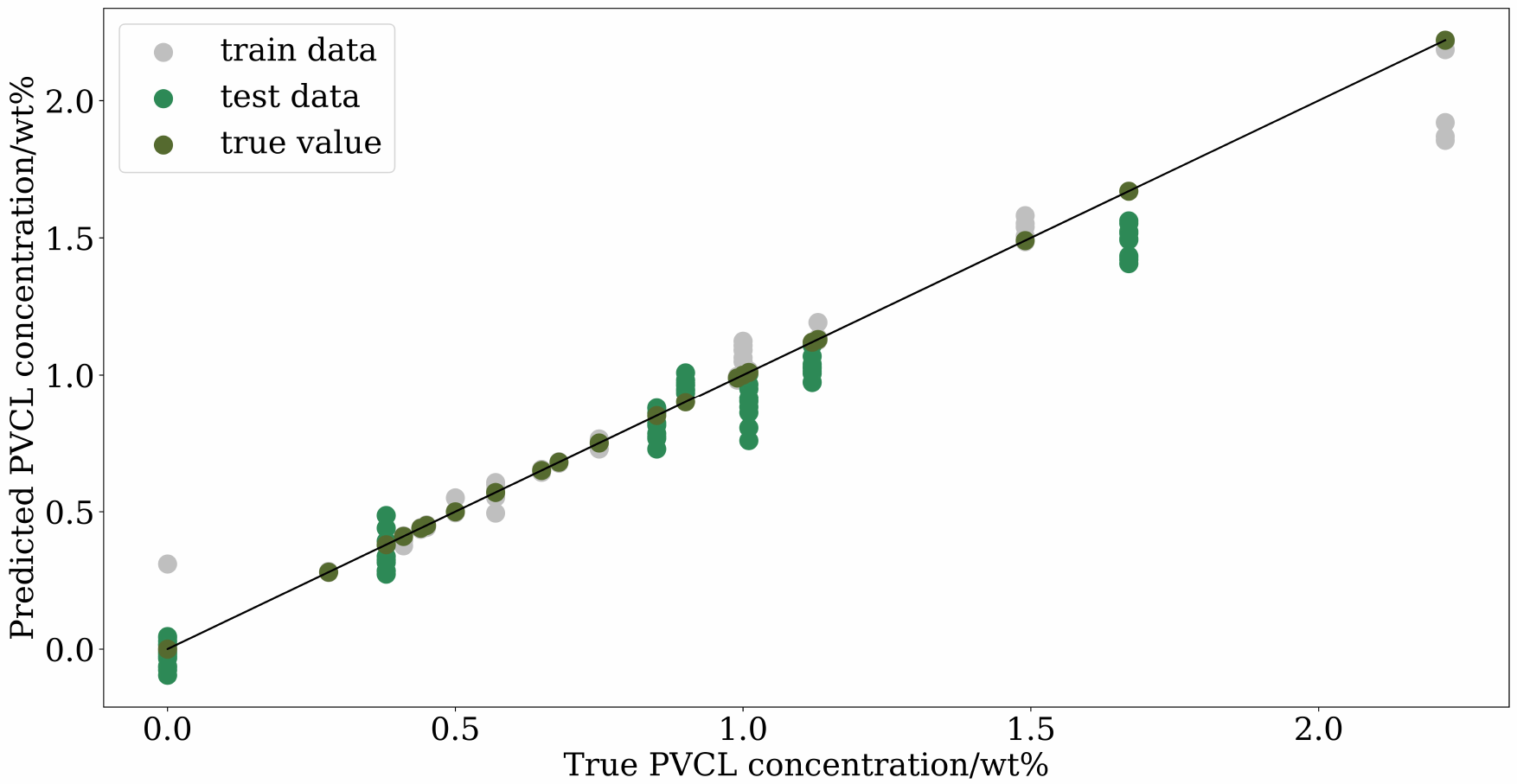}
    \caption{}
    \label{fig:UQ_ANN_pol_case 1}
    \end{subfigure}
    \caption{Uncertainty quantification study plots from the separate target prediction with ANN for the first case study, for water (a), monomer (b), and polymer (c). There is a higher discrepancy for the predicted test datapoints compared to the predicted train datapoints, because the test datapoints are new data for the models. The predictions from the ten models are on the same range.}
    \label{fig:UQ_ANN_case 1}
\end{figure}

\begin{figure}[ht] 
    \centering
    \begin{subfigure}{0.4\textwidth}
    \includegraphics[width=1\linewidth]{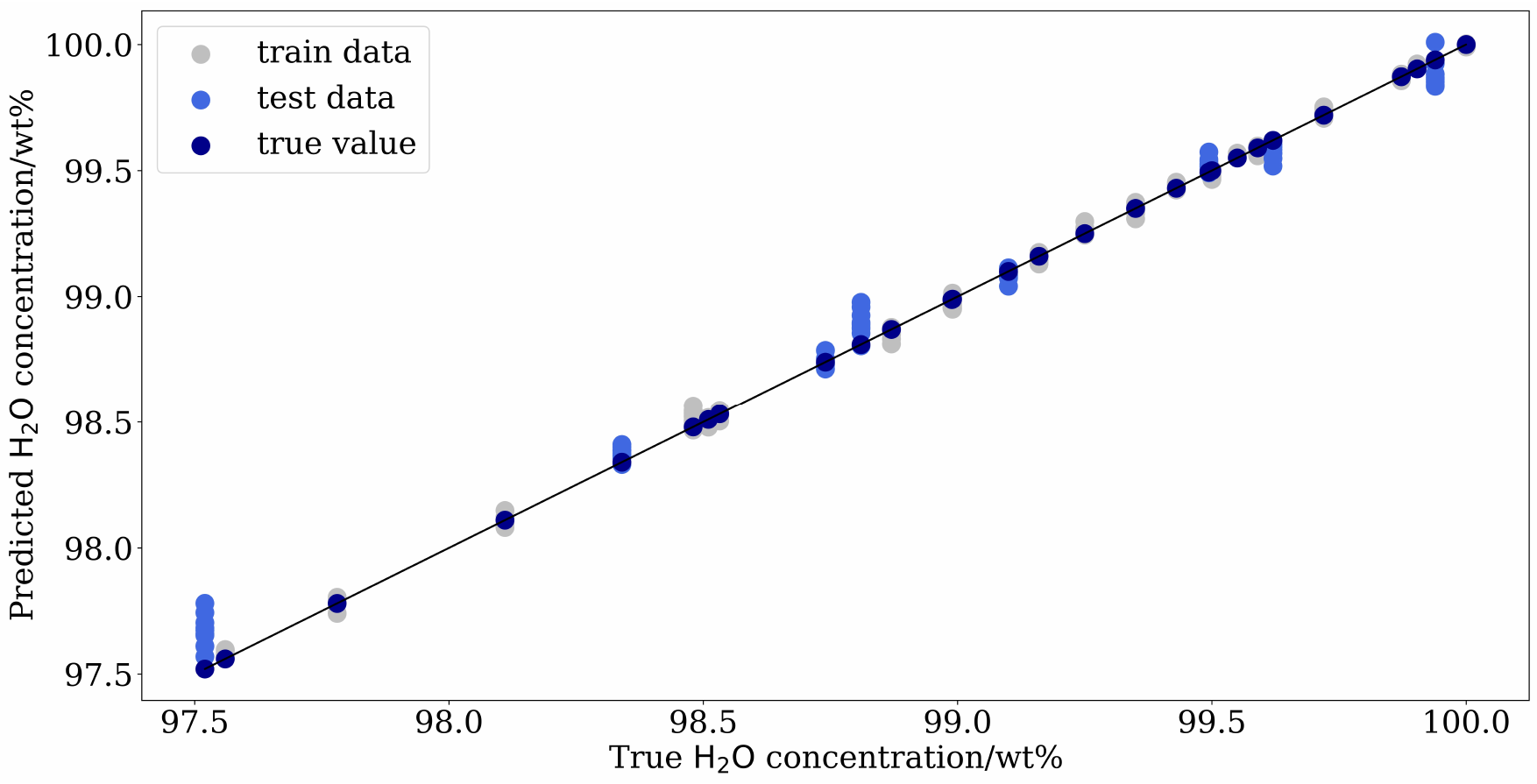}
    \caption{}
    \label{fig:UQ_sim_Y-AE_H2O_case 1}
    \end{subfigure}
    \hspace{0.1\textwidth}  
    \vspace{0.05\textwidth}
    \begin{subfigure}{0.4\textwidth}
    \includegraphics[width=1\linewidth]{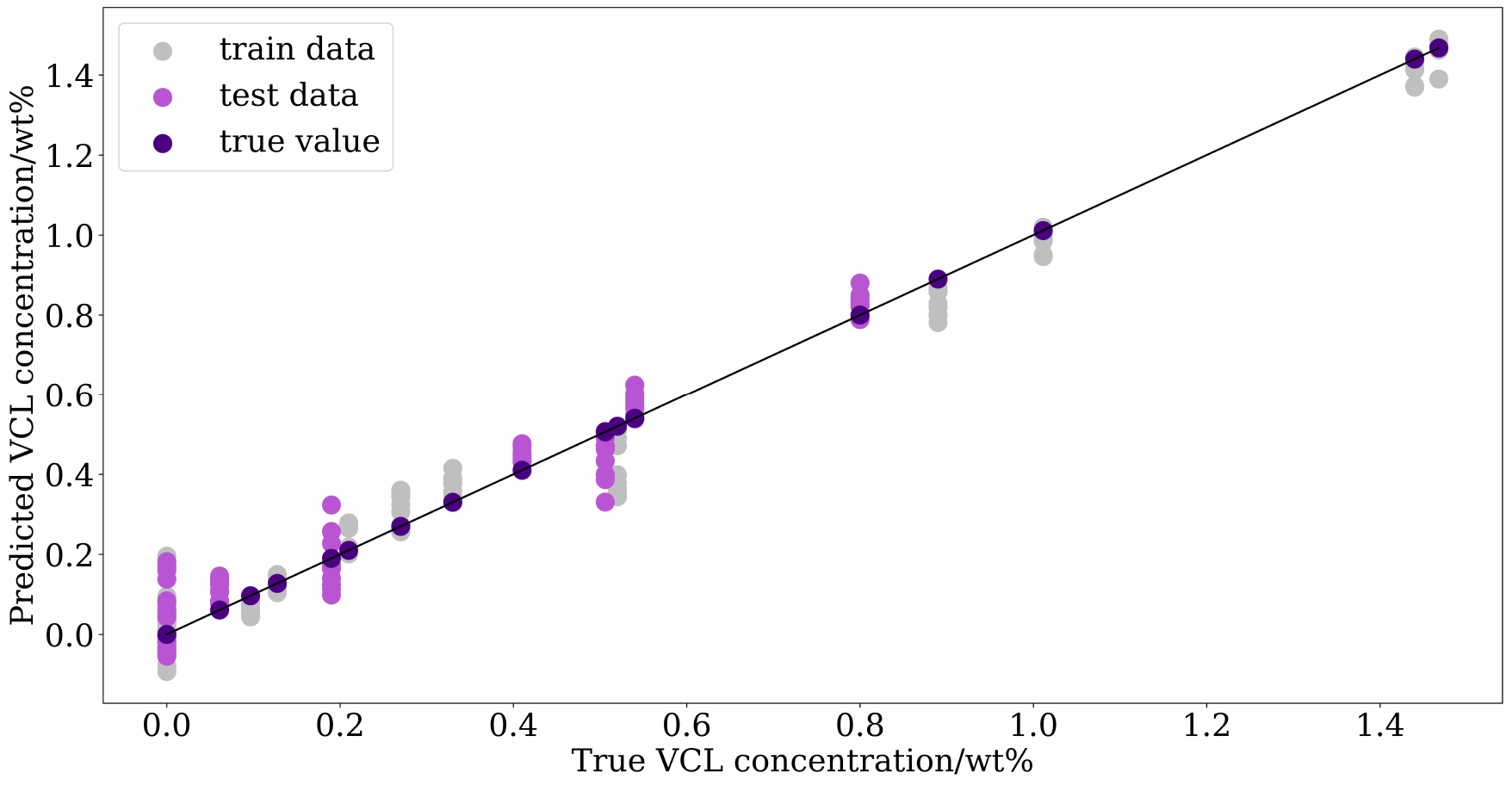}
    \caption{}
    \label{fig:UQ_sim_Y-AE_mon_case 1}
    \end{subfigure}
    \begin{subfigure}{0.4\textwidth}
    \includegraphics[width=1\linewidth]{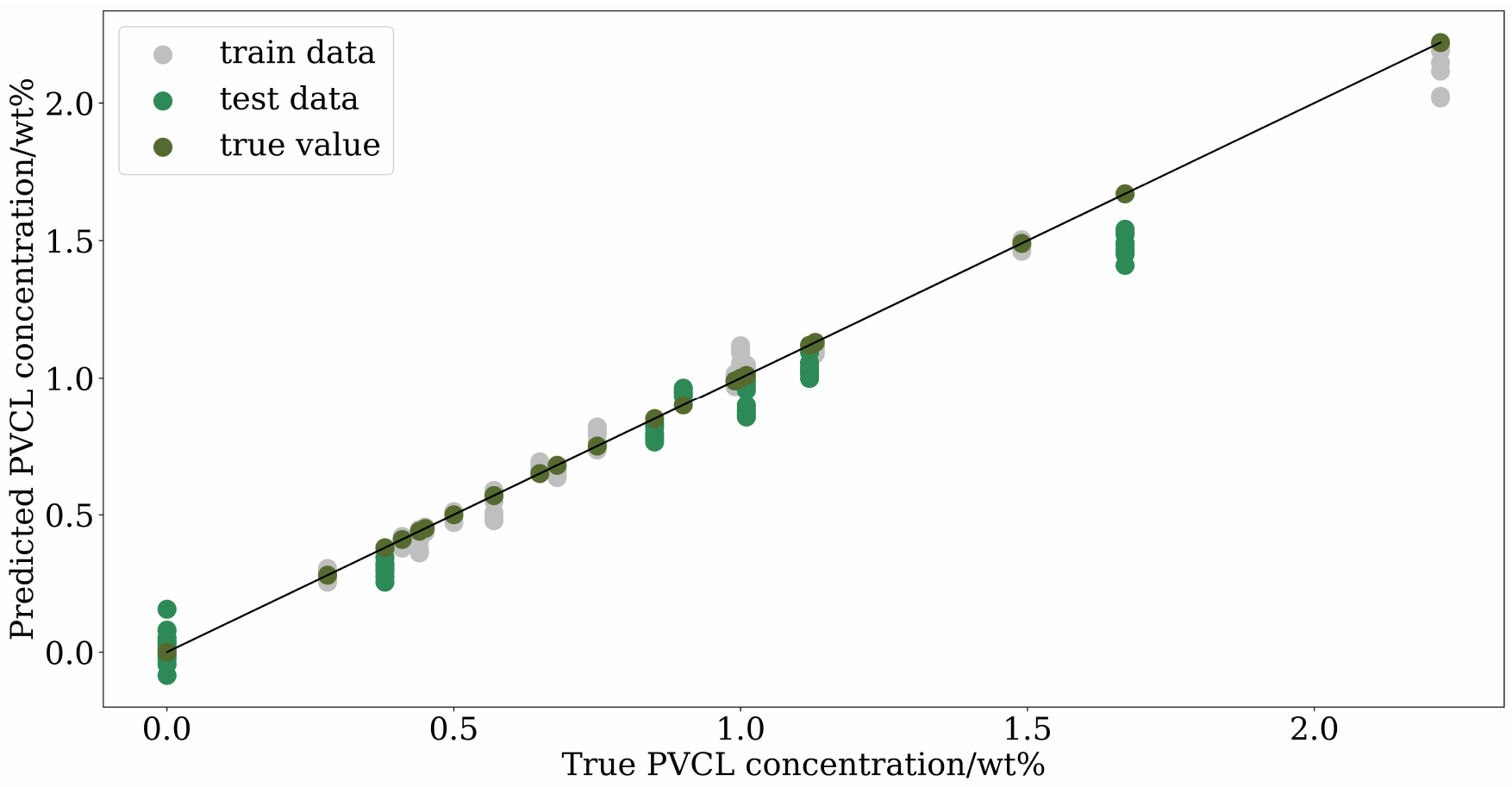}
    \caption{}
    \label{fig:UQ_sim_Y-AE_pol_case 1}
    \end{subfigure}
    \caption{Uncertainty quantification study plots from the simultaneous target prediction with Y-AE for the first case study, for water (a), monomer (b), and polymer (c). There is a higher discrepancy for the predicted test datapoints compared to the predicted train datapoints, because the test datapoints are new data for the models. The predictions from the ten models are on the same range.}
    \label{fig:UQ_simultaneous Y-AE_case 1}
\end{figure}

We do an uncertainty quantification study on the separate target predictions from Y-AE and ANN, and on the simultaneous target predictions from Y-AE, by training ten models for each method. For the simultaneous target predictions we consider the model from the second approach, predicting three targets from two latent variables. The parity plots obtained from the UQ study on the separate predictions from Y-AE for each target are shown in Fig.~\ref{fig:UQ_sep Y-AE_case 1}. The parity plots for the UQ study on the separate predictions from ANN and on the simultaneous predictions from Y-AE for each target are shown in Fig.~\ref{fig:UQ_ANN_case 1} and in Fig.~\ref{fig:UQ_simultaneous Y-AE_case 1} respectively.

We see that for all the methods, for all the targets we generally have higher discrepancy for the predicted test datapoints compared to the predicted train datapoints. This is because the test datapoints are new data for the models.
We also see that for each datapoint the predictions have generally a positive or a negative error, meaning that for each sample some of the predictions are above or below the true value. These predictions are biased. For some samples we get predictions with both positive and negative errors, meaning that these predictions are centered and unbiased.
When observing the parity plots for the separate predictions (Fig.~\ref{fig:UQ_sep Y-AE_case 1}) and the simultaneous predictions with Y-AE (Fig.~\ref{fig:UQ_simultaneous Y-AE_case 1}), we see that the behavior is roughly the same, meaning that the multi-target prediction with Y-AE is possible and of a similar accuracy with the single-target prediction. 
Overall, the predictions from ten models for each method are consistent and on the same range with each other and with the predictions presented on the results Section on the manuscript.

\subsection{Case study 2: citric acid dissociation data}\label{SI_UQ_case_2}
\begin{figure}[p] 
    \centering
    \begin{subfigure}{0.4\textwidth}
    \includegraphics[width=1\linewidth]{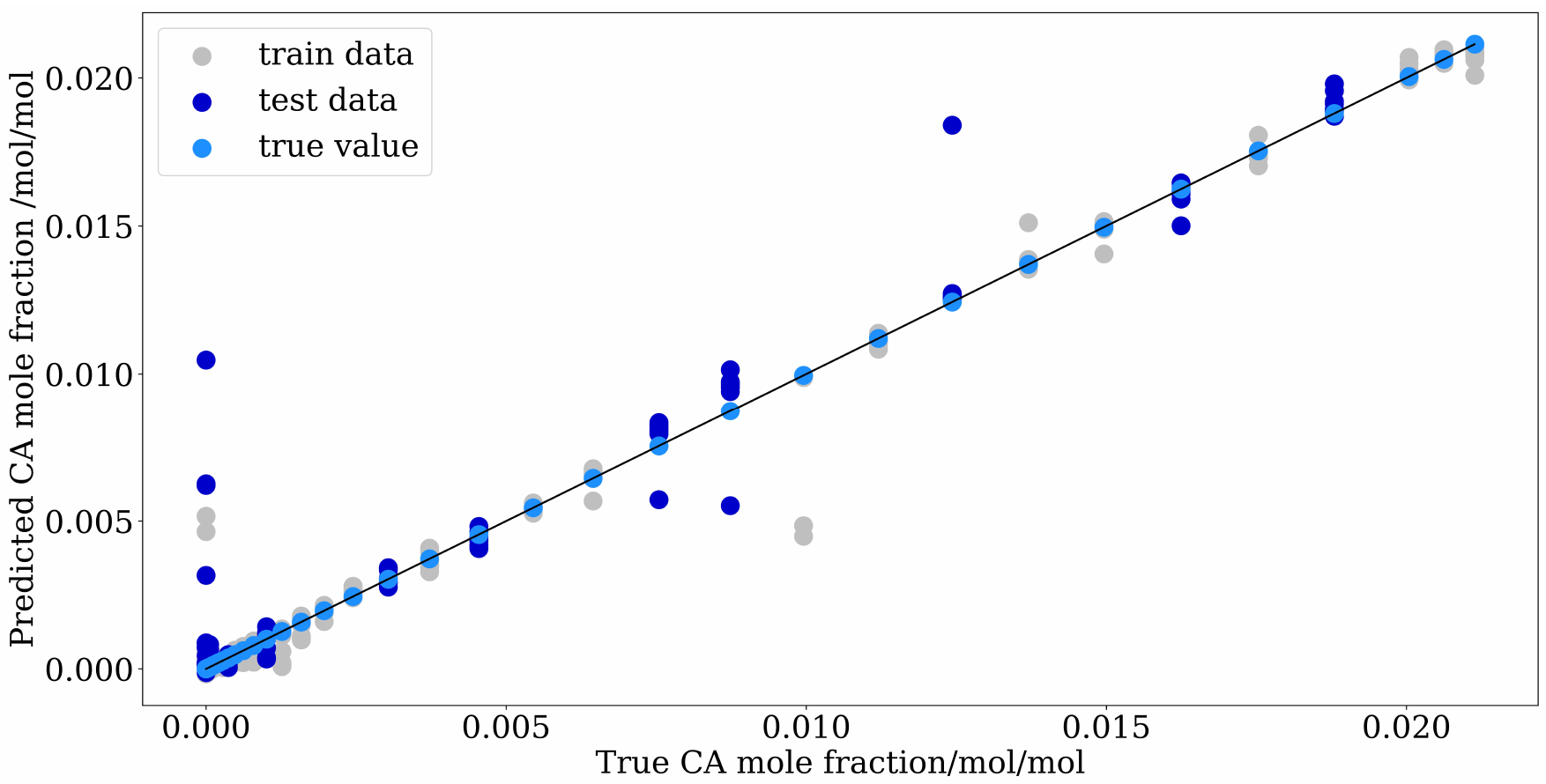}
    \caption{}
    \label{fig:UQ_sep_Y-AE_CA_case 2}
    \end{subfigure}
    \hspace{0.1\textwidth}  
    \vspace{0.05\textwidth}
    \begin{subfigure}{0.4\textwidth}
    \includegraphics[width=1\linewidth]{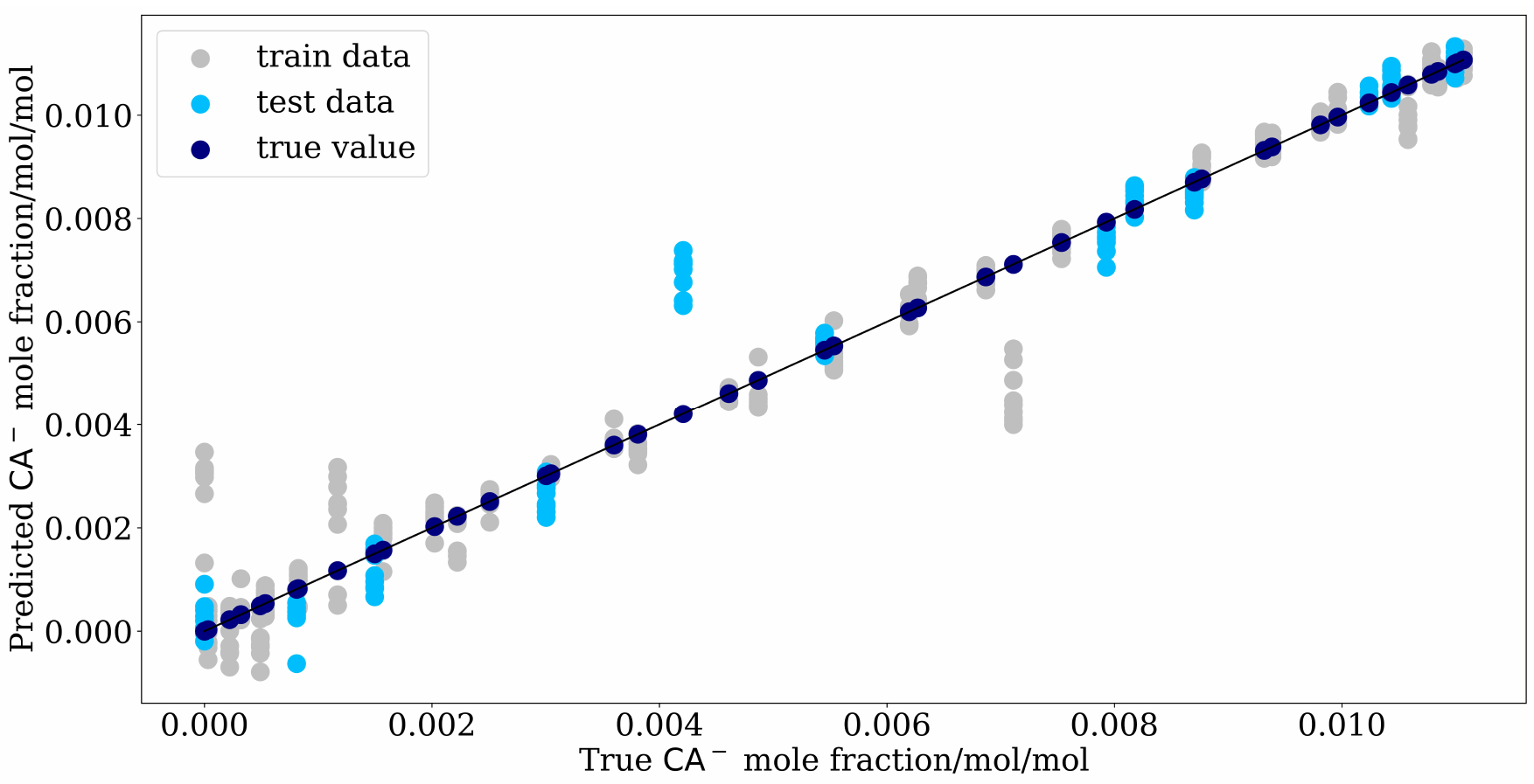}
    \caption{}
    \label{fig:UQ_sep_Y-AE_CA-_case 2}
    \end{subfigure}
    \begin{subfigure}{0.4\textwidth}
    \includegraphics[width=1\linewidth]{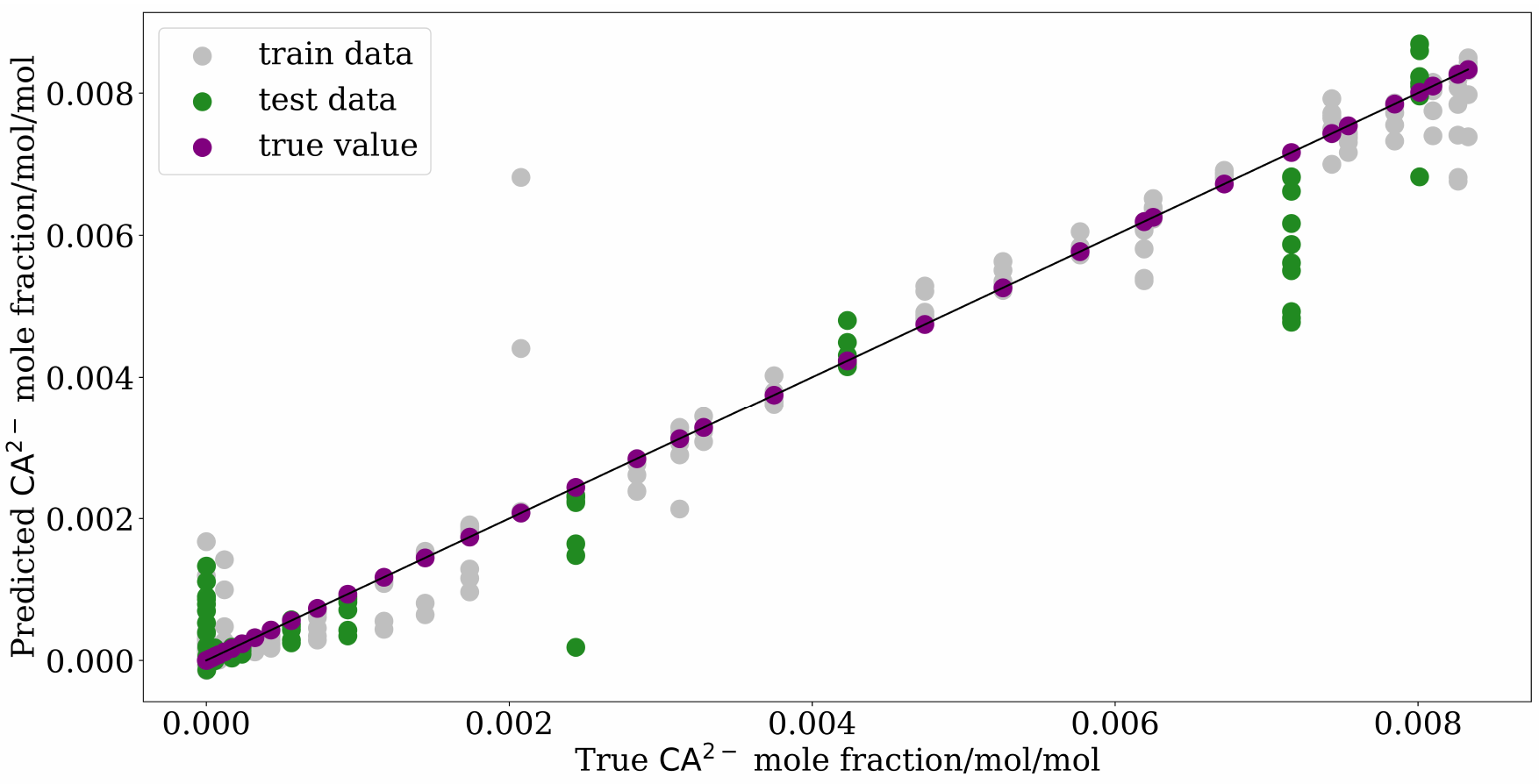}
    \caption{}
    \label{fig:UQ_sep_Y-AE_CA2-_case 2}
    \end{subfigure}
    \hspace{0.1\textwidth}  
    \vspace{0.05\textwidth}
    \begin{subfigure}{0.4\textwidth}
    \includegraphics[width=1\linewidth]{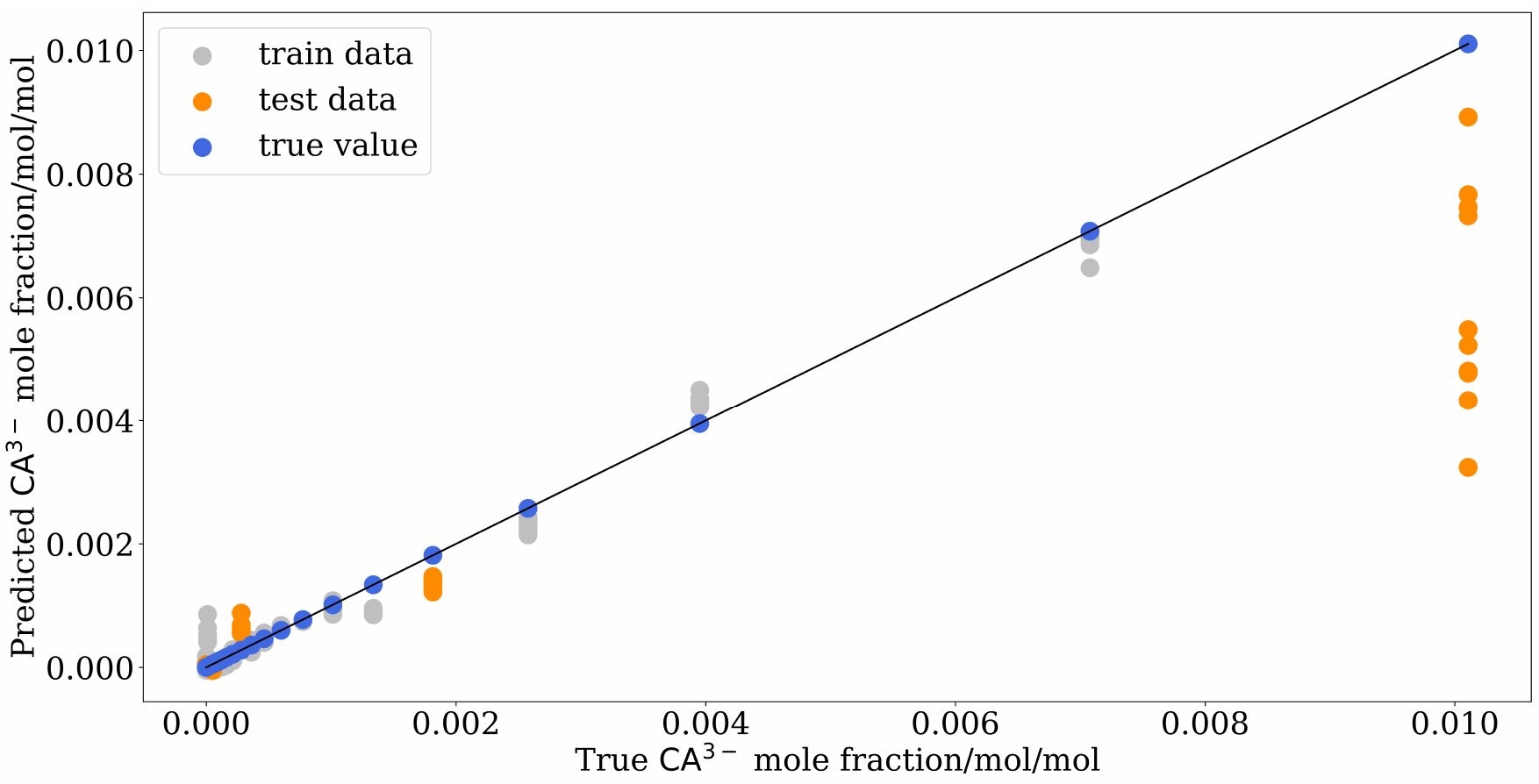}
    \caption{}
    \label{fig:UQ_sep_Y-AE_CA3-_case 2}
    \end{subfigure}
    \begin{subfigure}{0.4\textwidth}
    \includegraphics[width=1\linewidth]{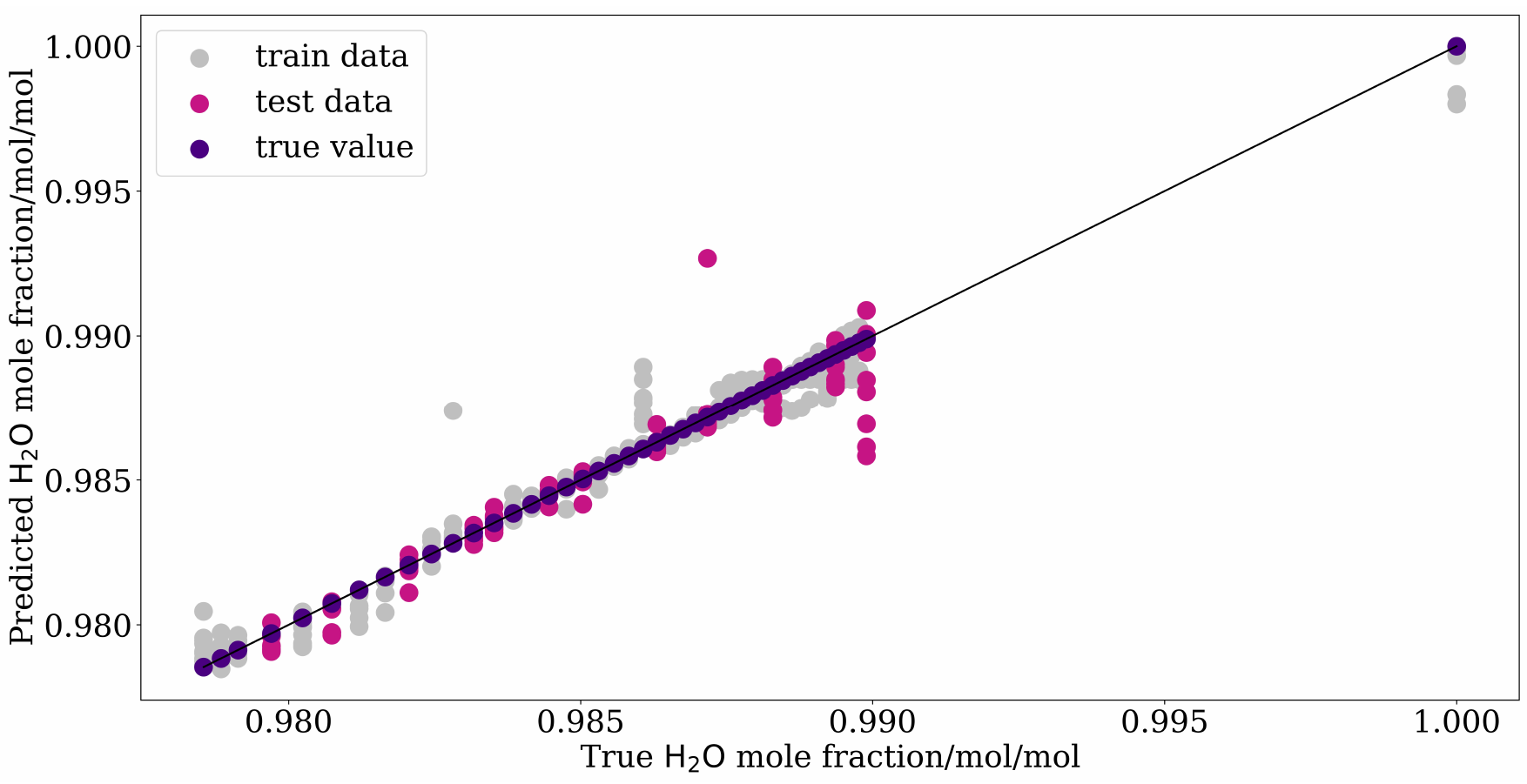}
    \caption{}
    \label{fig:UQ_sep_Y-AE_water_case 2}
    \end{subfigure}
    \caption{Uncertainty quantification study plots from the separate target prediction with Y-AE for the second case study, for citric acid (a), CA \(^{-}\) (b), CA \(^{2-}\) (c), CA \(^{3-}\) (d), and water (e). There is a higher discrepancy for the predicted test datapoints compared to the predicted train datapoints, because the test datapoints are new data for the models.}
    \label{fig:UQ_sep Y-AE_case 2}
\end{figure}

\begin{figure}[p] 
    \centering
    \begin{subfigure}{0.4\textwidth}
    \includegraphics[width=1\linewidth]{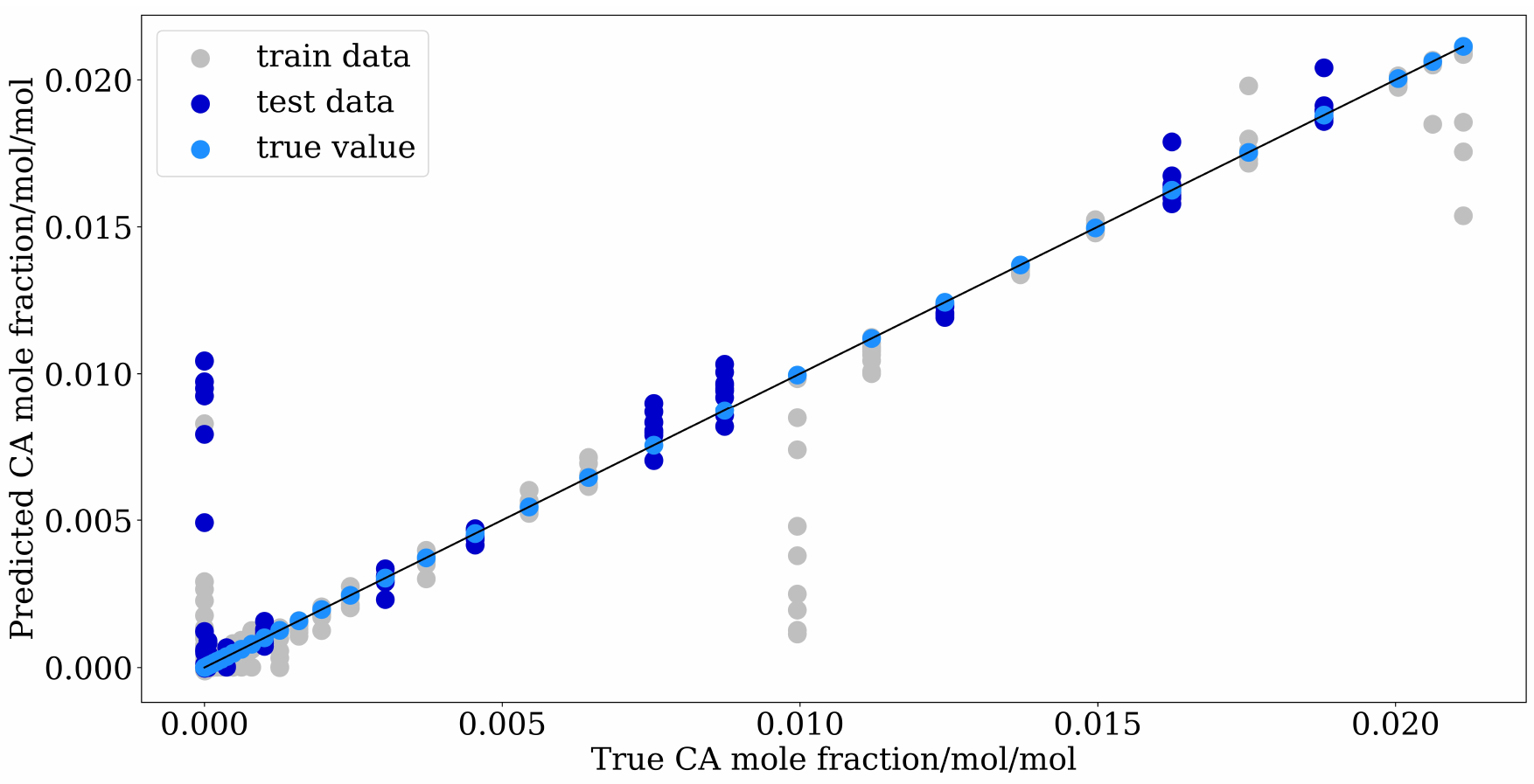}
    \caption{}
    \label{fig:UQ_ANN_CA_case 2}
    \end{subfigure}
    \hspace{0.1\textwidth}  
    \vspace{0.05\textwidth}
    \begin{subfigure}{0.4\textwidth}
    \includegraphics[width=1\linewidth]{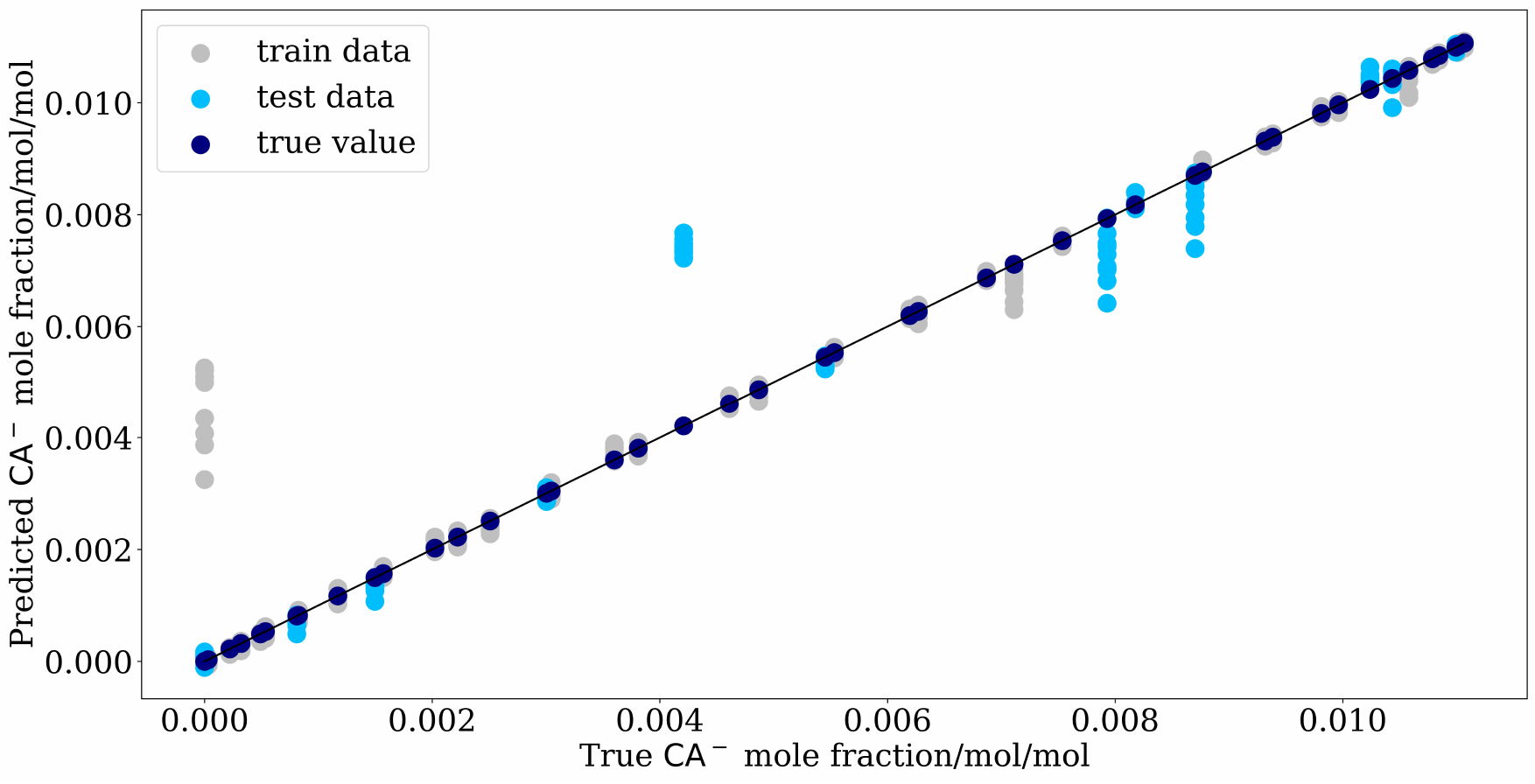}
    \caption{}
    \label{fig:UQ_ANN_CA-_case 2}
    \end{subfigure}
    \begin{subfigure}{0.4\textwidth}
    \includegraphics[width=1\linewidth]{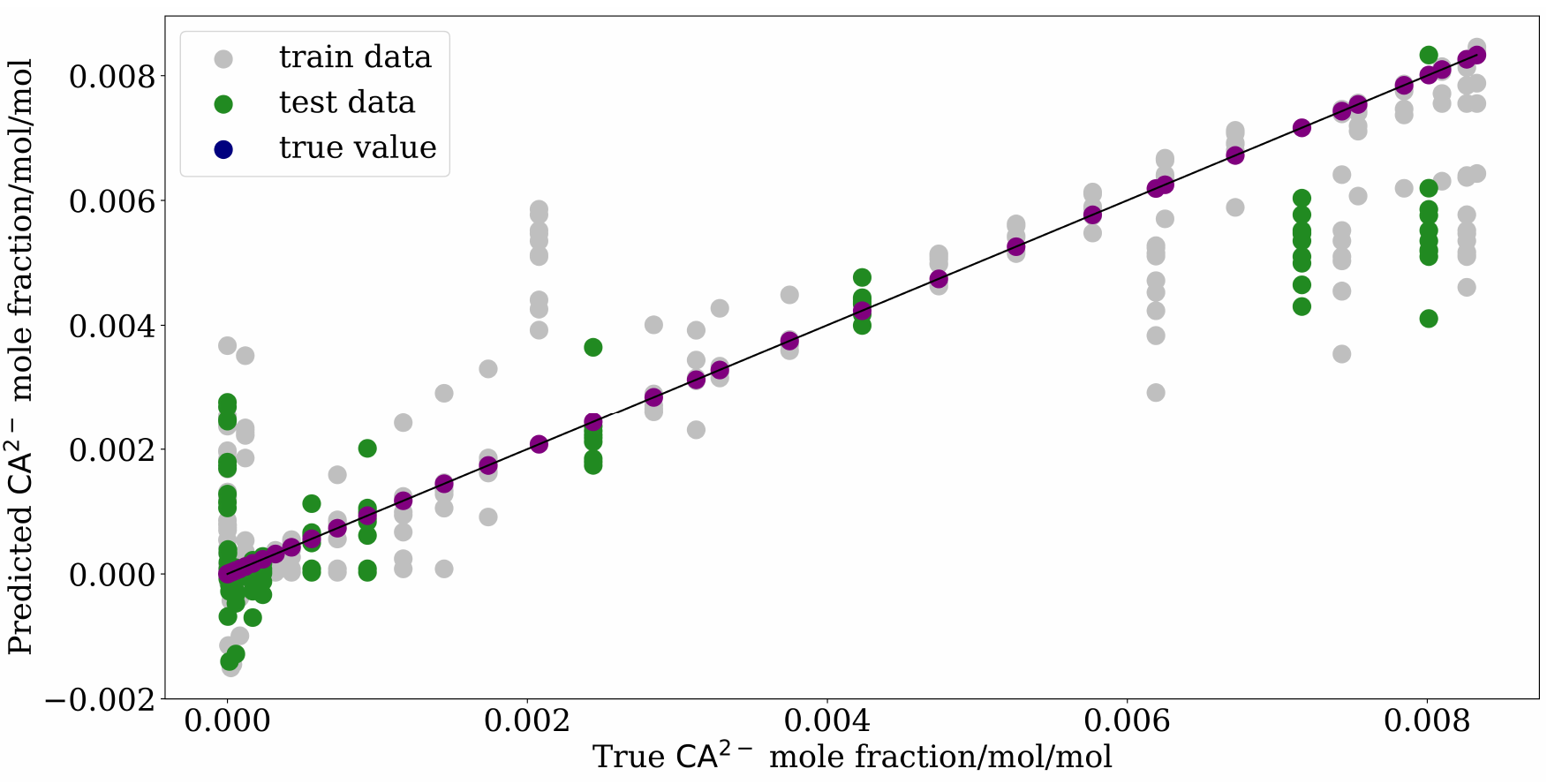}
    \caption{}
    \label{fig:UQ_ANN_CA2-_case 2}
    \end{subfigure}
    \hspace{0.1\textwidth}  
    \vspace{0.05\textwidth}
    \begin{subfigure}{0.4\textwidth}
    \includegraphics[width=1\linewidth]{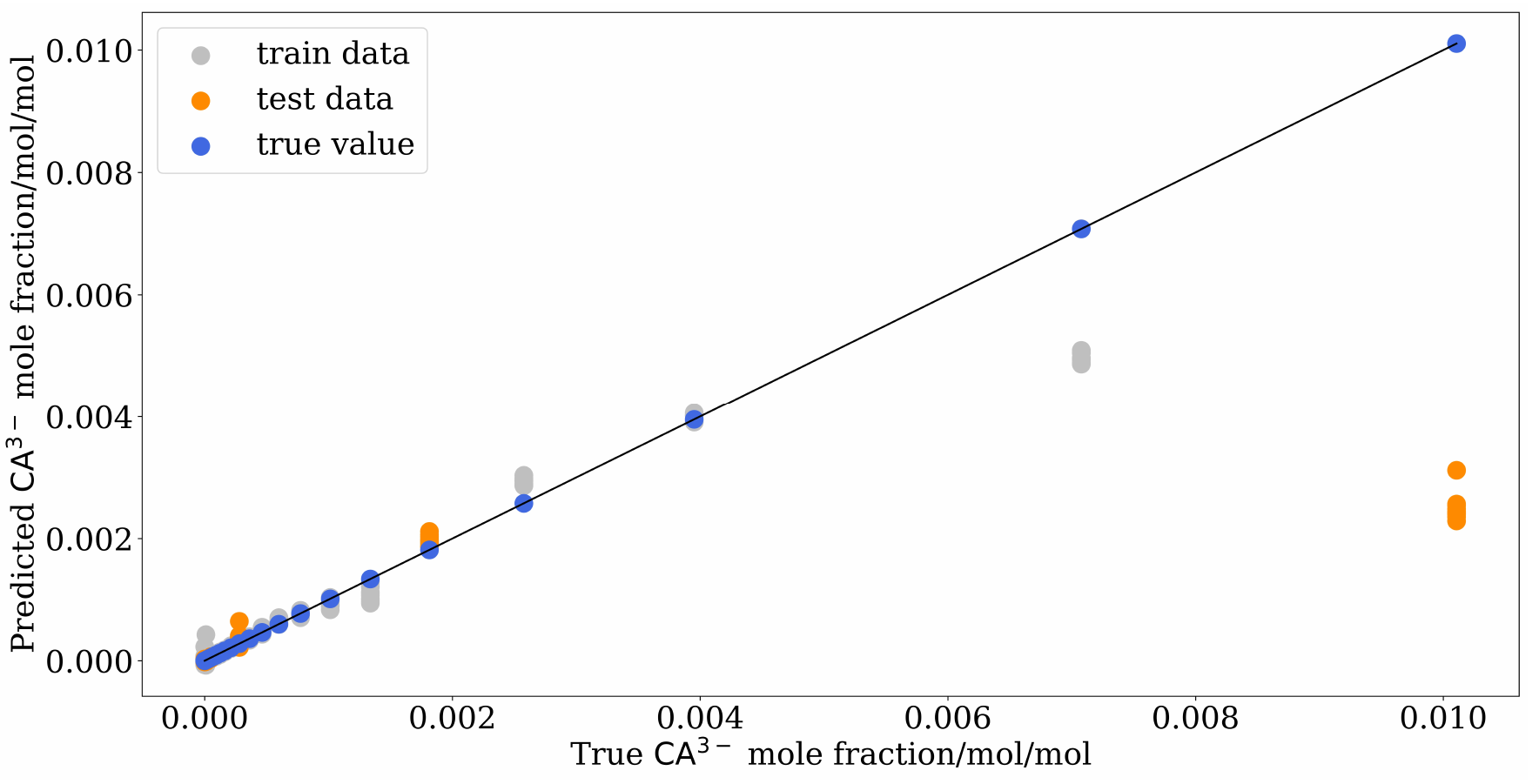}
    \caption{}
    \label{fig:UQ_ANN_CA3-_case 2}
    \end{subfigure}
    \begin{subfigure}{0.4\textwidth}
    \includegraphics[width=1\linewidth]{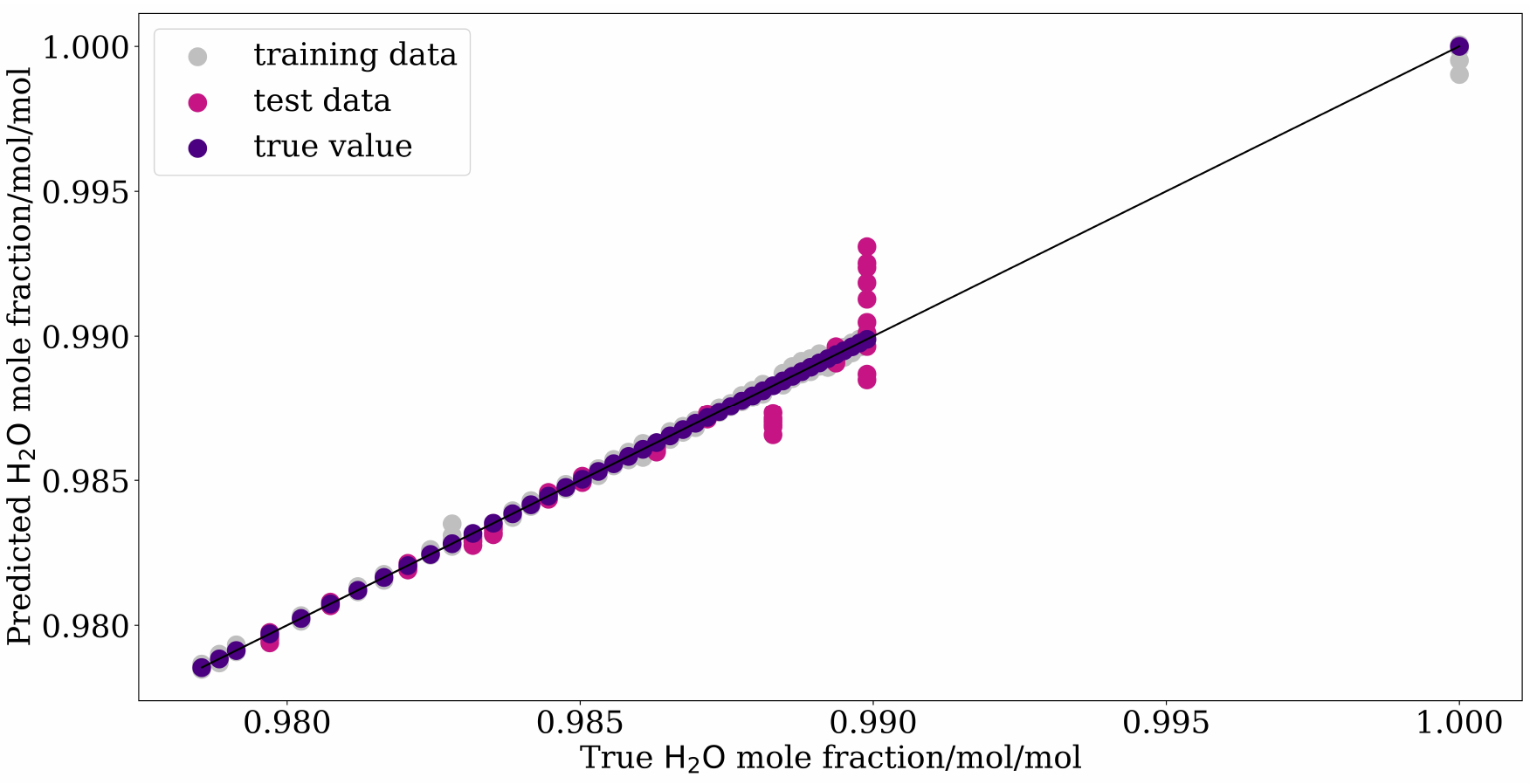}
    \caption{}
    \label{fig:UQ_ANN_water_case 2}
    \end{subfigure}
    \caption{Uncertainty quantification study plots from the separate target prediction with ANN for the second case study, for citric acid (a), CA \(^{-}\) (b), CA \(^{2-}\) (c), CA \(^{3-}\) (d), and water (e). There is a higher discrepancy for the predicted test datapoints compared to the predicted train datapoints, because the test datapoints are new data for the models.}
    \label{fig:UQ_ANN_case 2}
\end{figure}

\begin{figure}[p] 
    \centering
    \begin{subfigure}{0.4\textwidth}
    \includegraphics[width=1\linewidth]{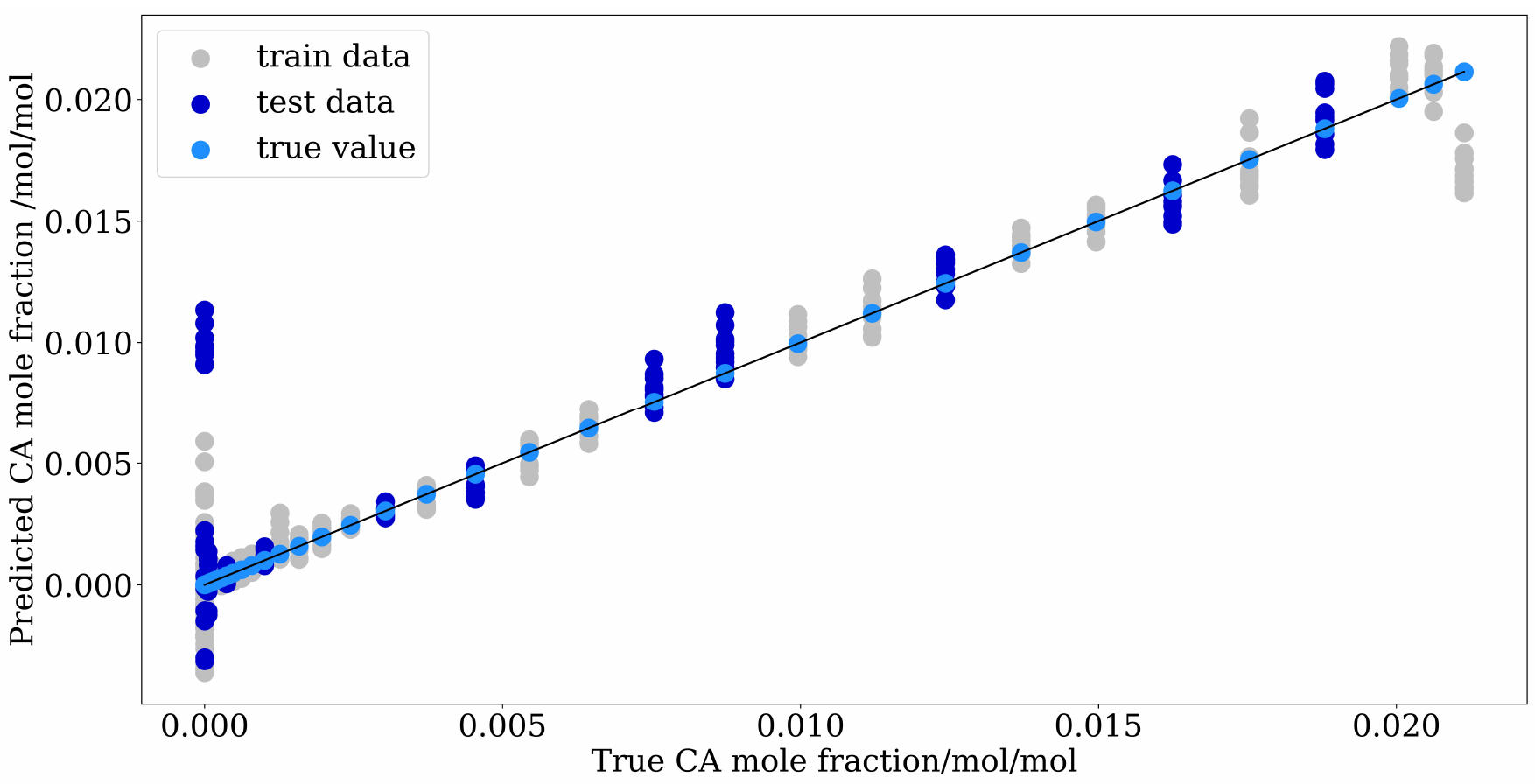}
    \caption{}
    \label{fig:UQ_sim_Y-AE_CA_case 2}
    \end{subfigure}
    \hspace{0.1\textwidth}  
    \vspace{0.05\textwidth}
    \begin{subfigure}{0.4\textwidth}
    \includegraphics[width=1\linewidth]{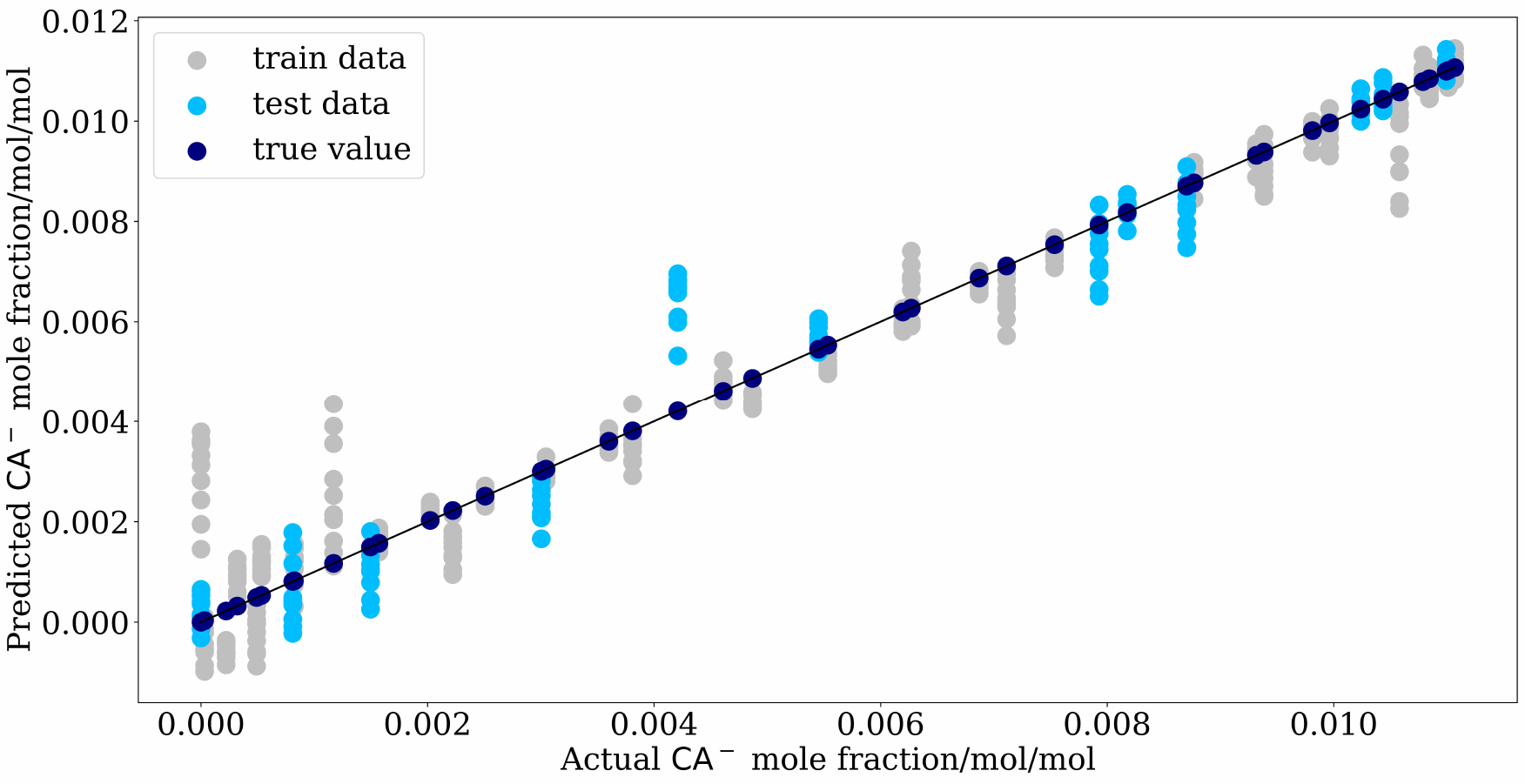}
    \caption{}
    \label{fig:UQ_sim_Y-AE_CA-_case 2}
    \end{subfigure}
    \begin{subfigure}{0.4\textwidth}
    \includegraphics[width=1\linewidth]{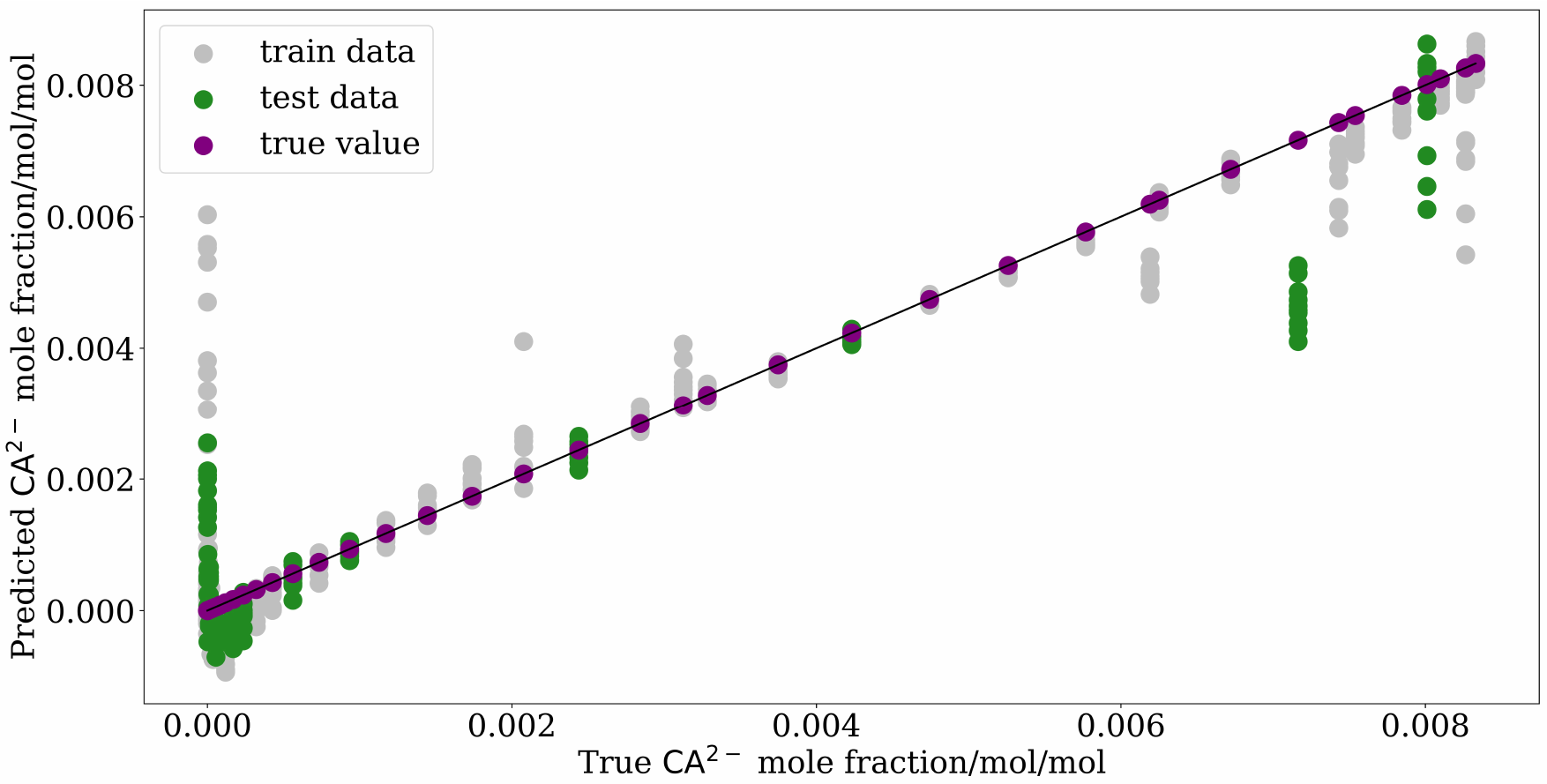}
    \caption{}
    \label{fig:UQ_sim_Y-AE_CA2-_case 2}
    \end{subfigure}
    \hspace{0.1\textwidth}  
    \vspace{0.05\textwidth}
    \begin{subfigure}{0.4\textwidth}
    \includegraphics[width=1\linewidth]{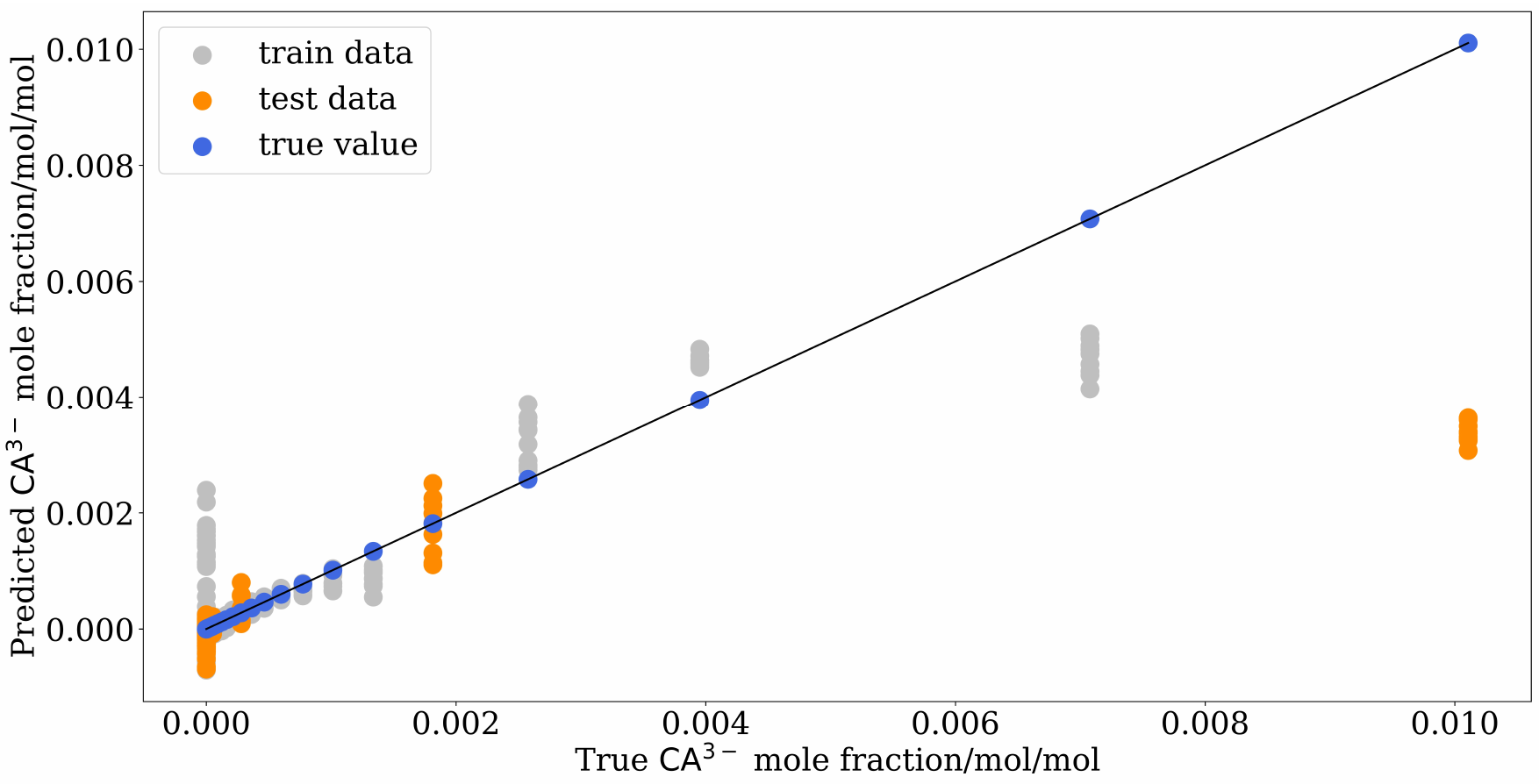}
    \caption{}
    \label{fig:UQ_sim_Y-AE_CA3-_case 2}
    \end{subfigure}
    \begin{subfigure}{0.4\textwidth}
    \includegraphics[width=1\linewidth]{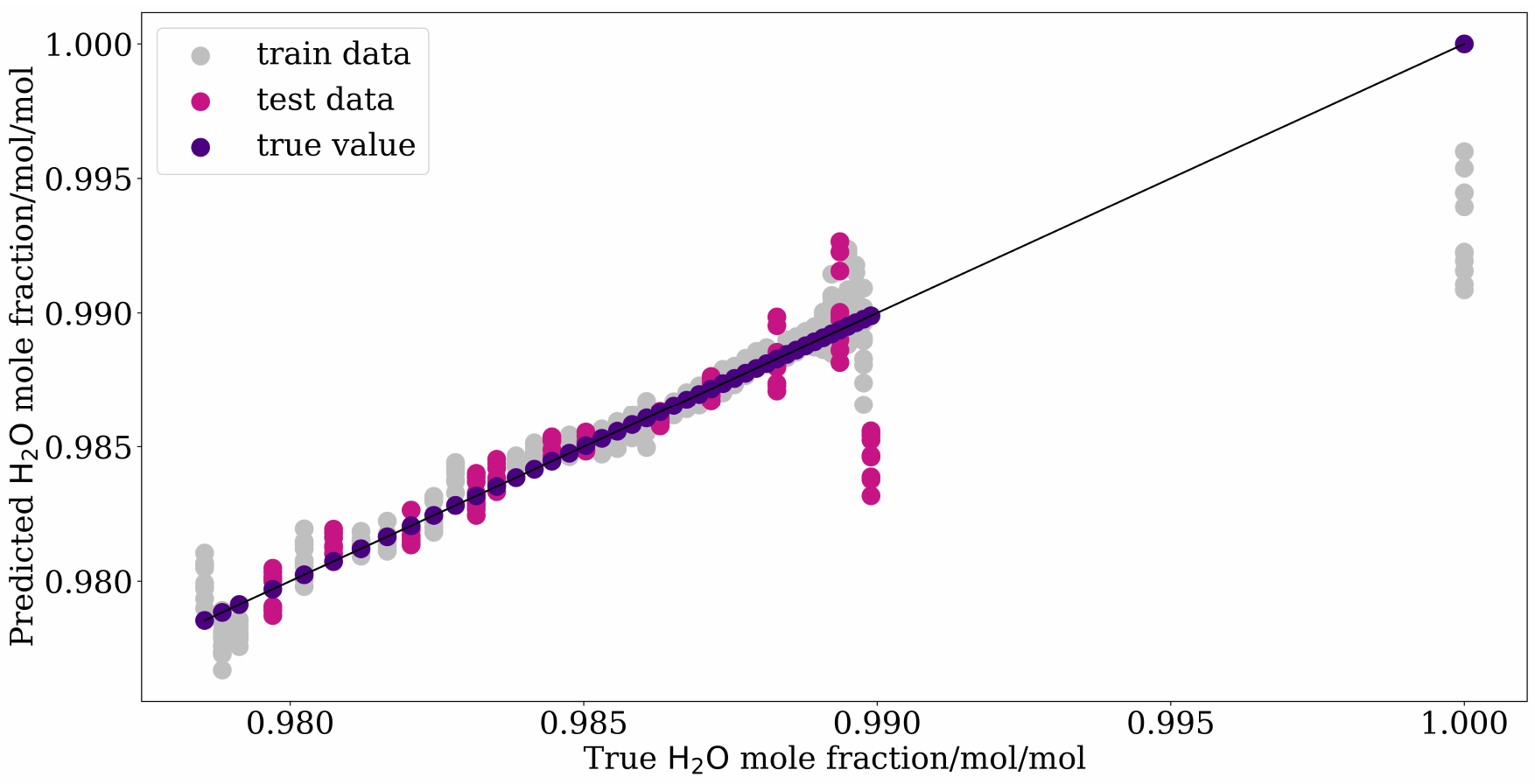}
    \caption{}
    \label{fig:UQ_sim_Y-AE_water_case 2}
    \end{subfigure}
    \caption{Uncertainty quantification study plots from the simultaneous target prediction with Y-AE for the second case study, for citric acid (a), CA \(^{-}\) (b), CA \(^{2-}\) (c), CA \(^{3-}\) (d), and water (e). There is a higher discrepancy for the predicted test datapoints compared to the predicted train datapoints, because the test datapoints are new data for the models.}
    \label{fig:UQ_simultaneous Y-AE_case 2}
\end{figure}

We do an uncertainty quantification study on the separate target predictions from Y-AE and ANN, and on the simultaneous target predictions from Y-AE, by training ten models for each method, also for the second case study. For the simultaneous target predictions we consider the model defined for the five-dimensional manifold, derived from the initial two-dimensional manifold, predicting five targets from five latent variables. The parity plots obtained from the UQ study on the separate predictions from Y-AE for each target are shown in Fig.~\ref{fig:UQ_sep Y-AE_case 2}. The parity plots for the UQ study on the separate predictions from ANN and on the simultaneous predictions from Y-AE for each target are shown in Fig.~\ref{fig:UQ_ANN_case 2} and in Fig.~\ref{fig:UQ_simultaneous Y-AE_case 2} respectively.

Similarly to the results of the UQ to the first case study, we see that there is a higher discrepancy for the predicted test datapoints than for the predicted train datapoints, even though the discrepancy for the predicted train datapoints for some of the targets is of a higher value in this case compared to what we see in the first case study. We also see that the parity plots for the single-target and the multi-target predictions from Y-AE have similar behavior, even though the single-target predictions are more accurate.

\section{DMAPs study: describing the low-dimensional manifold for both case studies} \label{SI_DMAPs study}
In the following, Section~\ref{SI_dmaps_case 1} and Section~\ref{SI_dmaps_case 2} present alternative representations of the underlying manifold of the Raman spectra for the first and the second case study respectively found with DMAPs. Section ~\ref{SI_dmaps_Nyström} presents the interpolation step with the Nyström extension using Geometric Harmonics.

\subsection{Case study 1: microgel synthesis data}\label{SI_dmaps_case 1}
\begin{figure} [H]
    \centering
   \includegraphics[width=0.5\linewidth]{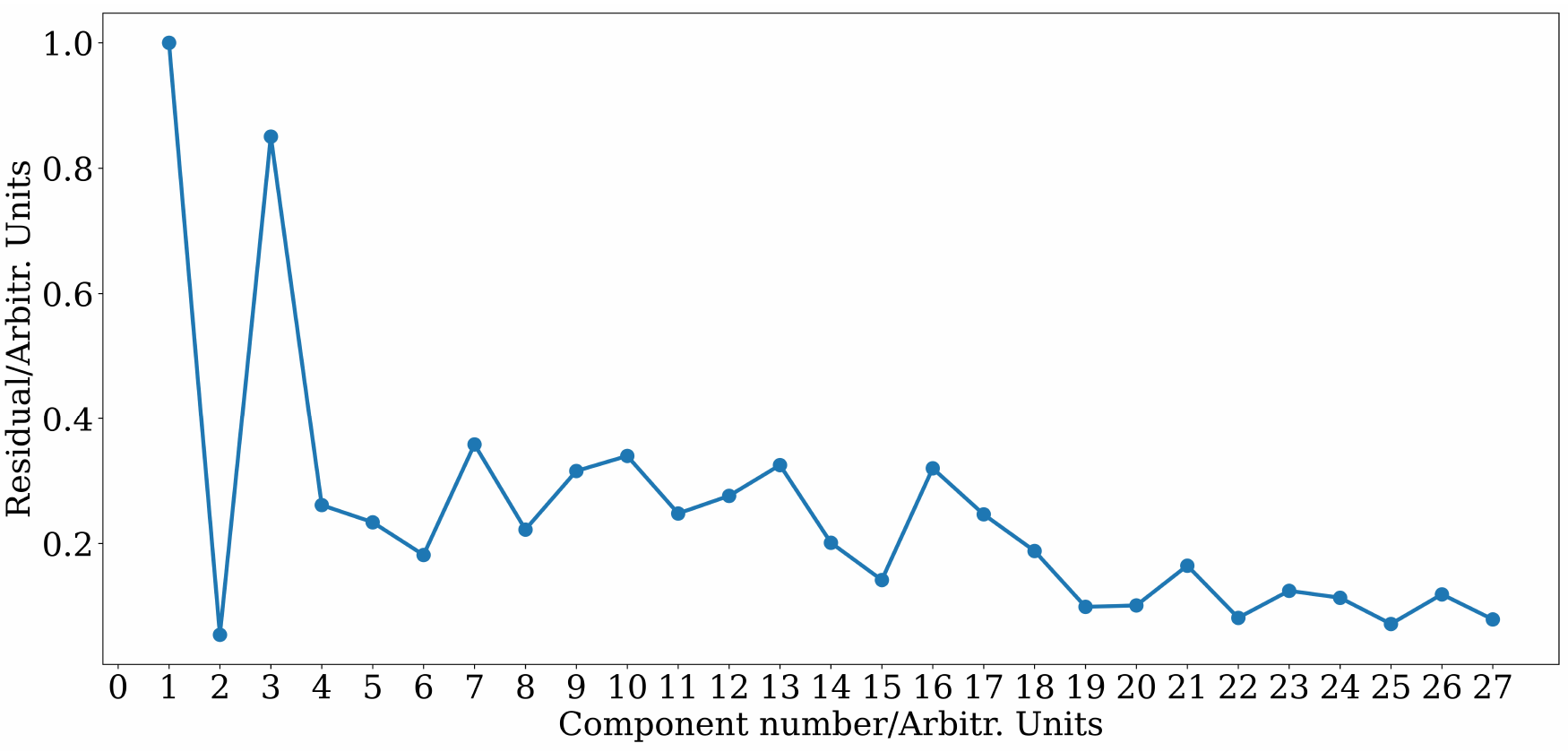}
    \caption{The residuals between the eigenvector sets after the local regression for the first case study. $\psi_1$ and $\psi_3$ have the highest residual values, followed by $\psi_7$, $\psi_{17} $ and so forth.}
    \label{fig:residuals_localregression_case 1}
\end{figure}

In order to describe the underlying manifold of the Raman spectra with DMAPs, we select the seven-dimensional representation shown in Fig.~\ref{fig:manifold_Raman_case 1} based on the decay of the eigenvalues of the matrix illustrated in Fig.~\ref{fig:scree_plot_case_1}, and the accuracy of the reconstruction of the original spectra, as described on the manuscript.
Besides the decay of the eigenvalues, we can also consider the residuals between the eigenvector sets after performing a local regression~\cite{LocalRegressionSelection_Dsilva2018}. We choose the eigenvectors with larger residuals, as they add more information. Fig.~\ref{fig:residuals_localregression_case 1} shows the residual values for the eigenvectors.

\begin{figure} [ht]
    \centering
    \includegraphics[width=0.5\linewidth]{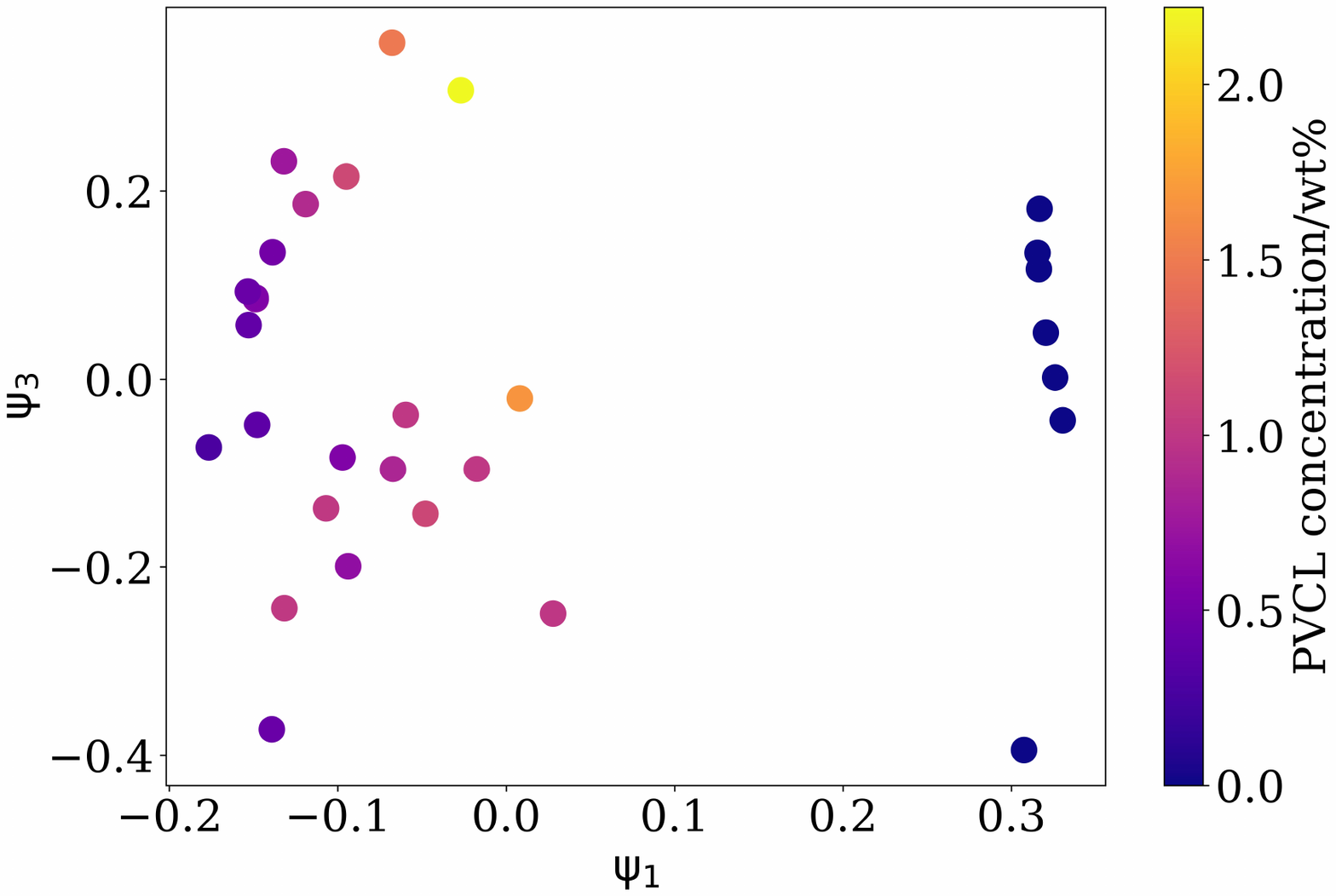}
    \caption{Low-dimensional underlying manifold based on the residuals between the eigenvector sets for the first case study. There are again two clusters, since the kernel scale is high, but the larger cluster is more spread than on the manifold where we considered the eigenvalue decay.}
    \label{fig:residuals manifold_localregression_case 1}
\end{figure}

From the figure we can see that $\psi_1$ and $\psi_3$ have the highest residual values, followed by $\psi_7$, $\psi_{17} $ and so forth. In order to choose how many from these eigenvectors to add in the embedding, we again consider the accuracy of the reconstruction of the original spectra, and we choose a two-dimensional manifold illustrated in Fig.~\ref{fig:residuals manifold_localregression_case 1}.
 
 We again see the two clusters, since the kernel scale is high (different than the previous one), but the larger cluster is more spread than on the parametrization where we considered the eigenvalue decay (Fig.~\ref{fig:manifold_Raman_case 1} on the manuscript). The reconstruction from this manifold however is less accurate, with the coefficient of determination for the test dataset being $R^2=0.8919$.

\subsection{Case study 2: citric acid dissociation data}\label{SI_dmaps_case 2}
Similarly to the first case study, we do the DMAPs analysis for the second case study and considering the decay of the eigenvalues illustrated in Fig.~\ref{fig:scree_plot_case_2}, we select the two-dimensional representation shown in Fig.~\ref{fig:manifold_Raman_case 2} on the manuscript.
For the simultaneous target predictions we define four more parametrizations of the manifold consisting of four, five and six dimensions, by adding more eigenvectors to the initially defined representation, and allowing for lower reconstruction accuracy.

\begin{figure} [H]
    \centering
    \includegraphics[width=0.5\linewidth]{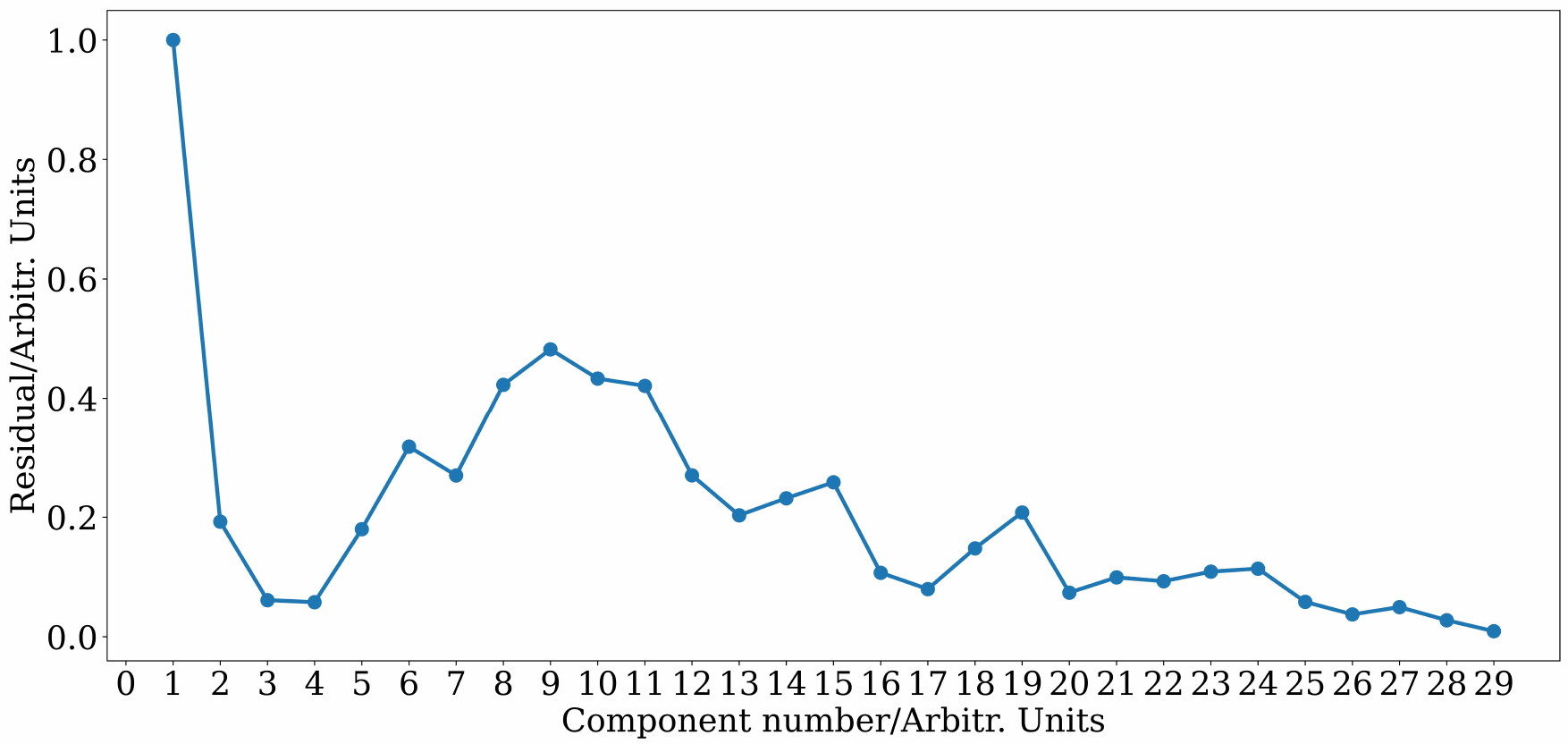}
    \caption{The residuals between the eigenvector sets after the local regression for the second case study. $\psi_1$ and $\psi_9$ have the highest residual values, followed by $\psi_8$, $\psi_{10} $ and so forth.}
    \label{fig:residuals_localregression_case 2}
\end{figure}

Since from the manifold in Fig.~\ref{fig:manifold_Raman_case 2} on the manuscript can be seen that the two eigenvectors might not be independent, we also consider the residuals between the eigenvector sets after performing the local regression and we get the residual values illustrated in Fig.~\ref{fig:residuals_localregression_case 2}.

\begin{figure} [H]
    \centering
    \includegraphics[width=0.5\linewidth]{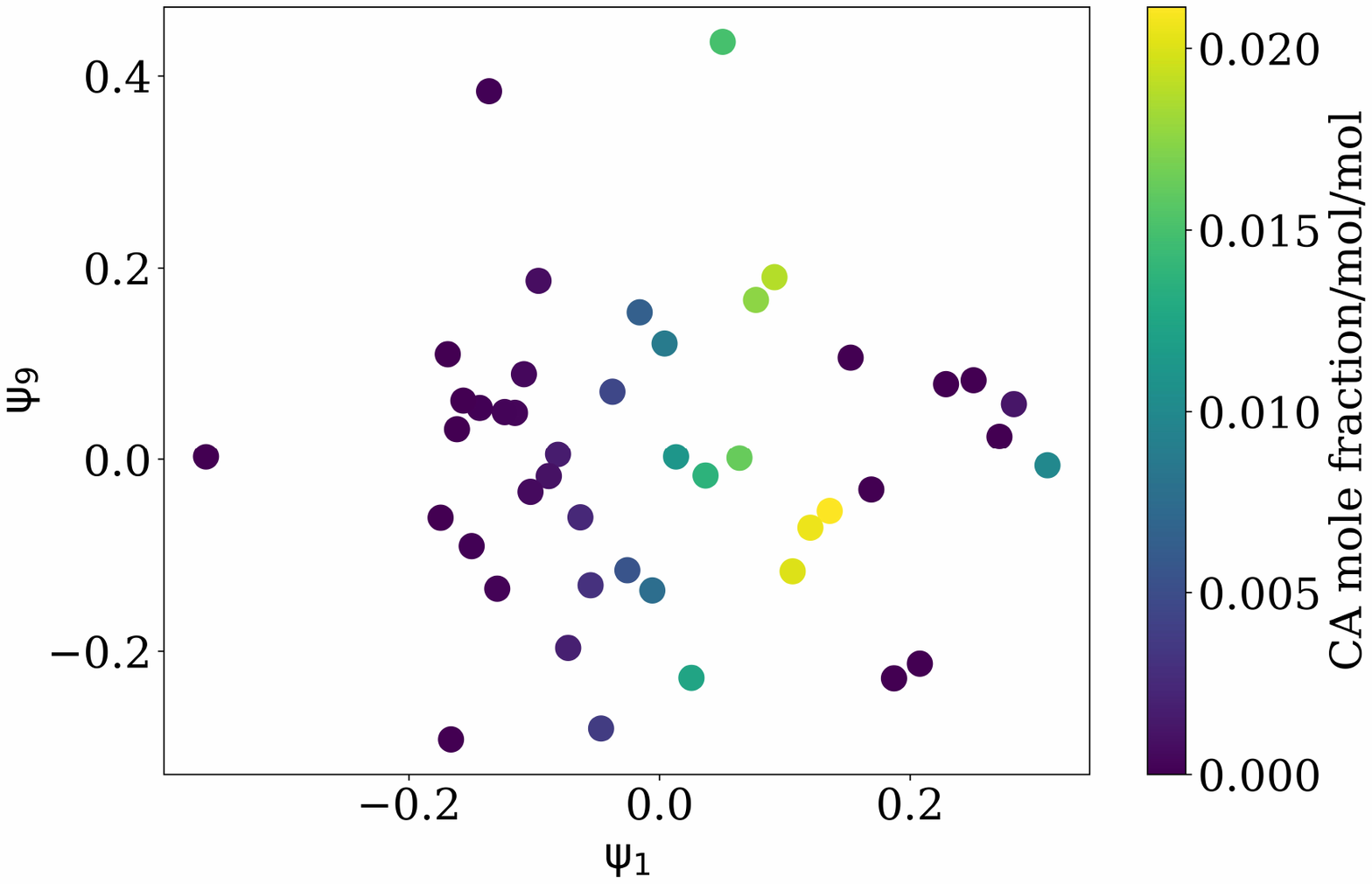}
    \caption{Low-dimensional underlying manifold based on the residuals between the eigenvector sets for the second case study. There are again two clusters, since the kernel scale is high, but the smaller cluster has only one datapoint and the larger cluster is more spread, and we cannot see a correlation between the two eigenvectors, as we see on the manifold where we considered the eigenvalue decay.}
    \label{fig:residuals manifold_localregression_case 2}
\end{figure}

From the figure we can see that $\psi_1$ and $\psi_9$ have the highest residual values, followed by $\psi_8$, $\psi_{10} $ and so forth. In order to choose how many from these eigenvectors to add in the embedding, we consider the accuracy of the reconstruction of the spectra, and we choose a two-dimensional manifold illustrated in Fig.~\ref{fig:residuals manifold_localregression_case 2}.

 We again see the two clusters, since the kernel scale is high (different than the previous one), but the smaller cluster has only one datapoint and the larger cluster is more spread, and we cannot see a correlation between the two eigenvectors, as we see on the parametrization where we considered the eigenvalue decay (Fig.~\ref{fig:manifold_Raman_case 2} on the manuscript). The reconstruction from this representation however is again less accurate, with the coefficient of determination for the test dataset being $R^2=0.7187$.
 
\begin{figure} [H]
    \centering
    \includegraphics[width=0.5\linewidth]{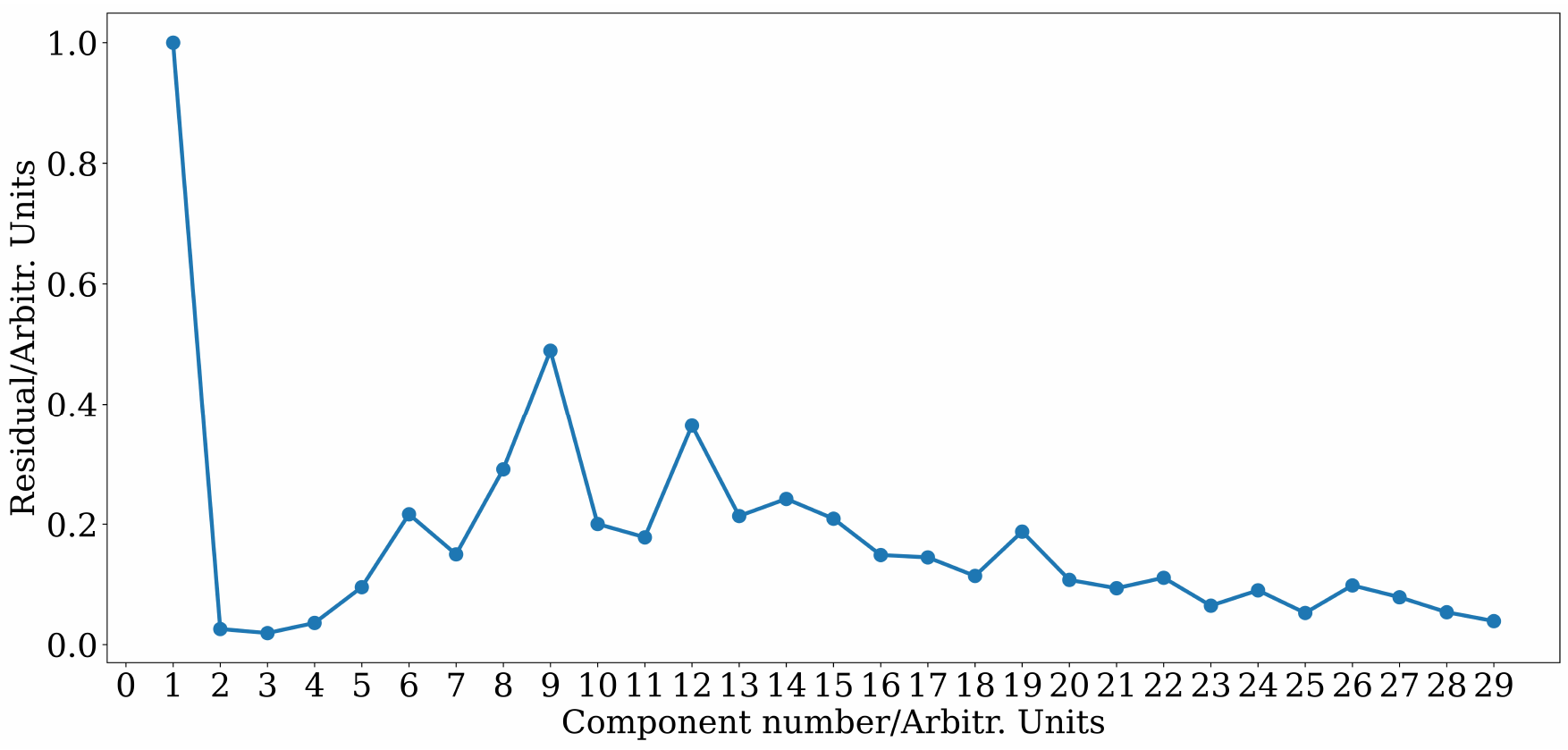}
    \caption{The residuals between the eigenvector sets after the local regression for the second case study with a different kernel scale. $\psi_1$ and $\psi_9$ have the highest residual values, followed by $\psi_8$, $\psi_{12} $ and so forth.}
    \label{fig:residuals_localregression_4D case 2}
\end{figure}

For the simultaneous target predictions, since the representation on Fig.~\ref{fig:residuals manifold_localregression_case 2} is again a two-dimensional parametrization, we need to define another one with at least four dimensions. By adding more eigenvectors to the already defined parametrization, we get even lower reconstruction of the spectra, with the values of $R^2=0.1042$ for four dimensions, $R^2=0.3417$ for five, and $R^2=0.3176$ for six dimensions. By further tuning the kernel scale we get the residual values shown on Fig.~\ref{fig:residuals_localregression_4D case 2}.

\begin{figure} [ht]
    \centering
    \includegraphics[width=0.5\linewidth]{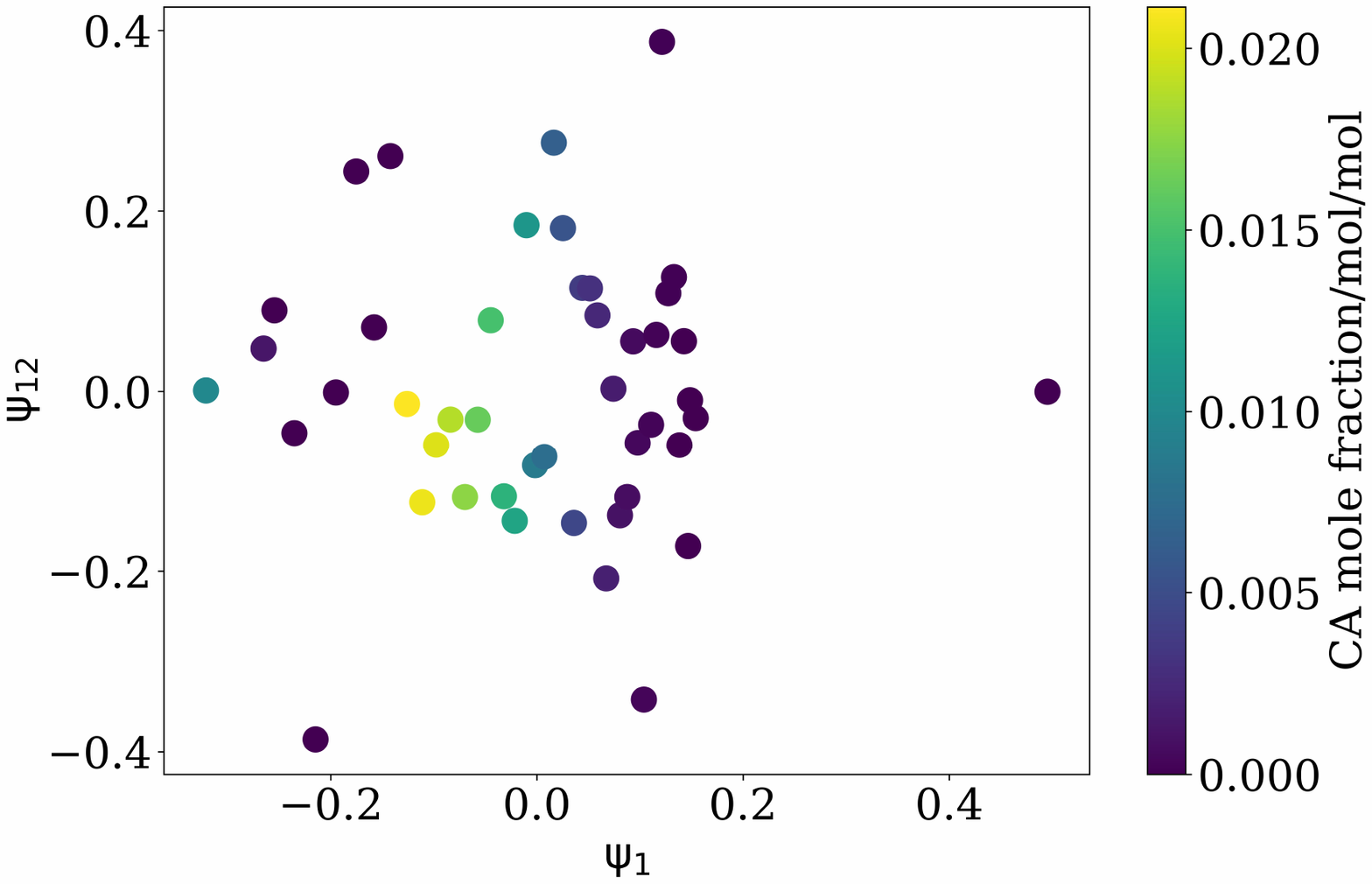}
    \caption{Four-dimensional parametrization of the manifold based on the residuals between the eigenvector sets for the second case study. There are two clusters, with the large cluster being less spread compared to the one in Fig.~\ref{fig:residuals manifold_localregression_case 2}, since the kernel scale here is lower.}
    \label{fig:residuals manifold_localregression_4D case 2}
\end{figure}

From Fig.~\ref{fig:residuals_localregression_4D case 2} we can see that $\psi_1$ and $\psi_9$ have the highest residual values, followed by $\psi_8$, $\psi_{12} $ and so forth. Based on the accuracy of the reconstruction of the original spectra, we choose a four-dimensional representation illustrated in Fig.~\ref{fig:residuals manifold_localregression_4D case 2}.

We see the two clusters, with the large cluster being less spread compared to the one in Fig.~\ref{fig:residuals manifold_localregression_case 2}, since the kernel scale here is lower. The reconstruction from this representation is again less accurate compared to the one in Fig.~\ref{fig:manifold_Raman_case 2} on the manuscript, with the coefficient of determination for the test dataset being $R^2=0.6901$.

\begin{figure} [ht]
    \centering
    \includegraphics[width=0.5\linewidth]{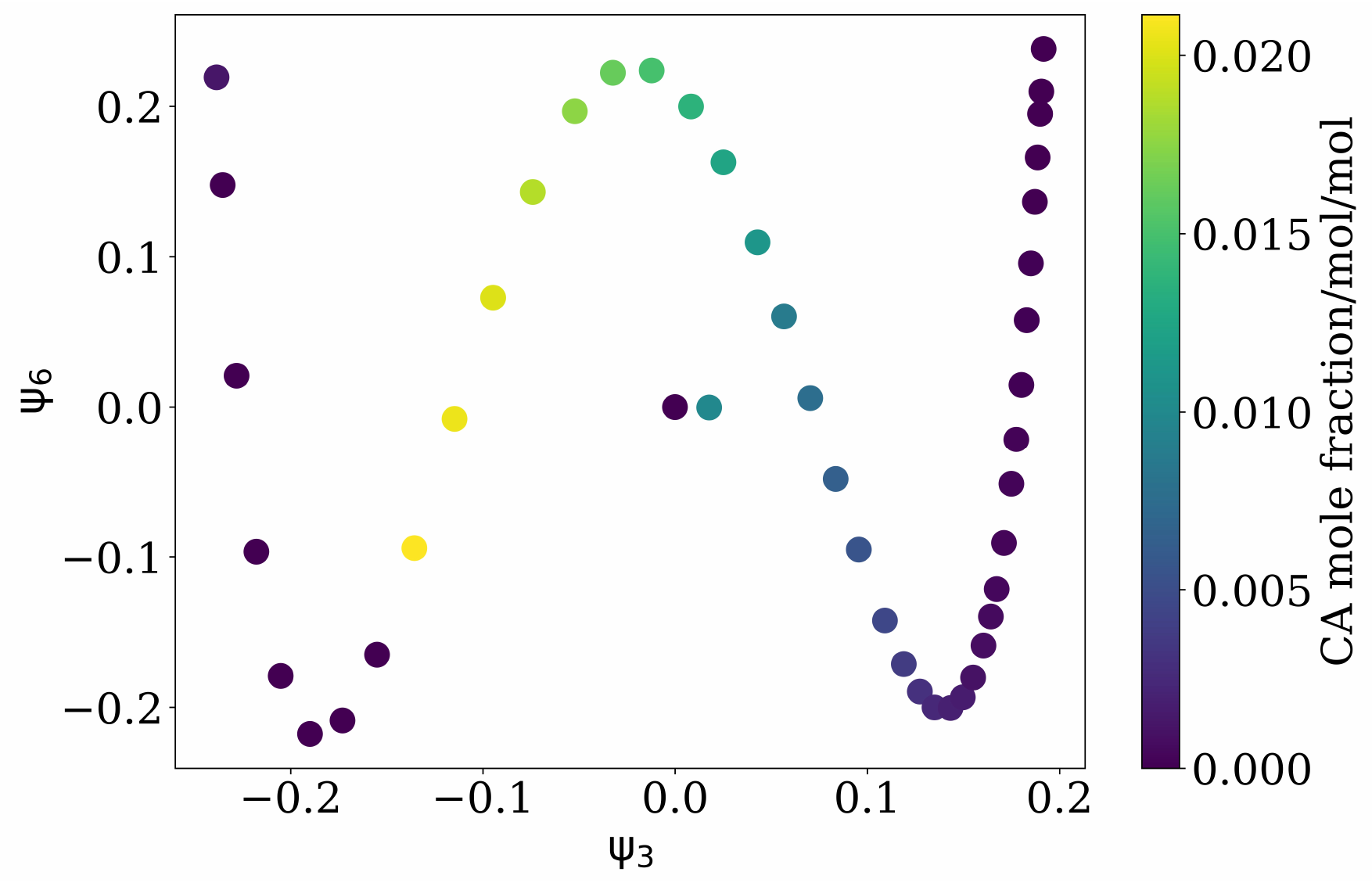}
    \caption{Five-dimensional parametrization of the underlying manifold based on the decay of the eigenvalues. This parametrization is the input on the Y-AE for the simultaneous prediction of the targets.}
    \label{fig:manifold_simultaneous_case_2}
\end{figure}

Therefore we return to the method based on the decay of the eigenvalues, and we select a five-dimensional parametrization of the manifold from the same kernel scale as the two-dimensional representation shown in Fig.~\ref{fig:manifold_Raman_case 2} on the manuscript. The five-dimensional representation is shown in Fig.~\ref{fig:manifold_simultaneous_case_2} and it is more accurate than the representations from the local regression, with a coefficient of determination for the test dataset of $R^2=0.9554$.

\subsection{Geometric Harmonics interpolation for an ``out-of-sample'' mapping}\label{SI_dmaps_Nyström}
After finding the low-dimensional manifolds shown in Fig.~\ref{fig:manifolds_Raman} on the manuscript, it is possible to map between the initial ambient space and the reduced space. This is to extend functions that are defined on the dataset and evaluate them outside of it, for points called ``out-of-sample''. This mapping from high to low dimensions is performed with the Nyström extension, which uses some functions defined on the dataset, called Geometric Harmonics~\cite{koronaki24, coifman06}.
This way we learn an ``out-of-sample'' mapping between the two spaces with a geometric harmonics interpolation. For the interpolations we need more eigenvalues and eigenvectors than for the dimensionality reduction, and thus a larger kernel scale. But since the interpolation includes a term where the eigenvalues are on the denominator, and eigenvalues become smaller as their number increases, numerical instabilities can occur for large a number of eigenvalues~\cite{koronaki24, DATAFOLD_Lehmberg2020}.

For the first case study we initially create the maximum achievable number of eigenvalues and eigenvectors for the dimensionality reduction, based on the number of the samples. 
After finding a parametrization for the low-dimensional manifold as described on Section~\ref{manifolds} on the manuscript, with 75$\%$ of the datasets for the training and 25$\%$ for the testing, we create the maximum achievable number of eigenvalues and eigenvectors for the mapping between the initial ambient space and the reduced space, which are less than those created for the dimensionality reduction. We keep the same kernel scale for both steps, since no better eigenvector predictions can be found based on the values of the coefficients of determination (\(R^2\)), Root-Mean-Squared-Error (RMSE) and Mean-Absolute-Error (MAE) metrics for the training and the testing.

\begin{figure}[ht] 
    \centering
    \begin{subfigure}{0.4\textwidth}
    \includegraphics[width=1\linewidth]{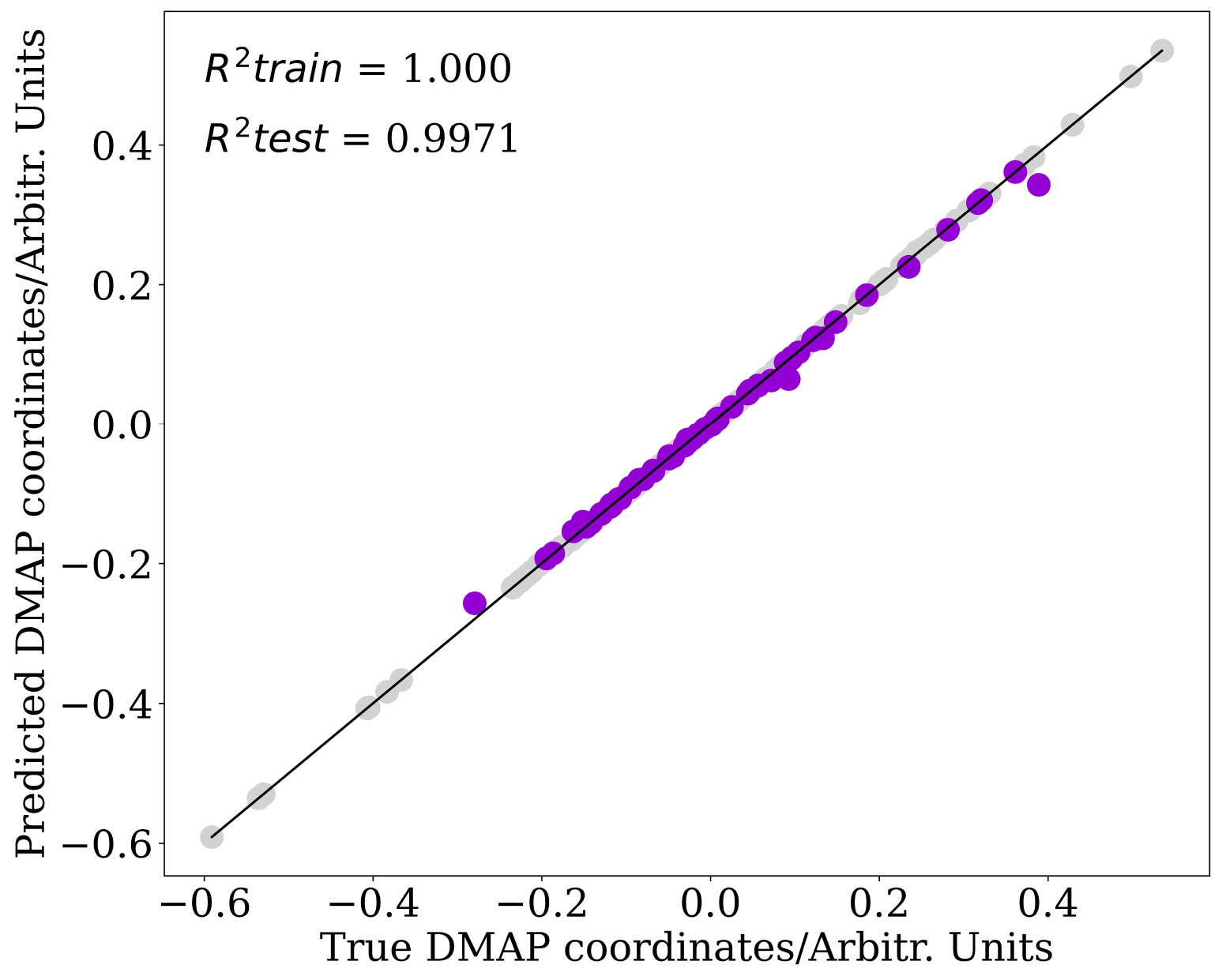}
    \caption{}
    \label{fig:dmaps_Nystroem_case_1}
    \end{subfigure}
    \hspace{0.07\textwidth}  
    \begin{subfigure}{0.4\textwidth}
    \includegraphics[width=1\linewidth]{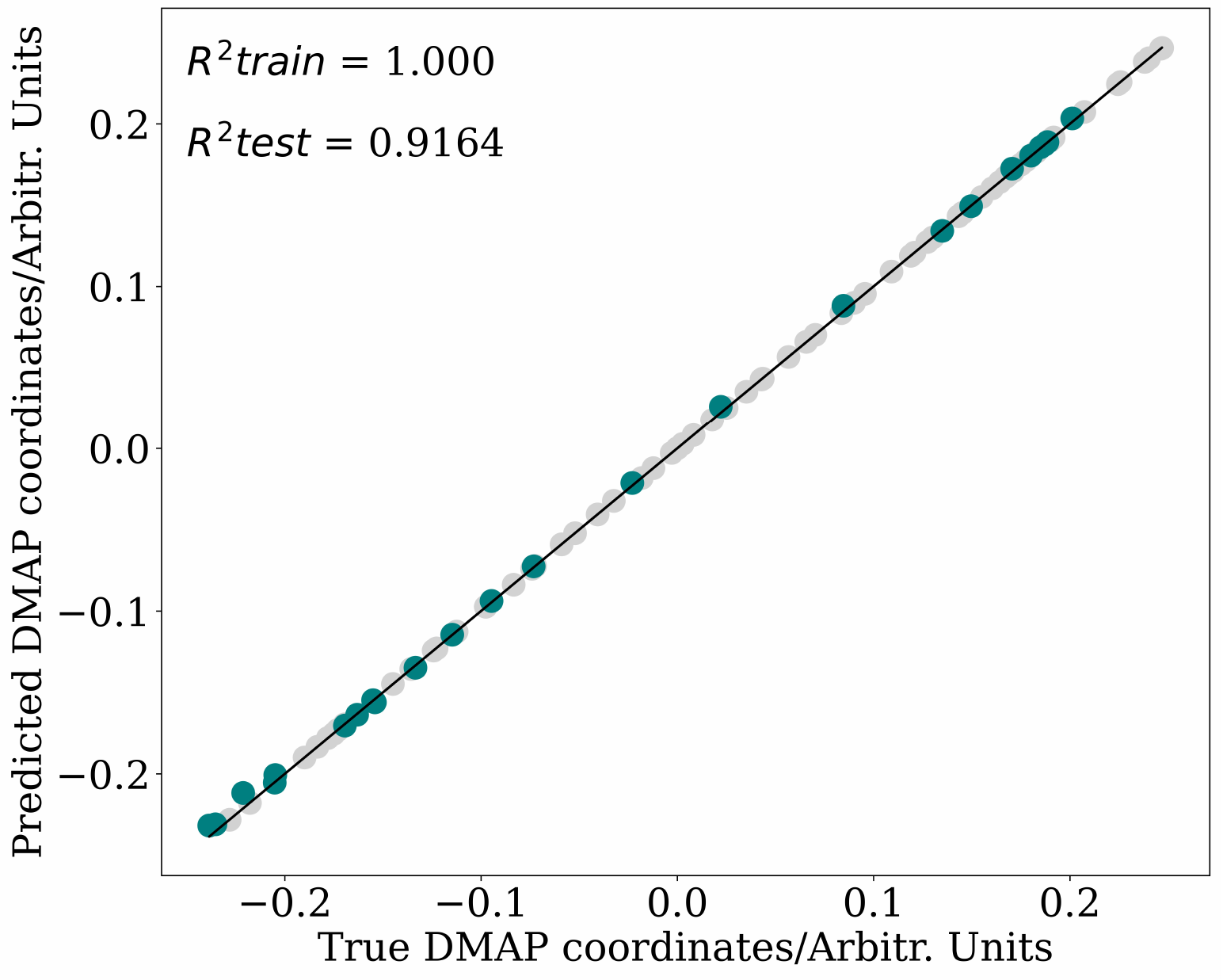}
    \caption{}
    \label{fig:dmaps_Nystroem_case_2}
    \end{subfigure}
    
    \caption{Predictions results from the interpolation with the Nyström extension, (a) for the first case study  and (b) for the second case study. The gray dots represent the training datapoints, while the colored ones represent the testing datapoints.}
    \label{fig:dmaps_Nystroem}
\end{figure}

For the second case study, since we have a larger dataset compared to the first case study, we initially create a few eigenvalues and eigenvectors for the dimensionality reduction, and then we create the maximum achievable number of eigenvalues and eigenvectors for the interpolation with the Nyström extension, which resulted to be one more than those created for the dimensionality reduction. The kernel scale however is neither the same nor larger than the one on the dimensionality reduction.
The parity plots with the prediction results for both case studies are shown in Fig.~\ref{fig:dmaps_Nystroem}

For the dimensionality reduction step, for both cases studies, we create also a different number of eigenvalues and eigenvectors than the initially ones, to see the impact on the maximum achievable number for interpolation step, but it remained the same. Additionally, we observe that the change on the number of eigenvalues and eigenvectors on the dimensionality reduction step does not affect their actual value in total, meaning that there are minor differences on the values or the signs of the eigenvectors, which at the end result on a similar, if not the same representation of the parametrization and reconstruction of the initial spectra.

\putbib 
\end{bibunit} 

\end{document}